%% file: NLTS.tex
\documentclass[letterpaper]{article} 
\usepackage{aaai2026}  

\usepackage{times}  
\usepackage{helvet}  
\usepackage{courier}  
\usepackage[hyphens]{url}  
\usepackage{graphicx} 
\usepackage{natbib}  
\usepackage{caption} 
\usepackage[most]{tcolorbox}   
\usepackage{listings}          
\usepackage{xcolor}            

\lstdefinestyle{pythonstyle}{
    language=Python,
    basicstyle=\ttfamily\scriptsize,
    keywordstyle=\color{blue!70!black}\bfseries,
    commentstyle=\color{green!50!black}\itshape,
    stringstyle=\color{orange!80!black},
    breaklines=true,
    breakatwhitespace=false,
    tabsize=4,
    showstringspaces=false
}

\newtcblisting{pythoncodebox}[1][]{
    listing only,
    listing options={style=pythonstyle},
    colback=black!5,
    colframe=blue!50!black,
    arc=0mm,
    boxrule=0.8pt,
    fonttitle=\bfseries,
    title=#1,
    left=1mm,
    enhanced,
    sharp corners
}

\usepackage{algorithm}
\usepackage{algpseudocode}
\usepackage{booktabs} 
\usepackage{multirow}
\usepackage{amsmath}
\usepackage{amsfonts}
\usepackage{mathtools}
\usepackage{amsthm}
\usepackage{algpseudocode}
\usepackage[capitalize,noabbrev]{cleveref}
\usepackage{placeins} 
\theoremstyle{plain}
\newtheorem{theorem}{Theorem}

\newtheorem{lemma}{Lemma}

\theoremstyle{definition}

\theoremstyle{remark}

\newcommand{\vx}{{\mathbf{x}}}

\newcommand{\vw}{{\mathbf{w}}}

\newcommand{\thref}[1]{Theorem~\ref{thm:#1}}
\newcommand{\lref}[1]{Lemma~\ref{lemma:#1}}

\newcommand{\eref}[1]{Eq.~\eqref{eq:#1}}
\newcommand{\fref}[1]{Fig.~\ref{fig:#1}}

\newcommand{\tref}[1]{Table~\ref{tab:#1}}
\newcommand{\aref}[1]{Algorithm.~\ref{algo:#1}}
\newcommand{\Aref}[1]{Appendix~\ref{appendix:#1}}

\def\figref#1#2{\includegraphics[width=#1\columnwidth]{figures/#2}}
\def\figclip#1#2#3{\includegraphics[trim=#1, clip, width=#2\columnwidth]{figures/#3}}


\nocopyright

\title{Enhancing Zero-Shot Time Series Forecasting in Off-the-Shelf LLMs via Noise Injection
}

\author{
Xingyou Yin\textsuperscript{\rm 1},
Ceyao Zhang\textsuperscript{\rm 2},
Min Hu\textsuperscript{\rm 1},
Kai Chen\textsuperscript{\rm 1}
}

\affiliations{
\textsuperscript{\rm 1}School of Mathematics and Statistics, Central South University (CSU), Changsha, China\\
\textsuperscript{\rm 2}Peking University, Beijing, China\\
232111109@csu.edu.cn, ceyaozhang@gmail.com, 242111129@csu.edu.cn, kaichen6@csu.edu.cn
}

\usepackage{bibentry}

\begin{document}

\maketitle


\begin{abstract}
Large Language Models (LLMs) have demonstrated effectiveness as zero-shot time series (TS) forecasters.
The key challenge lies in tokenizing TS data into textual representations that align with LLMs’ pre-trained knowledge. 
While existing work often relies on fine-tuning specialized modules 
to bridge this gap, 
a distinct, yet challenging, paradigm aims to leverage truly off-the-shelf LLMs without any fine-tuning whatsoever, relying solely on strategic tokenization of numerical sequences. The performance of these fully frozen models is acutely sensitive to the textual representation of the input data, as their parameters cannot adapt to distribution shifts. In this paper, we introduce a simple yet highly effective strategy to overcome this brittleness: injecting noise into the raw time series before tokenization. This non-invasive intervention acts as a form of inference-time augmentation, compelling the frozen LLM to extrapolate based on robust underlying temporal patterns rather than superficial numerical artifacts. 
We theoretically analyze this phenomenon and empirically validate its effectiveness across diverse benchmarks. 
Notably, to fully eliminate potential biases from data contamination during LLM pre-training, we introduce two novel TS datasets that fall outside all utilized LLMs' pre-training scopes, and consistently observe improved performance.
This study provides a further step in directly leveraging off-the-shelf LLMs for time series forecasting.
\end{abstract}


\section{Introduction}\label{sec:Intro}
\textbf{Time series (TS)} modeling plays a critical role in various real-world applications, including climate, economics, energy, and operations~\cite{wu2021autoformer,liu2023itransformer}. 
Accurate TS forecasting relies on the ability to model complex temporal dependencies in data, such as trends, seasonality, and nonlinearity, and to predict future values based on historical observations~\cite{montgomery2015introduction,che2018recurrent}. 
Traditional TS forecasters, such as ARIMA~\cite{box1968some}, nonlinear models~\cite{zhang2021fast,zhang2023data}, and Gaussian Processes~\cite[GP;][]{xu2020exact, dai2020interpretable,chen2022recent}, rely on human prior knowledge to select appropriate model configurations, for instance, kernel choices in GPs, to capture the underlying patterns and achieve accurate predictions.
While \textbf{deep learning}~\cite[DL;][]{lecun2015deep} has made impressive advances in NLP and CV, demonstrating that learned features can outperform human-designed features, these DL-based methods~\cite{zeng2023transformers} have also been extended to the TS domain.
However, both traditional methods and DL-based TS methods require training from scratch for a specific TS task.
Recently, \textbf{large language models (LLMs)}, such as GPT-3~\cite{brown2020language}, have demonstrated the ability to perform downstream tasks without the need for fine-tuning, enabling zero-shot learning. 
Based on this capability, ~\citet{LLMTime} proved that LLMs can also serve as zero-shot TS forecasters by tokenizing time series data.

As the use of LLM-based time series methods—hereafter referred to as \textit{LLMs-for-TS}—forecasting has increased, more strategies have been developed to improve their performance in TS modeling. 
There are two primary paradigms for TS forecasting of LLMs~\cite{TEST}. 
The first paradigm focuses on designing and training a dedicated TS-specific large model from scratch or by fine-tuning an existing pre-trained LLM to transfer it from textual to temporal domains.
Representatives include LLM4TS~\cite{LLM4TS}, which adapts LLMs to time series through autoregressive supervised fine-tuning, and CALF~\cite{liu2025calf}, which employs a dual-branch architecture with attention-based embeddings.
The second paradigm involves freezing the parameters of the existing LLM and designing TS representations that are compatible with them.
LLMTime~\cite{LLMTime} directly inputs raw numeric time series data as text prompts, and TEST~\cite{TEST} aligns temporal instances with textual prototypes.
This paradigm maximizes the model's capabilities by aligning the input data with the LLM's architecture, allowing it to process TS effectively without requiring extensive retraining.

Meanwhile, the term \textbf{\textit{zero-shot}} is ambiguous in different LLM-based time series methods; we disambiguate it as follows. The first type is \textbf{zero-shot under trainable settings}, which covers all first LLM-for-TS paradigm and parts of the second paradigm, such as TEST, which freeze the LLM but train the other modules. 
Under this setting, \textit{zero-shot} refers to the model is first fine-tuned on source time series A before inference on target time series B.
In comparison, \textbf{zero-shot under training-free setting} directly applies LLMs to input queries without any task-specific adaptation or fine-tuning, such as PromptCast~\cite{xue2023promptcast}, LLMTime~\cite{LLMTime}. 
We focus on this more challenging yet practically valuable \textbf{Off-the-Shelf LLMs} setting, as it most purely tests an LLM's inherent reasoning capabilities and offers the lowest barrier to deployment by eliminating the need for any training data or computational budget for fine-tuning.

However, the performance of off-the-shelf LLMs is acutely dependent on the textual representation of the continuous time series data. A central challenge in this paradigm is the brittleness of the tokenized input: the forecasting accuracy can be highly sensitive to the specific numerical representation and vulnerable to distribution shifts, as the model's parameters are frozen and cannot adapt. 
To address this, we turn to a powerful yet under-explored strategy for enhancing model robustness: input-level \textbf{noise injection}.
Traditionally employed during model training for regularization and data augmentation~\cite{trirat2024universal,ma2024survey,abnar2021exploring,jin2023large}, 
we reconceptualize its application for inference-time robustness in frozen LLMs. We hypothesize that strategically perturbing the time series with noise before tokenization can act as a powerful form of input augmentation, compelling the LLM to base its predictions on more stable, underlying temporal structures rather than on the precise, and potentially misleading, numerical representation. This approach is fundamentally non-invasive, aligning perfectly with the off-the-shelf paradigm by enhancing performance without any manipulation of the model's internal parameters or the need for retraining.

Translating this hypothesis into a practical framework, we introduce Noise-injected LLM for Time Series (NLTS).
Our method operates by first injecting controlled stochastic noise into the raw time series. This perturbed series is then converted into a discrete token sequence through a meticulous textualization and tokenization process that preserves bijectivity between numerical and symbolic representations. This noised, tokenized prompt is fed directly into the frozen LLM, which then performs the autoregressive \textit{next-token prediction}. To aggregate predictions and estimate uncertainty, we sample multiple forecasts by generating varied noisy instances of the input prompt, then compute the median and variance across these samples. This entire pipeline requires no backpropagation, fine-tuning, or internal access to the LLM.
Specifically, our contributions are as follows.
\begin{itemize}

\item We propose a noise-injection strategy that equips off-the-shelf LLMs with enhanced zero-shot forecasting capabilities without any task-specific fine-tuning.

\item 
We provide both theoretical guarantees and exhaustive empirical validation on established time series benchmarks, demonstrating consistent gains across different LLMs and various settings.


\item We expose data contamination risks inherent in LLM-based zero-shot forecasting, and design two new datasets to eliminate this risk. Experiments further validate the effectiveness of our method.

\end{itemize}

\section{Related Work}\label{sec:Bg}
\textbf{LLMs for time series forecasting}
PromptCast~\cite{xue2023promptcast} was the first study to apply pre-trained LLMs to TS forecasting. 
Subsequently, LLMTime~\cite{LLMTime} innovatively inputs numeric TS directly into LLMs, transforming TS forecasting into a "next-word prediction" task. This shift has enabled LLMs like GPT-3.5 ~\cite{brown2020language}and LLaMA-2~\cite{touvron2023llama2} to excel in zero-shot forecasting tasks, further demonstrating the potential of LLMs in TS tasks.
On the other hand, significant progress has been made in block-based representation methods for TS~\cite{nie2022time}. 
Examples include One Fits All (OFA)~\cite{zhou2023one}, LLM4TS~\cite{LLM4TS}, TEST~\cite{TEST}, TEMPO~\cite{Tempo} and TimeLLM~\cite{Time-llm}, all of which employ block-based methods to tokenize TS, making them compatible with LLM and improving performance. 
All these methods ignore the impact of noise injection in LLMs-for-TS. 

\noindent\textbf{Impact of noise in machine learning}
In traditional TS analysis~\cite{gao2009denoising}, noise is often considered a disruptive factor, prompting widespread use of denoising techniques to reduce noise levels and improve model prediction accuracy and robustness. 
However, recent research~\cite{zhang2007neural,nourani2018hybrid} has revealed that rather than simply removing noise, injecting noise as a data augmentation strategy can significantly enhance model robustness under certain conditions. 
Nourani et al.~\cite{nourani2018hybrid} demonstrated the potential of noise to increase training data diversity and enhance model performance. Magklaras et al.~\cite{magklaras2019noise} suggest that noise injection helps identify malicious data. Kim et al.~\cite{kim2024extraction} integrated noise injection with digital signal processing techniques, using frequency feature extraction to improve the anti-interference ability of classification models. 

\section{Method}
\label{sec:framework}
We systematically investigate how to enhance TS forecasting in LLMs through data perturbation with noise injection and how noisy prompts influence predictive robustness and generalization. 
We introduce a novel framework, the \textbf{N}oise Injection Augmented \textbf{L}LM for \textbf{TS} (NLTS), which comprises several critical components, including: 
1) noise design and sampling strategies for generating diverse noise patterns, 2) data perturbation with noise injection, 3) textualization of a point in TS to convert numerical data of a point into a descriptive textual format, 4) data transformation and tokenization of full TS to enable a LLM-friendly representation of TS, 5) prompt formulation and token prediction, and 6) sampling and aggregation approaches to compute performance consistency across different forecasting tasks. 
Beyond its simplicity, this approach provides a significant advantage over prevalent LLM-based zero-shot TS forecasting methods. 
As shown in~\fref{overview}, we present two approaches for zero-shot TS forecasting using LLMs.
Our NLTS (bottom subplot) is purely zero-shot, data-agnostic, and LLM-agnostic, allowing it to seamlessly adapt to various LLMs and tasks.

\subsection{TS Forecasting Problem Formulation}

Mathematically, a TS $ \vx = \{ x_t \}_{t=1}^{T} $ can be represented as the sum of a signal $ \{ f(t) \}_{t=1}^{T} $ and noise $ \{ \epsilon_t \}_{t=1}^{T} $, such that: $x_t = f(t) + \epsilon_t$,
where $ f(t) $ denotes the true underlying signal at time $ t $, and $ \epsilon_t $ represents the noise term at time $ t $. In the context of TS forecasting, noise refers to random fluctuations or disturbances within the data that cannot be accounted for by the underlying trend, seasonality, or other systematic patterns. The goal of TS forecasting is to predict the future values $ \{ x_{T+1}, x_{T+2}, \dots, x_{T+H} \} $, with $ H $ representing the forecast horizon. Thus, the TS forecasting problem can be formulated as estimating the conditional distribution of future values given past observations: $ p(\{ x_t \}_{t=T+1}^{T+H} | \{ x_t \}_{t=1}^{T}) $.

\subsection{TS Forecasting in Off-the-Shelf LLMs}
\subsubsection{Token modeling in LLM.}
LLMs are trained on sequential data, $ \mathcal{S} = \{ S_1, S_2, \ldots, S_i, \ldots, S_N \} $, where each sequence $ S_i $ consists of tokens $(\mathbf{s}_{i,1}, \mathbf{s}_{i,2}, \ldots, \mathbf{s}_{i,j}, \ldots, \mathbf{s}_{i,n_i})$, with each token $\mathbf{s}_{i,j} $ from a vocabulary $ \mathcal{V} $. These models encode an autoregressive distribution, where the probability of each token depends only on preceding tokens: $ p_\Theta(S_i) = \prod_{j=1}^{n_i} p_\Theta(\mathbf{s}_j \mid \mathbf{s}_{0:j-1}) $, and the model parameters $ \Theta $ are optimized by maximizing the likelihood of the entire dataset: $ p_\Theta(\mathcal{S}) = \prod_{i=1}^N p_\Theta(S_i) $.

\subsubsection{Token prediction of TS.}
Sampling from a trained language model typically begins with an initial prompt $\mathbf{s}_{0:k}$ and progresses iteratively, selecting the subsequent token based on $ p_\Theta(\mathbf{s}_j \mid \mathbf{s}_{0:j-1}) $. 
In the case of LLMs, TS forecasting is conceptualized as a sequence generation task. 
This autoregressive process can be mathematically expressed as: $p(\text{Token}(\tilde{x}_{T+k}) | \{\text{Token}(\tilde{x}_{t})\}_{t=1}^{T+k-1})$.
Consequently, the conditional distribution is approximated as $$p(\tilde{x}_{T+k}| \{ \tilde{x}_t \}_{t=1}^{T+k-1}) \approx p(\text{Token}(\tilde{x}_{T+k}) | \{\text{Token}(\tilde{x}_{t})\}_{t=1}^{T+k-1}).$$

\subsubsection{Sampling and aggregating outputs of LLM.}
In addition to the insights gleaned from individual model predictions, drawing multiple predictions from an LLM can yield valuable indications regarding the overall central tendency and confidence of LLM-based forecasting. 
Suppose we intend to approximate the central tendency of the forecasting function $ f(t) $ under a probability distribution $ p(t) $ across a domain $ \mathcal{X} $. For the testing set $\vx_{*} = \{\tilde{x}_{T+1}, \tilde{x}_{T+2}, \dots, \tilde{x}_{T+H}\}$, we define 
\begin{align}\label{eq:reverse}
\vx_{*} = \mathcal{Q}^{-1}(\text{Token}(S_{*})),
\end{align}
where 
$S_{*} = \{\text{Token}_{t}(\tilde{x}_{t}) \}_{t=T+1}^{T+H}$. 
Additionally, we assess the uncertainty of each point via the empirical variance, furnishing a measure of the prediction's reliability.
We estimate the central tendency of the prediction by using the sample median. Specifically, we order multiple predictive points in non-decreasing sequence to obtain the order statistics:  $x_{h}^{(1)} \leq x_{h}^{(2)} \leq \dots \leq x_{h}^{(m)}$. The sample median \(\bar{x}_{*}\) is defined as:  
\begin{align}
   \bar{x}_{*} = 
\begin{cases} 
x_{*, \frac{h + 1}{2}}^{(m)} & \text{if } h \text{ is odd}, \\
\frac{1}{2} \left( x_{*, \frac{h}{2}}^{(m)} + x_{*,  \frac{h}{2} + 1}^{(m)} \right) & \text{if } h \text{ is even},
\end{cases}
\end{align}
where $x_{*, i}^{(m)}$ denotes the $i$-th point of LLM-generated forecasting, and $ m $ is the number of LLM model generations. This aggregation contributes to reducing the effect of individual responses while offering a more robust estimate of the true forecast. 
In addition, the confidence interval of pointwise prediction can be achieved by leveraging the quantile and variance of $m$ predictions. 
Finally, we adopt $\bar{\vx}_{*}$ as the overall ultimate TS forecast in LLM. Note that we do not have to eliminate the introduced noise from $\vx_{*}^{(m)}$ or $\bar{\vx}_{*}$, because we believe the LLM has the capacity of noise-based self-correction, adaptively learns to disregard "irrelevant" noise, and concentrates on salient patterns, thereby substantially bolstering its overall predictive performance and forecasting accuracy.

\begin{figure}[t]
\renewcommand{\tabcolsep}{-0.5mm}
\centering
\begin{tabular}{p{0.0mm}*{1}{c}}
& \figref{0.95}{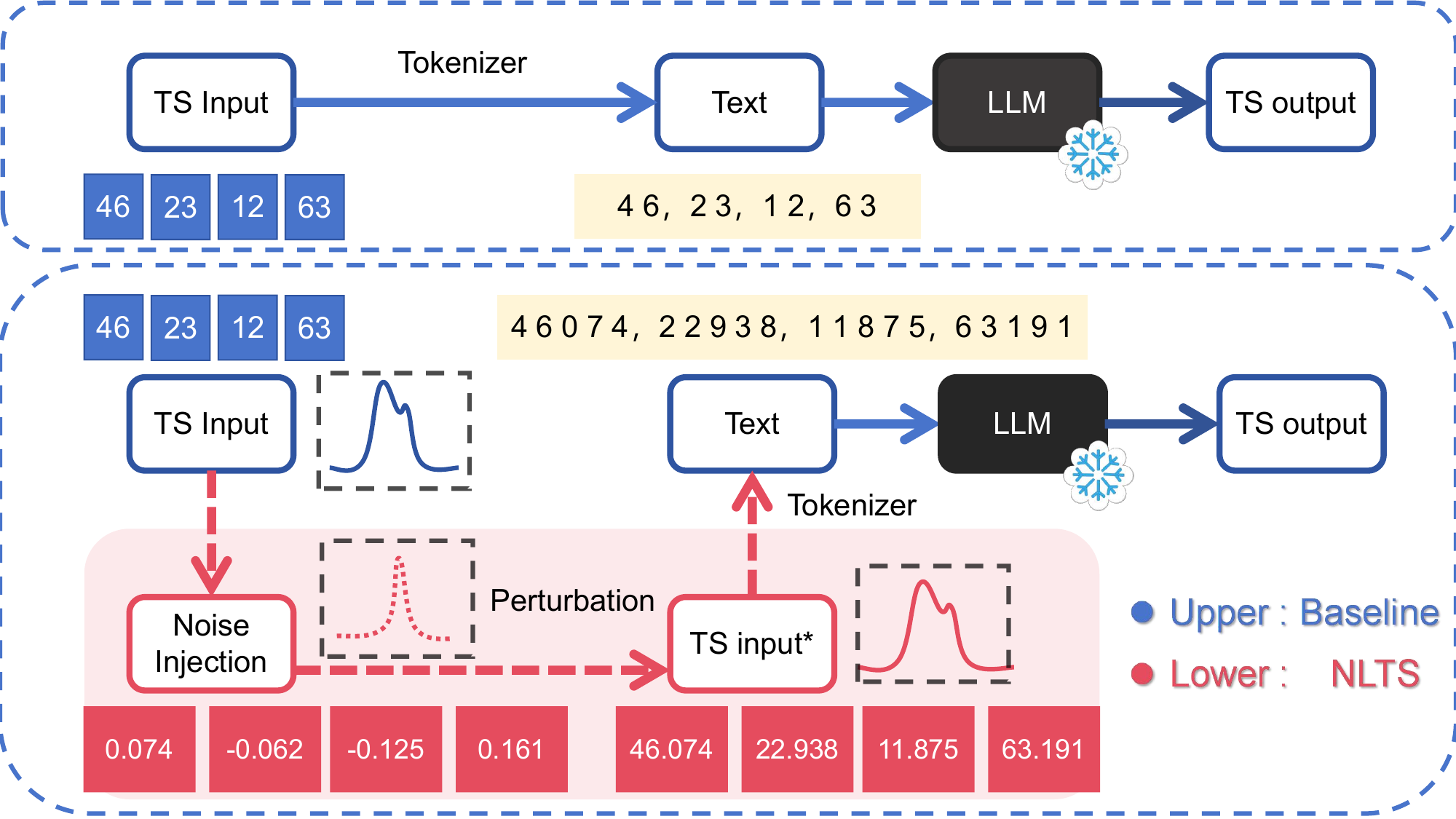} \\
\end{tabular}
\caption{Overview of zero-shot TS forecasting in off-the-shelf LLMs: the top is a vanilla usage of off-the-shelf LLM for TS, where the numerical values are tokenized and directly converted into a string, and then fed into a frozen LLM for prediction. The bottom is our NLTS framework, which introduces noise injection.}
\label{fig:overview}
\end{figure}

\subsection{TS Forecasting in Off-the-Shelf LLMs via Noise Injection}

\subsubsection{Noise design and sampling strategies.} 
The noise term is commonly represented as a random variable with a mean of zero and a variance of $ \sigma^2 $. The variance $ \sigma^2 $ reflects the level of uncertainty or randomness inherent in the observations. For instance, noise that follows a Gaussian distribution is expressed as $\epsilon_i \sim \mathcal{N}(0, \sigma^2)$, where $\epsilon_i$ denotes the noise to be injected. 
In addition to the Gaussian distribution, we examine four other noise distributions, including uniform, Laplace, Gamma, and Beta distributions. Further details can be found in~\Aref{noise_distribution}.

We also introduce noise sampling with noise scaling, which entails the controlled adjustment of the noise magnitude or intensity relative to the underlying data, ensuring that the noise does not overwhelm the signal, yet remains sufficiently impactful to affect model behavior and performance. This noise scaling is essential for balancing the model's sensitivity and robustness. Specifically, we parameterize the noise variance $ \sigma^2 $ as $\sigma^2 = \alpha^2 \sigma_x^2$, 
where $ \sigma_x^2 $ is the variance of the original TS, and $ \alpha $ is the scaling factor that governs the noise magnitude relative to the original value $ x_t $. The noise is jointly determined by the standard deviation $ \sigma_x $ of the original data and the scaling factor $ \alpha $.

\subsubsection{Noise injection on TS.}
The external noise injection introduces an element of stochasticity into the model's input, effectively simulating data perturbation.
Given a TS $ x_t \in \mathbb{R} $, \textit{noise injection} is a stochastic perturbation operator $ \mathcal{P}: \mathbb{R} \rightarrow \mathbb{R} $ defined as:
\begin{equation}\label{equ:noise_injection}
    \mathcal{P}(x_t) = \tilde{x}_t = x_t + \epsilon_i,
\end{equation}
where $ \epsilon_i$ is sampled from a noise distribution. This operator induces controlled variability into the input.

By enforcing $ \tilde{x}_t $ to approximate $ x_t $ under $ \mathbb{E}[\epsilon_i] = 0 $, the model is regularized to prioritize latent signal structures over spurious fluctuations, thereby enhancing generalization without architectural modifications or retraining.  
The perturbed series $ \{\tilde{x}_t\} $ is propagated through subsequent LLM-based forecasting steps, optimizing robustness to distributional shifts. 

\subsubsection{Textualization of point in TS.}
The textualization of a TS point involves the conversion of quantitative data, such as numerical values, into a natural language format. This process is essential for enabling LLMs to comprehend, interpret, and generate human-readable representations of numerical information.

Given a point of TS $x_{t}$,  its \textit{textualization} is a bijective mapping $ \mathcal{T}: \mathbb{R} \rightarrow \mathcal{S} $ to a linguistically structured format via:  
1) digit separation: $ \mathcal{T}_1(x_t) = d_1 \ \text{\textvisiblespace}\ d_2 \ \text{\textvisiblespace}\ \cdots\ \text{\textvisiblespace}\ d_n $, where $ d_i $ are decimal digits of $ x_t $, ensuring token-wise independence;  
2) precision scaling: For fixed $ k \in \mathbb{N} $, the integer representation $ \tilde{x}_{t, \text{int}}=\mathcal{T}_2(x_t) = \lfloor x_t \times 10^k \rfloor $, discarding decimals while preserving invertibility via $ \mathcal{T}_2^{-1}(\tilde{x}_{t, \text{int}}) = \tilde{x}_{t, \text{int}} / 10^k $.  
The composite function $ \mathcal{T}(x_t) = \mathcal{T}_1(\mathcal{T}_2(x_t)) $ encodes $x$ into interpretable textual tokens, bridging continuous TS with discrete symbolic frameworks~\cite{LLMTime}.  

Specifically, 
inserting spaces between the digits of a point, ensuring that each digit is treated as a distinct token. For instance, the number "56789" becomes "5 6 7 8 9", preventing standard tokenization methods from treating the entire number as a single token and instead enabling the model to process each digit independently. 
Furthermore, by removing the decimal point and representing numbers as integers, we achieve a more concise and efficient representation. 
For example, if $ \tilde{x}_{t} = 2.718 $ and $ k = 3 $, the integer representation becomes $ 2718 $. To reconstruct $ \tilde{x}_{t} $, we compute $ \tilde{x}_{t} = 2718 / 10^3 = 2.718 $.

\subsubsection{Tokenization of noised TS.} 
The meticulous tokenization of TS is pivotal in improving the precision and dependability of predictions made by LLMs. Transformer-based LLMs, such as GPT-3.5 and DeepSeek, are inherently designed to process token sequences, typically in textual form, which necessitates the adaptation of TS into a compatible format for these models.

Let $ \{\tilde{x}_t\}_{t=1}^T \subset \mathbb{R} $ denote a noised TS derived from data perturbation (Eq. \eqref{equ:noise_injection}). A \textit{tokenization operator} $ \mathcal{Q}: \mathbb{R}^T \rightarrow \mathcal{S}^T $ is a bijective mapping that converts $ \{\tilde{x}_t\} $ into a discrete token sequence $ S = \{\mathrm{Token}_t(\tilde{x}_t)\}_{t=1}^T $, where $ \mathcal{S} $ is the token vocabulary. The operator satisfies:  
\begin{align}\label{eq:token}
S = \mathcal{Q}(\{\tilde{x}_t\})  \quad \text{and} \quad \{\tilde{x}_t\} = \mathcal{Q}^{-1}(S),
\end{align}
ensuring invertibility between numerical and symbolic representations. 

The bijectivity constraints $ \mathcal{Q} \circ \mathcal{Q}^{-1} = \mathrm{Id}_{\mathcal{S}^T} $ and $ \mathcal{Q}^{-1} \circ \mathcal{Q} = \mathrm{Id}_{\mathbb{R}^T} $ guarantee structural fidelity, where $ \mathrm{Id} $ denotes identity mapping. Crucially, $ \mathcal{Q} $ preserves temporal semantics by aligning token embeddings $ \{\mathrm{Token}_t\} $ with the perturbed dynamics $ \{\tilde{x}_t\} $, enabling LLMs to process noisy numerical sequences as contextually coherent text. This tokenization-reconstruction duality ensures that stochastic variations introduced during noise injection remain interpretable within the LLM's embedding space.  
Specifically, we adopt the tokenization method for TS as described in LLMTime~\cite{LLMTime}. The tokenization function $\mathcal{Q}$ utilizes commas to delineate individual time points, treating each time step as a discrete input, with time steps separated by commas. For example, the TS $[22, 25, 28]$ is encoded as "2 2, 2 5, 2 8". The comma separation is crucial as it allows the model to distinguish between distinct time steps and preserve their sequential order. For instance, by combining textualization and tokenization, the TS $[0.314, 3.14, 31.4, 314.0]$ is represented as "3 1, 3 1 4, 3 1 4 0, 3 1 4 0 0". This sequence $ S $ can now be input into the LLM. 



\section{Theoretical Analysis of LLM with NLTS}
\begin{theorem}[First- and second-order optimality for well-trained LLMs] \label{thm:optimality}  
Let $ \hat{\mathcal{L}}(\Theta) =p_\Theta(\mathcal{S})
$ denotes the empirical log-likelihood of an LLM over training datasets. A parameter configuration $ \Theta^* $ maximizes $ \hat{\mathcal{L}}(\Theta) $ and defines a well-trained LLM if: first-order optimality with $\nabla_\vw \hat{\mathcal{L}}(\Theta^*) = 0$ and second-order optimality with the expected Hessian $ \mathbb{E}[H_{f}(\Theta^*)] = \mathbb{E}[\nabla_\vw^2 \hat{\mathcal{L}}(\Theta^*)] $ is negative definite. 
\end{theorem}  
The negative definiteness of the expected Hessian reflects the intrinsic concavity of the likelihood function in identifiable parameter regimes.

\begin{lemma}[Perturbation stability of well-trained LLMs]\label{lemma:perturbation_stability}  
Let $ f_\Theta$ be a well-trained LLM satisfying \thref{optimality}. For any perturbed input $ \tilde{x}_t$ and target function $ h(x) $, the following inequality holds: $\mathbb{E}_{\epsilon}\left[ f_{\Theta^*}(\tilde{x}_t
) - h(x) \right] \leq \mathbb{E}_{\epsilon}\left[ f_{\Theta^*}(x) - h(x) \right]$.
\end{lemma}
\lref{perturbation_stability} indicates that the generalization error of LLM can be reduced by noise injection on the input for zero-shot forecasting. We present the proofs of \thref{optimality} and \lref{perturbation_stability} in~\Aref{Proof of Theorem}. 

\section{Zero-Shot Forecasting Experiments}\label{sec:exp}
In this section, we comprehensively assess the NLTS framework's effect on the TS forecasting performance across LLMs. 

\subsection{Setup} 
\textbf{Datasets}.
We employ three recognized benchmark datasets: Autoformer~\cite{wu2021autoformer}, Darts~\cite{herzen2022darts}, and Memorization~\cite{LLMTime}.  
The details of these datasets are presented in ~\Aref{dataset}.  

\begin{table*}[tbph]
\centering
\small
\setlength{\tabcolsep}{2pt} 
\begin{tabular}{llccccccccccccccc}
\toprule
\multirow{2}{*}{Benchmarks} & \multirow{2}{*}{Datasets} & \multirow{2}{*}{\shortstack{Prediction \\ Length}} & 
\multicolumn{2}{c}{\shortstack{NLTS 
}} & 
\multicolumn{2}{c}{\shortstack{LLMTime 
}} & 
\multicolumn{2}{c}{\shortstack{N-HiTS 
}} & 
\multicolumn{2}{c}{\shortstack{N-BEATS 
}} & 
\multicolumn{2}{c}{\shortstack{TCN 
}} & 
\multicolumn{2}{c}{\shortstack{SM-GP 
}} & 
\multicolumn{2}{c}{\shortstack{ARIMA 
}} \\
\cmidrule(lr){4-5} 
\cmidrule(lr){6-7} 
\cmidrule(lr){8-9} 
\cmidrule(lr){10-11} 
\cmidrule(lr){12-13} 
\cmidrule(lr){14-15} 
\cmidrule(lr){16-17}
& & & MSE & MAE & MSE & MAE & MSE & MAE & MSE & MAE & MSE & MAE & MSE & MAE & MSE & MAE \\
\midrule
\multirow{8}{*}{Darts} 
& AirPassengers & 29 & \textbf{0.003} & \textbf{0.044} & 0.010 & 0.075 & 0.025 & 0.118 & 0.028 & 0.126 & 0.048 & 0.172 & 0.008 & 0.073 & \underline{0.006} & \underline{0.060} \\
& AusBeer & 43 & \textbf{0.001} & \textbf{0.020} & \underline{0.001} & \underline{0.026} & 0.027 & 0.154 & 0.006 & 0.059 & 0.011 & 0.094 & 0.056 & 0.186 & 0.002 & 0.032 \\
& GasRateCO2 & 60 & \textbf{0.019} & \textbf{0.106} & 0.038 & 0.160 & 0.052 & 0.183 & 0.106 & 0.271 & 0.050 & 0.186 & 0.033 & 0.157 & \underline{0.029} & \underline{0.151} \\
& HeartRate & 180 & \textbf{0.027} & \textbf{0.131} & \underline{0.035} & \underline{0.157} & 0.114 & 0.279 & 0.051 & 0.174 & 0.048 & 0.184 & 0.063 & 0.208 & 0.039 & 0.162 \\
& MonthlyMilk & 34 & \textbf{0.002} & \textbf{0.042} & \underline{0.005} & 0.062 & 0.006 & \underline{0.060} & 0.011 & 0.088 & 0.025 & 0.138 & 0.014 & 0.104 & 0.014 & 0.108 \\
& Sunspots & 141 & \textbf{0.051} & \textbf{0.166} & 0.082 & 0.208 & 0.070 & 0.191 & 0.135 & 0.287 & 0.068 & 0.181 & 0.088 & 0.223 & \underline{0.061} & \underline{0.176} \\
& Wine & 36 & \textbf{0.008} & \textbf{0.067} & 0.014 & \underline{0.086} & 0.044 & 0.164 & 0.046 & 0.140 & 0.025 & 0.127 & 0.047 & 0.182 & \underline{0.012} & 0.087 \\
& Wooly & 24 & \textbf{0.010} & \textbf{0.079} & 0.014 & 0.103 & \underline{0.012} & \underline{0.088} & 0.036 & 0.178 & 0.022 & 0.132 & 0.031 & 0.140 & 0.028 & 0.156 \\
\midrule
\multirow{3}{*}{Memorization}
& IstanbulTraffic & 30 & \textbf{0.060} & \textbf{0.181} & 0.136 & 0.330 & 0.259 & 0.401 & 0.399 & 0.573 & \underline{0.122} & \underline{0.304} & 0.229 & 0.385 & 0.154 & 0.310 \\
& TSMCStock & 30 & \textbf{0.0003} & \textbf{0.014} & \underline{0.0004} & \underline{0.016} & 0.001 & 0.018 & 0.029 & 0.161 & 0.002 & 0.039 & 0.014 & 0.108 & 0.465 & 0.584 \\
& TurkeyPower & 30 & \textbf{0.001} & \textbf{0.023} & \underline{0.002} & \underline{0.032} & 0.018 & 0.114 & 0.019 & 0.126 & 0.004 & 0.046 & 0.015 & 0.103 & 0.003 & 0.047 \\
\bottomrule 
\end{tabular}
\caption{
Zero-Shot Forecasting Performance on \textit{Short-Term Time Series}. The evaluation setting follows LLMTime~\cite{LLMTime}: Darts uses 80\% of each time series as prompt input and reserves 20\% for testing. Memorization benchmark forecasts the next 30 time steps. \textbf{Bold} and \underline{underline}: the best and the second best performance.}
\label{tab:mse-mae-short}
\end{table*}

\begin{table*}[tbph]
\centering
\small
\setlength{\tabcolsep}{3pt}
\begin{tabular}{llccccccccccccccc}
\toprule
\multirow{2}{*}{Benchmarks} & 
\multirow{2}{*}{Datasets} & 
\multirow{2}{*}{\shortstack{Prediction \\ Length}} & 
\multicolumn{2}{c}{\shortstack{NLTS 
}} & 
\multicolumn{2}{c}{\shortstack{iTransformer 
}} & 
\multicolumn{2}{c}{\shortstack{LLMTime 
}} & 
\multicolumn{2}{c}{\shortstack{PatchTST 
}} & 
\multicolumn{2}{c}{\shortstack{TimesNet 
}} & 
\multicolumn{2}{c}{\shortstack{Autoformer 
}} & 
\multicolumn{2}{c}{\shortstack{Informer 
}} \\
\cmidrule(lr){4-5} 
\cmidrule(lr){6-7} 
\cmidrule(lr){8-9} 
\cmidrule(lr){10-11} 
\cmidrule(lr){12-13} 
\cmidrule(lr){14-15} 
\cmidrule(lr){16-17}
& & & MSE & MAE & MSE & MAE & MSE & MAE & MSE & MAE & MSE & MAE & MSE & MAE & MSE & MAE \\
\midrule
\multirow{7}{*}{Autoformer}
& ECL & 96 & \textbf{0.014} & \textbf{0.094} & 0.084 & 0.210 & \underline{0.027} & \underline{0.127} & 0.124 & 0.289 & 0.076 & 0.208 & 0.235 & 0.364 & 0.124 & 0.364 \\
& ETTh1 & 96 & \textbf{0.005} & \textbf{0.052} & 0.038 & 0.163 & \underline{0.008} & \underline{0.077} & 0.071 & 0.239 & 0.080 & 0.262 & 0.071 & 0.234 & 0.039 & 0.234 \\
& ETTh2 & 96 & \textbf{0.024} & \textbf{0.117} & 0.214 & 0.359 & \underline{0.042} & \underline{0.148} & 0.331 & 0.478 & 0.314 & 0.483 & 0.184 & 0.324 & 0.174 & 0.324 \\
& ETTm1 & 96 & \textbf{0.002} & \textbf{0.035} & \underline{0.003} & \underline{0.044} & 0.004 & 0.045 & 0.006 & 0.056 & 0.008 & 0.069 & 0.009 & 0.078 & 0.004 & 0.078 \\
& ETTm2 & 96 & \textbf{0.020} & \textbf{0.114} & \underline{0.021} & \underline{0.117} & 0.028 & 0.142 & 0.026 & 0.139 & 0.021 & 0.124 & 0.077 & 0.249 & 0.075 & 0.249 \\
& Traffic & 96 & \textbf{0.001} & \textbf{0.027} & 0.091 & 0.257 & \underline{0.010} & \underline{0.069} & 0.075 & 0.227 & 0.077 & 0.228 & 0.073 & 0.213 & 0.079 & 0.213 \\
& ILI & 24 & \textbf{0.081} & \textbf{0.106} & 0.426 & 0.486 & \underline{0.084} & \underline{0.115} & 0.979 & 0.687 & 0.766 & 0.765 & 1.077 & 0.969 & 11.047 & 3.312 \\
\bottomrule
\end{tabular}
\caption{
Zero-Shot Forecasting Performance on \textit{Long-Term Time Series}. All baselines follow the experimental setup from LLMTime, using the prediction length as the number of test steps. Additional results for various prediction horizons are provided in Appendix~\ref{appendix:forecasting_horizons}.\textbf{ Bold}: best performance, \underline{underline}: the second best.}
\label{tab:mse-mae-long}
\end{table*}

\noindent\textbf{Models and baselines}. 
We examine nine representative LLMs 
: GPT-4~\cite{openai2023gpt4}, GPT-3.5-Turbo-Instruct~\cite{brown2020language}, Moonshot-V1-8k, Claude-3-Opus, Claude-3.5-Sonnet, Claude-3.5-Haiku~\cite{anthropic2024claude35sonnet}, DeepSeek-V3~\cite{deepseekv3}, GLM-4-Air~\cite{GLM2024chatGLM}, and Qwen3-4B~\cite{qwen34b}.
Observe that we still choose not to employ the newest or more sophisticated LLMs, including higher versions of ChatGPT, owing to their prohibitive cost.
In contrast, we choose popular open-source and closed-source LLMs that span various native language backgrounds, ensuring substantial representativeness and diversity.
Furthermore, the study performs rigorous comparative analysis, contrasting these LLMs' performance with that of non-LLM baseline models, including ARIMA, SM-GP, Temporal Convolution Networks~\cite[TCN;][]{lea2016temporal}, N-BEATS, N-HiTS, and a range of advanced TS forecasting models such as Informer~\cite{zhou2021informer},  Autoformer, NSTransformer~\cite{liu2022non}, TimesNet~\cite{wutimesnet}, PatchTST~\cite{nietime}, and iTransformer~\cite{liu2023itransformer}. Additionally, the zero-shot forecasting model LLMTime is included to further highlight the relative effectiveness of LLMs in this domain. It is important to emphasize that our method is directly built upon the LLMTime framework, using the same model architecture and data splits. 

While recent works have proposed several LLM-based forecasting frameworks, such as TEST~\cite{TEST}, TimeLLM~\cite{Time-llm}, and CALF~\cite{liu2025calf}, we do not include them as direct baselines, as their so-called zero-shot forecasting involves optimizing the model on one dataset and evaluating it on another. In contrast, we directly apply off-the-shelf LLMs without any task-specific fine-tuning.

\noindent\textbf{Metrics}. For performance evaluation of all methods, we employ two widely used regression metrics: Mean Squared Error (MSE) and Mean Absolute Error (MAE). 
\begin{figure*}[th]
\renewcommand{\tabcolsep}{1.0mm}
\scriptsize
\centering
\begin{tabular}{p{0.0mm}*{3}{c}}
& \figclip{0 0 5 30}{0.600}{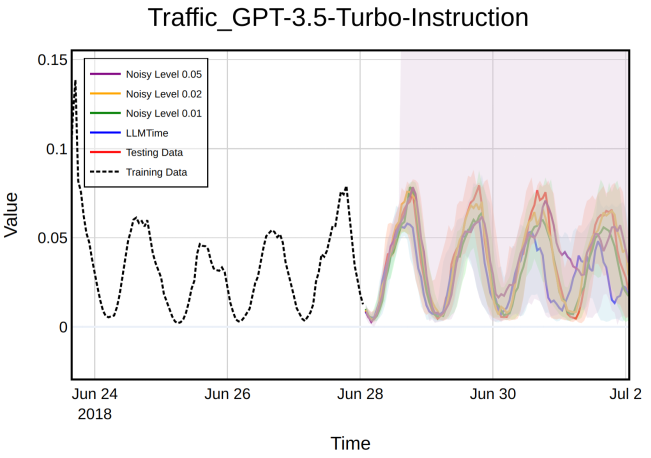} & \figclip{0 0 20 25}{0.60}{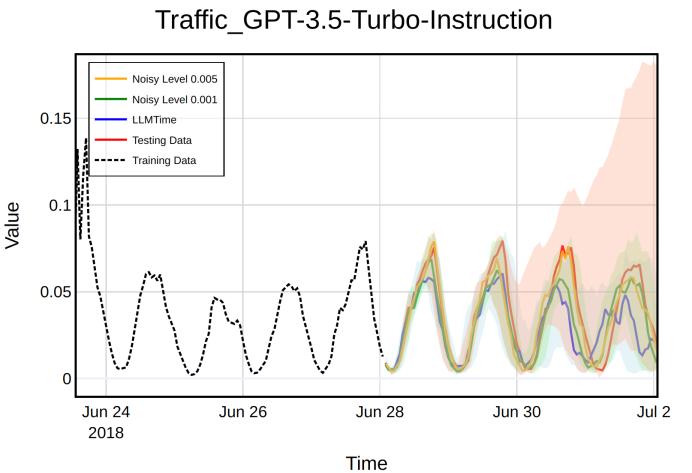} & \figclip{10 10 0 10}{0.60}{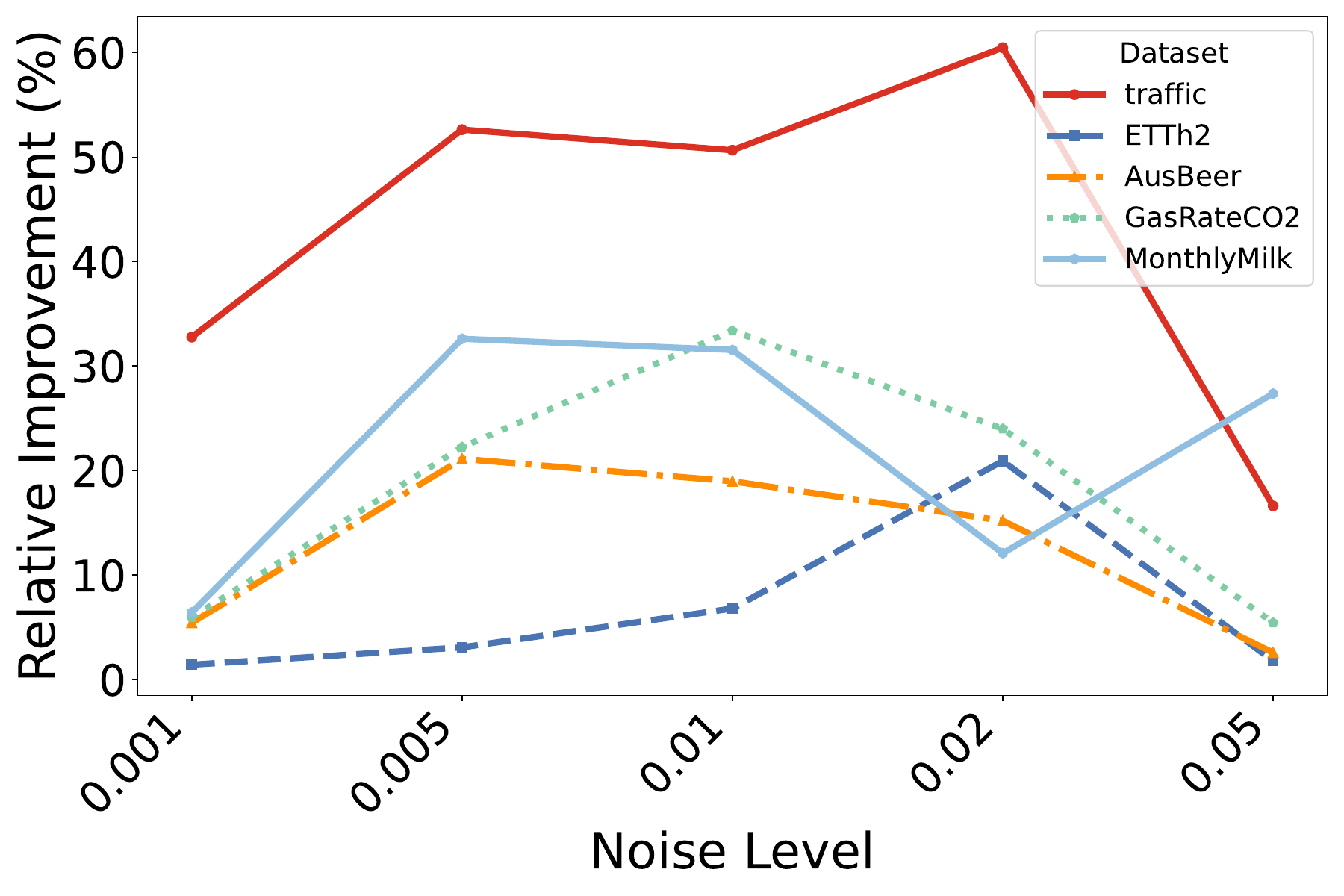} \\
& (a) NLTS with low-level noise ($\alpha<0.01$) & (b) NLTS with high-level noise ($\alpha\geq0.01$) & (c) Relative improvements of different noises
\end{tabular}
\caption{Effect of noise level with NLTS. (a) and (b) illustrate predictions on the Traffic dataset under low and high noise levels. (c) summarizes the relative improvements of NLTS over the LLMTime baseline across multiple datasets under varying noise levels. 
}\label{fig:GPT-Traffic}
\end{figure*}

\subsection{Main results}

\tref{mse-mae-short} and \tref{mse-mae-long} report the zero-shot forecasting results of NLTS on short-term and long-term time series benchmarks, respectively. 
For each benchmark, we consider the best-performing results of LLM-based models enhanced with NLTS, regardless of the noise levels applied. 

NLTS demonstrates a clear advantage in both settings, consistently achieving the lowest MSE and MAE across the vast majority of datasets, and outperforming all baseline methods. For example, in the short-term IstanbulTraffic dataset, NLTS reduces the MAE by approximately 45\% compared to LLMTime; in the more challenging long-term setting (e.g., the ILI dataset), where all methods exhibit high prediction errors, NLTS still achieves the best performance.

Notably, LLMTime, which is also a zero-shot method, shows competitive results on multiple tasks. This suggests that the success of LLMs in zero-shot time series forecasting is not primarily due to memorizing answers from data contamination, but more likely stems from their genuine ability to understand data patterns and make accurate predictions. We further analyze this aspect in Section~\ref{sec:ood}, where we specifically examine forecasting performance without data contamination.

\begin{figure}[tb]
\centering
\figref{1}{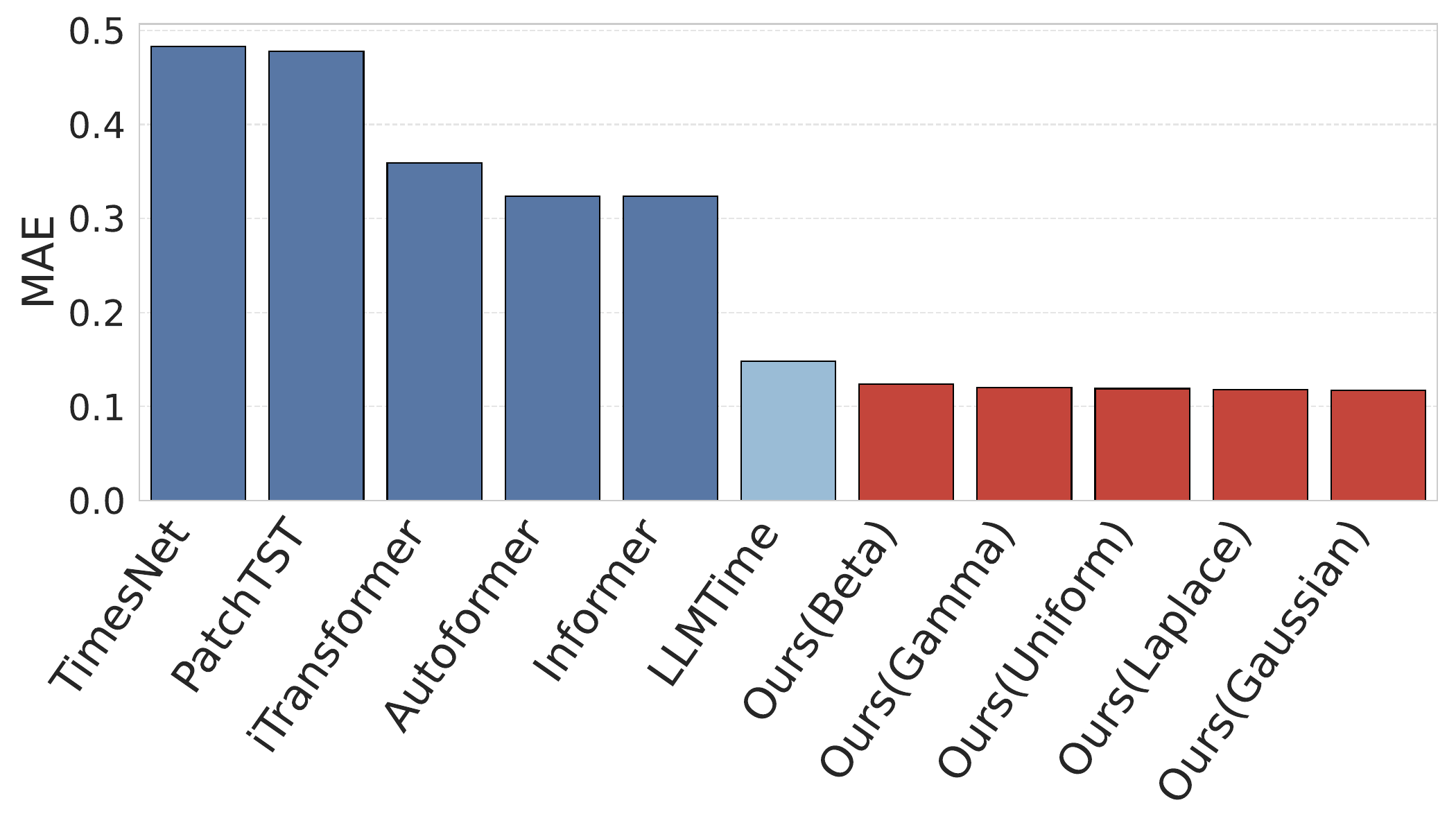}
\caption{Performance of our NLTS on ETTh2 under different noise types.}
\label{fig:noise_types}
\end{figure}


\subsection{Quantitative analysis}

\noindent\textbf{Impact of noise level.}
The scale of noise critically determines the ability of LLMs with NLTS to produce accurate forecasts across diverse domains. In this section, we investigate the effect of noise level in LLMs with NLTS. 
To keep the evaluation focused and informative, we report results on five representative real-world datasets: Traffic, ETTh2, AusBeer, GasRateCO2, and MonthlyMilk. From \fref{GPT-Traffic} (c), we can observe that the forecasting accuracy of GPT-based models with NLTS varies significantly across different noise intensities. Crucially, moderate noise levels, particularly those in the range of 0.005 to 0.02, consistently yield superior and more stable performance. This trend suggests the presence of a noise augmentation "sweet spot", where the benefits of regularization are maximized without introducing excessive distortion.
For instance, in \fref{GPT-Traffic} (a) and (b), the model achieves its highest accuracy (60.48\%) on the Traffic dataset with an MAE improvement at the noise level of 0.02, demonstrating that a carefully calibrated level of perturbation can effectively enhance model generalization in volatile environments. Similar performance trends are also observed in the AusBeer and MonthlyMilk datasets.
In practice, model selection and capacity implicitly interact with noise tolerance: larger models, due to their inherent representation richness, may implicitly adjust to optimal noise levels during inference. This phenomenon points to a potential synergy between model size and noise-aware augmentation strategies, warranting further investigation.

\begin{table}[tb]
\centering
\small
\setlength{\tabcolsep}{1.8pt}
\begin{tabular}{cccccc}
\toprule
Model / Noise Level & 0.001 & 0.005 & 0.010 & 0.020 & 0.050 \\
\midrule
GPT-4                      & 22.82\% & -2.19\% & 11.00\% & \textbf{29.43\%} & 3.17\%  \\
GPT-3.5                    & 1.41\%  & 3.07\%  & 6.79\%  & \textbf{20.91\%} & 1.78\%  \\
Moonshot-V1-8k             & 12.49\% & 9.18\%  & 25.40\% & -3.03\% & 9.76\%  \\
Claude-3-Opus              & 4.69\%  & 13.08\% & 14.23\% & \textbf{26.94\%} & 23.18\% \\
Claude-3.5-Haiku           & 1.67\%  & 4.55\%  & 23.09\% & \textbf{25.88\%} & 22.39\% \\
Claude-3.5-Sonnet          & 62.71\% & \textbf{62.97\%} & 13.72\% & -4.02\% & 15.97\% \\
Deepseek-V3                & 0.09\%  & 1.90\%  & 0.84\%  & \textbf{3.60\%}  & 3.07\%  \\
GLM-4-Air                  & 14.71\% & \textbf{50.09\%} & 41.83\% & 45.05\% & 27.44\% \\
Qwen3-4B                      & 8.75\%  & 12.56\% & 10.69\% & 6.87\%  & \textbf{29.34\%} \\
\bottomrule
\end{tabular}
\caption{Performance improvement (\%) across different LLMs on the ETTh2 dataset achieved through NLTS. \textbf{Bold:} the LLM with the greatest improvement at each noise scale.}
\label{tab:etth2_nlts_gain}
\end{table}

\begin{table*}[t]
\centering
\small
\setlength{\tabcolsep}{1.8pt}
\begin{tabular}{llcccccccccccccc}
\toprule
\multirow{2}{*}{Datasets} & \multirow{2}{*}{Sources} 
& \multicolumn{2}{c}{\shortstack{NLTS 
}} 
& \multicolumn{2}{c}{\shortstack{iTransformer 
}} 
& \multicolumn{2}{c}{\shortstack{LLMTime 
}}
& \multicolumn{2}{c}{\shortstack{TimesNet 
}} 
& \multicolumn{2}{c}{\shortstack{NSTransformer 
}} 
& \multicolumn{2}{c}{\shortstack{Autoformer 
}}
& \multicolumn{2}{c}{\shortstack{ARIMA 
}}\\
\cmidrule(lr){3-4} \cmidrule(lr){5-6} \cmidrule(lr){7-8} \cmidrule(lr){9-10} \cmidrule(lr){11-12} \cmidrule(lr){13-14} \cmidrule(lr){15-16}
& & MSE & MAE & MSE & MAE & MSE & MAE & MSE & MAE & MSE & MAE & MSE & MAE & MSE & MAE \\
\midrule
\multirow{6}{*}{Synthetic TS} 
& ExpSineSquared & \textbf{0.013} & \textbf{0.091} & 1.028 & 0.808 & \underline{0.016} & \underline{0.104} & 0.965 & 0.769 & 0.952 & 0.783 & 0.946 & 0.777 & 0.025 & 0.123\\
& Linear & \textbf{0.018} & \textbf{0.110} & 1.807 & 1.139 & \underline{0.019} & \underline{0.119} & 1.741 & 1.109 & 1.734 & 1.125 & 1.752 & 1.135 & 0.041 & 0.171\\
& Matern & \textbf{0.012} & \textbf{0.083} & 1.020 & 0.754 & \underline{0.012} & \underline{0.089} & 1.013 & 0.775 & 1.004 & 0.771 & 0.979 & 0.746 & 0.022 & 0.114\\
& Polynomial & \textbf{0.012} & \textbf{0.086} & 1.031 & 0.831 & \underline{0.016} & \underline{0.104} & 1.228 & 0.914 & 1.081 & 0.843 & 1.117 & 0.864 & 0.024 & 0.122\\
& RQ & \textbf{0.013} & \textbf{0.086} & 1.190 & 0.883 & \underline{0.014} & \underline{0.096} & 1.222 & 0.892 & 1.106 & 0.857 & 1.168 & 0.870 & 0.029 & 0.140\\
& RBF & \textbf{0.015} & \textbf{0.092} & 1.213 & 0.852 & \underline{0.017} & \underline{0.100} & 1.235 & 0.855 & 1.285 & 0.863 & 1.222 & 0.836 & 0.031 & 0.138\\
\midrule
\multirow{2}{*}{Latest Stock TS} 
& DJIAh   & \textbf{0.0004} & \textbf{0.017} & 0.039 & 0.156 & \underline{0.0006} & \underline{0.020} & 0.034 & 0.142 & 0.025 & 0.132 & 0.137 & 0.329 & 0.0016 & 0.033 \\
& SZ300m  & \textbf{0.00001} & \textbf{0.003} & 0.001 & 0.022 & 0.00017 & 0.011 & 0.001 & 0.018 & 0.001 & 0.028 & 0.092 & 0.292 & \underline{0.00005} & \underline{0.006} \\
\bottomrule
\end{tabular}
\caption{Zero-shot forecasting performance on our synthetic dataset and latest stock market datasets, which are chronologically and substantively disjoint from all evaluated LLMs' training data. \textbf{Bold} and \underline{underline}: the best and the second best performance.}
\label{tab:no-data-contamination}
\end{table*}

\noindent\textbf{Impact of noise types.}
The selection of noise type plays a pivotal role in determining the effectiveness of NLTS. Different types of noise—ranging from simple, unstructured forms like uniform noise to more complex, structured distributions—can exert diverse effects on model dynamics and forecasting performance. 
To assess the influence of noise distribution, we conduct controlled experiments on the ETTh2 dataset by injecting various types of noise into the input. As shown in \fref{noise_types}, among all noise types, Gaussian noise achieves the best performance, with a minimum MAE of 0.120. Moreover, regardless of the noise type used, all noise‑augmented variants of NLTS outperform both traditional forecasting models and the LLMTime baseline.

\noindent\textbf{Impact of LLM choices.}
We evaluate a diverse collection of LLMs, covering both open-source and proprietary models, on the ETTh2 dataset using NLTS, without domain-specific fine-tuning. We further emphasize that our focus is on a low-cost, easily deployable off-the-shelf setup that requires no additional pre-training, aiming to facilitate the adoption of LLMs in resource-constrained environments. 
As shown in \tref{etth2_nlts_gain}, NLTS yields substantial improvements in forecasting accuracy across most models and noise levels, although the extent and consistency of these gains vary by model. For example, Claude-3.5-Sonnet and GLM-4-Air exhibit the most pronounced enhancements at a low noise level ($\alpha=0.005$), with relative improvements of \textbf{62.97\%} and \textbf{50.09\%}, respectively. Claude-3.5-Haiku and Claude-3-Opus display more stable performance across different noise intensities, each peaking above \textbf{25\%} at $\alpha=0.02$. In contrast, GPT-4 and Moonshot-V1-8k present more variability, with fluctuations in performance depending on the noise scale.
The consistent improvements observed across diverse LLMs underscore NLTS's promise as a lightweight and effective method for enhancing robustness in TS forecasting tasks, with no model-specific adaptation.


\section{Zero-Shot Forecasting without Data Contamination}\label{sec:ood}


Considering that millions of information sources are fed into the LLM pre-training and fine-tuning stage, the test data benchmark may be included in the training data of LLM, a phenomenon known as \textbf{data contamination}~\cite{dong2024generalization,xu2024benchmark}. 
Under data contamination, LLMs superficially perform zero-shot prediction, while in reality merely conducting in-domain prediction. 
Consequently, traditional benchmarks tend to overestimate the capabilities of LLM-based methods.
To address this concern, we design two new benchmarks without data contamination: (1) synthetic data generated by various GP kernels, and (2) the latest stock market datasets that are chronologically and substantively disjoint from all evaluated LLMs' training data.

\subsection{Setup}
For synthetic data, we generate TS samples from GPs with various kernel functions, including ExpSineSquared, Linear, Matérn, Polynomial, Rational Quadratic (RQ), and Radial Basis Function (RBF). 
For real-world stock datasets, we choose two time series:
SZ300m (minute-level data from May 6–9, 2025) and DJIAh (hourly data from May 1 to July 31, 2025). More details can be found in ~\Aref{without-data-conta}.

\subsection{Main results}
As shown in~\tref{no-data-contamination}, our method, which is based on GPT-3.5-Turbo-Instruct, consistently outperforms all baselines in two datasets. 
Notably, another LLM-based zero-shot method, LLMTime~\cite{LLMTime}, also achieves competitive performance. 
This result indicates that the success of LLMs as zero-shot TS forecasters is not necessarily due to memorization of answers caused by data contamination, but is likely due to truly understanding the data pattern and making correct predictions.
Moreover, several specialized deep learning methods perform poorly compared to traditional methods such as ARIMA. 
This finding prompts us to reconsider whether conventional TS forecasting benchmarks have become outdated, and designing more representative benchmarks to accurately evaluate the actual predictive capacity of models is an urgent challenge.



\section{Conclusion
}\label{sec:conclude}
In this work, we demonstrated that the zero-shot time series forecasting capability of off-the-shelf large language models can be significantly enhanced through a simple, non-invasive strategy: injecting noise into the input time series prior to tokenization. Unlike methods that require fine-tuning auxiliary modules or model parameters, our approach (NLTS) operates entirely through input-space perturbation, making it both computationally efficient and universally applicable to any pre-trained LLM. Through theoretical analysis, we showed that well-trained LLMs are naturally predisposed to benefit from such input perturbations. Empirically, NLTS consistently outperformed existing methods across diverse benchmarks, with its effectiveness on our newly introduced contamination-free datasets—synthetic data and recent stock prices—confirming that the improvement stems from genuine generalization rather than data memorization. Future work should investigate more adaptive noise injection strategies, precisely tailored to specific model characteristics and dataset properties, to further optimize the performance of LLMs in TS applications.

\FloatBarrier  


\bibliography{aaai2026}

\clearpage

\input{appendix.tex}


\end{document}

%% file: appendix.tex

\appendix

\setcounter{secnumdepth}{2} 

\renewcommand{\thesubsection}{\Alph{section}.\arabic{subsection}}

\onecolumn

\section*{Appendix}


This appendix provides rigorous technical substantiation and extended empirical analysis to support the methodological and experimental claims in the main text. Structured into four thematic sections, it delivers:  
\begin{itemize}
    \item 1. Formal Theoretical Foundations: Complete proofs of Theorem 1 and Lemma 1 (\Aref{Proof of Theorem}), ensuring the mathematical validity of our core propositions.  
\item 2. Methodological Transparency: Detailed exposition of the NLTS framework with noise-augmented prompts (\Aref{NLTS_prompt}), including noise distributions, perturbation mechanics, prompt design paradigms, and algorithmic implementation.  
\item 3. Reproducibility Protocols and Extended Quantitative Evidence : Comprehensive experiment specifications (\Aref{Experiment_Details})—datasets, implementation hyperparameters, and LLM access cost analysis—enabling exact replication. Granular performance analyses addressing multi-LLM benchmarking, horizon sensitivity, noise-level efficacy, and visualizations of forecasting behavior across all evaluated scenarios.
\item 4. Data Contamination Experiments and Extended Results: Describe the experimental design and datasets used to evaluate the impact without data contamination, and present results on additional datasets, including per-dataset performance and average performance visualizations (see \Aref{without-data-conta}).
\end{itemize}

Collectively, these materials fortify the study's academic integrity, offer actionable insights for practitioners, and establish a foundation for further research in time-series augmentation strategies. 

\section{Proof of Theorem}
\label{appendix:Proof of Theorem}
\subsection{Proof of theorem 1}
\begin{proof}
\textbf{Step 1}: Let $ \hat{\mathcal{L}}(\Theta) = \log p_\Theta(\mathcal{S}) $ denote the empirical log-likelihood of an LLM over a training dataset $ \mathcal{S} $, where $ p_\Theta(\mathcal{S}) $ is the probability assigned to $ \mathcal{S} $ under parameters $ \Theta $. 
By definition, $ \Theta^* $ maximizes $ \hat{\mathcal{L}}(\Theta) $ if $ \hat{\mathcal{L}}(\Theta^*) \geq \hat{\mathcal{L}}(\Theta) $ for all $ \Theta $ in a neighborhood of $ \Theta^* $. A necessary condition for this is that the gradient vanishes:  
\begin{align}  
\nabla_\Theta \hat{\mathcal{L}}(\Theta^*) = \mathbb{E}_{S\sim \mathcal{S}} \left[ \nabla_\Theta \log p_\Theta(\mathcal{S}) \right] \bigg|_{\Theta=\Theta^*} = 0.  
\end{align}  
This follows from Fermat's theorem in optimization: extrema of differentiable functions occur at critical points where the gradient is zero.  

\textbf{Step 2}: To ensure $ \Theta^* $ is a local maximum, we examine the second-order Taylor expansion of $ \hat{\mathcal{L}}(\Theta) $ around $ \Theta^* $:  
\begin{align}
\hat{\mathcal{L}}(\Theta^* + \Delta \Theta) = \hat{\mathcal{L}}(\Theta^*) + \Delta \Theta^\top \nabla_\Theta \hat{\mathcal{L}}(\Theta^*) + \frac{1}{2} \Delta \Theta^\top \nabla_\Theta^2 \hat{\mathcal{L}}(\Theta^*) \Delta \Theta + o(\|\Delta \Theta\|^2).  
\end{align}  
Substituting $ \nabla_\Theta \hat{\mathcal{L}}(\Theta^*) = 0 $, the dominant term is the quadratic form $ \frac{1}{2} \Delta \Theta^\top H_{f}(\Theta^*) \Delta \Theta $. For $ \Theta^* $ to be a local maximum, this term must be negative for all $ \Delta \Theta \neq 0 $, which requires $ H_{f}(\Theta^*) \prec 0 $.

\textbf{Step 3}: The empirical Hessian $ H_{f}(\Theta^*) = \nabla_\Theta^2 \hat{\mathcal{L}}(\Theta^*) $ is evaluated on the training set $ \mathcal{S} $. To generalize beyond $ \mathcal{S} $, consider the expected Hessian over the data distribution $ \mathcal{D}_{S} $:  
\begin{align}  
\mathbb{E}[H_{f}(\Theta^*)] = \mathbb{E}_{S\sim \mathcal{D}_{S}} \left[ \nabla_\Theta^2 \log p_\Theta(\mathcal{S}) \right] \bigg|_{\Theta=\Theta^*}.  
\end{align} 
By the law of large numbers, $ H_{f}(\Theta^*) \to \mathbb{E}[H_{f}(\Theta^*)] $ as $ |\mathcal{S}| \to \infty $. Negative definiteness of $ \mathbb{E}[H_{f}(\Theta^*)] $ ensures the curvature remains concave in expectation, preventing overfitting to $ \mathcal{S} $.  

\textbf{Step 4}: The Fisher information matrix $ {F}(\Theta^*) $ is defined as:  
\begin{align}  
{F}(\Theta^*) = \mathbb{E}_{S\sim \mathcal{D}_{S}} \left[ \nabla_\Theta \log p_\Theta(\mathcal{S}) \nabla_\Theta \log p_\Theta(\mathcal{S})^\top \right] \bigg|_{\Theta=\Theta^*}.  
\end{align}  
Under regularity conditions, the expected Hessian satisfies:  
\begin{align}  
\mathbb{E}[H_{f}(\Theta^*)] = -{F}(\Theta^*).  
\end{align}  
Since $ {F}(\Theta^*) \succ 0 $ for identifiable models, $ \mathbb{E}[H_{f}(\Theta^*)] \prec 0 $, confirming negative definiteness.  

\textbf{Step 5}: In over-parameterized LLMs, the Hessian $ H_{f}(\Theta^*) $ has a high-dimensional nullspace but satisfies $ \mathbb{E}[H_{f}(\Theta^*)] \prec 0 $ in non-degenerate directions. This ensures that, despite non-convexity, the model converges to a flatter minimum where the dominant curvature is concave, aligning with empirical observations of robust generalization.  

Therefore, a well-trained LLM satisfies $ \nabla_\Theta \hat{\mathcal{L}}(\Theta^*) = 0 $ and $ \mathbb{E}[H_{f}(\Theta^*)] \prec 0 $. These conditions jointly certify that $ \Theta^* $ is a strict local maximum of the empirical log-likelihood, with stable curvature properties that generalize beyond the training set.  
\end{proof}   

\subsection{Proof of Lemma 1}
\begin{proof}  
Let \( f_{\Theta^*}: \mathcal{X} \rightarrow \mathcal{Y} \) be a well-trained LLM satisfying the first- and second-order optimality conditions in Theorem \ref{thm:optimality}, and let \( h: \mathcal{X} \rightarrow \mathcal{Y} \) be a target function. 

\textbf{Step 1}: Let \( \tilde{x}_t = x + \alpha \epsilon \), where \( \alpha \) scales the noise magnitude. We expand \( f_{\Theta^*}(\tilde{x}_t) \) around \( x \) by using a second-order Taylor series:  
\begin{align}  
f_{\Theta^*}(\tilde{x}_t) = f_{\Theta^*}(x) + \alpha \epsilon^\top \nabla_x f_{\Theta^*}(x) + \frac{\alpha^2}{2} \epsilon^\top {H}_x(f_{\Theta^*}) \epsilon + o(\alpha^2),  
\end{align}  
where \( {H}_x(f_{\Theta^*}) = \nabla_x^2 f_{\Theta^*}(x) \) is the input-space Hessian of the LLM.  

\textbf{Step 2}: We take expectations over \( \epsilon \sim \mathcal{N}(0,\sigma^2) \):  
\begin{align}  
\mathbb{E}_{\epsilon}[f_{\Theta^*}(\tilde{x}_t)] = f_{\Theta^*}(x) + \frac{\alpha^2 \sigma^2}{2} \mathrm{Tr}({H}_x(f_{\Theta^*})) + o(\alpha^2),  
\end{align}  
since \( \mathbb{E}[\epsilon] = 0 \) and \( \mathbb{E}[\epsilon^\top {H}_x(f_{\Theta^*}) \epsilon] = \sigma^2 \mathrm{Tr}({H}_x(f_{\Theta^*}) \).  

\textbf{Step 3}: We subtract \( h(x) \) from both sides:  
\begin{align}  
\mathbb{E}_{\epsilon}[f_{\Theta^*}(\tilde{x}_t) - h(x)] = \underbrace{f_{\Theta^*}(x) - h(x)}_{\text{Baseline Error}} + \frac{\alpha^2 \sigma^2}{2} \mathrm{Tr}({H}_x(f_{\Theta^*})) + o(\alpha^2).  
\end{align}  
The inequality \( \mathbb{E}_{\epsilon}[f_{\Theta^*}(\tilde{x}_t) - h(x)] \leq \mathbb{E}_{\epsilon}[f_{\Theta^*}(x) - h(x)] \) reduces to:  
\begin{align}  
\frac{\alpha^2 \sigma^2}{2} \mathrm{Tr}({H}_x(f_{\Theta^*})) \leq 0.  
\end{align}  

\textbf{Step 4}: 
From Theorem \ref{thm:optimality}, the parameter-space Hessian \( \mathbb{E}[{H}_{f}(\Theta^*)] = \mathbb{E}[\nabla_\Theta^2 \hat{\mathcal{L}}(\Theta^*)] \prec 0 \). By the chain rule, the input-space Hessian relates to the parameter-space Hessian via:  
\begin{align}  
{H}_x(f_{\Theta^*}) = {J}_x(\Theta^*)^\top \mathbb{E}[{H}_{f}(\Theta^*)] {J}_x(\Theta^*),  
\end{align}  
where \( {J}_x(\Theta^*) = \nabla_\Theta f_{\Theta^*}(x) \) is the Jacobian of \( f_{\Theta^*} \) with respect to \( \Theta \). Since \( \mathbb{E}[{H}_{f}(\Theta^*)] \prec 0 \), it follows that \( {H}_x(f_{\Theta^*}) \preceq 0 \), and thus \( \mathrm{Tr}({H}_x(f_{\Theta^*})) \leq 0 \).  

Therefore, the inequality demonstrates that input perturbations reduce the expected error when the LLM is well-trained. This aligns with \thref{optimality}'s second-order condition, which enforces flatness in the loss landscape, making the model resilient to input variations.  
The lemma formalizes how noise injection regularizes LLM predictions by exploiting the concave curvature guaranteed by \thref{optimality}. 

\end{proof}

\section{Design of NLTS with Noise-Augmented Prompts}
\label{appendix:NLTS_prompt}

\subsection{Noise distribution}\label{appendix:noise_distribution}
The Gaussian distribution is defined by two parameters: the mean $ \mu $ and the variance $ \sigma^2 $. The probability density function (PDF) of the Gaussian distribution is given by:
\begin{align}
    f(\epsilon_{i}|\mu,\sigma) = \frac{1}{\sigma\sqrt{2\pi}} e^{-\frac{(\epsilon_{i}-\mu)^2}{2\sigma^2}}.
\end{align}
The Uniform distribution on the interval $[a, b]$ is a continuous distribution where every value within the interval is equally likely. The PDF is:
\begin{align}
    f(\epsilon_{i}|a,b) = \frac{1}{b-a} \quad \text{for } a \leq \epsilon_{i} \leq b.
\end{align}
The uniform distribution serves as a reference for other distributions and is often used in simulations and modeling when no prior information about the distribution of the noise is available.
The Laplace distribution, also known as the double exponential distribution, has a sharp peak at its mean and heavier tails compared to the normal distribution. The PDF of a Laplace distributed random variable $ \epsilon_{i} $ with location parameter $ \mu $ and scale parameter $ b $ is:
\begin{align}
    f(\epsilon_{i}|\mu,b) = \frac{1}{2b} e^{-\frac{|\epsilon_{i}-\mu|}{b}}.
\end{align}
The Gamma distribution is a two-parameter family of continuous probability distributions defined by shape parameter $ k $ and scale parameter $ \theta $. The PDF of a Gamma distributed random variable $ \epsilon_{i} $ is:
\begin{align}
    f(\epsilon_{i}|k,\theta) = \frac{\epsilon_{i}^{k-1}e^{-\frac{\epsilon_{i}}{\theta}}}{\theta^k \Gamma(k)} \quad \text{for } \epsilon_{i} > 0.
\end{align}
The Beta distribution is defined on the interval $[0, 1]$ and is characterized by two shape parameters, $ \alpha $ and $ \beta $. The PDF of a beta-distributed random variable $ \epsilon_{i} $ is:
\begin{align}
    f(\epsilon_{i}|\alpha,\beta) = \frac{\epsilon_{i}^{\alpha-1}(1-\epsilon_{i})^{\beta-1}}{B(\alpha,\beta)}.
\end{align}
where $ B(\alpha,\beta) $ is the beta function. 

\subsection{Distribution of data perturbation with noise injection}

In this section, we evaluate the impact of noise injection on the distributional alignment of LLM-based time series forecasting methods across two benchmark datasets: Traffic and ETTh2. Input data distributions are analyzed before and after noise injection (Figures 6a, 6d), revealing how stochastic perturbations modify the training data while preserving underlying temporal patterns. The core comparison involves two methodologies: LLMTime (a baseline without noise injection) and NLTS. Test set outputs are evaluated across three LLM architectures—GPT-3.5-Turbo-Instruct, GLM-Air, and Claude-3.5-Sonnet—to assess generalization.  

Results demonstrate that NLTS outputs exhibit significantly closer alignment with the true data distribution compared to LLMTime (Figures 6 b-c, 6e-f). This alignment is consistent across both datasets and all three LLMs, indicating that noise injection enhances robustness by reducing overfitting to training idiosyncrasies. The improved distributional fidelity suggests that NLTS mitigates the domain gap between training and testing phases, enabling better adaptation to real-world variability. The experiment underscores the critical role of structured noise in stabilizing LLM-based forecasts and highlights the framework's capacity to generalize across heterogeneous architectures and datasets.

\begin{figure*}[htbp]
\renewcommand{\tabcolsep}{-0.5mm}
\scriptsize
\centering
\begin{tabular}{p{0.0mm}*{3}{c}}
& \figref{0.33}{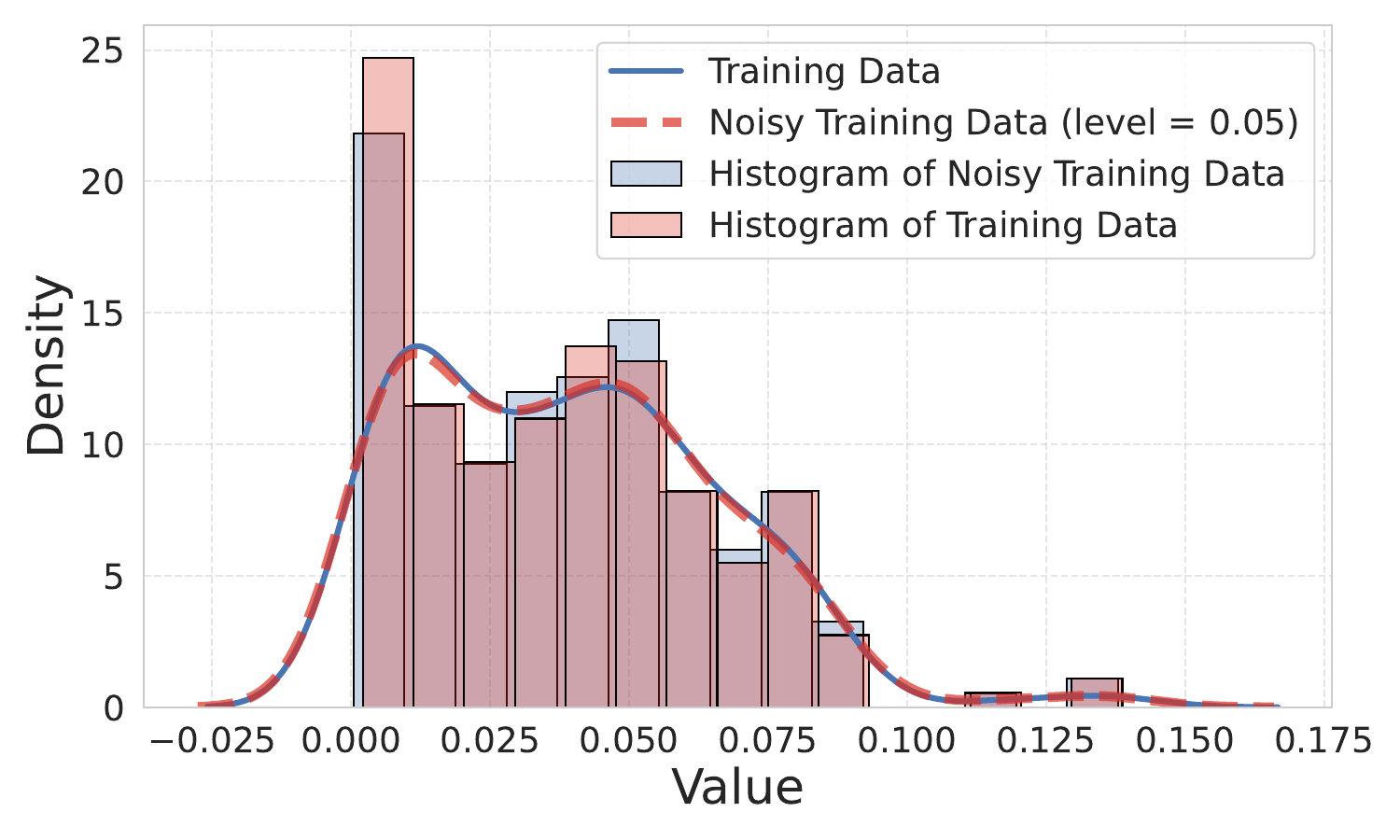} 
& \figref{0.33}{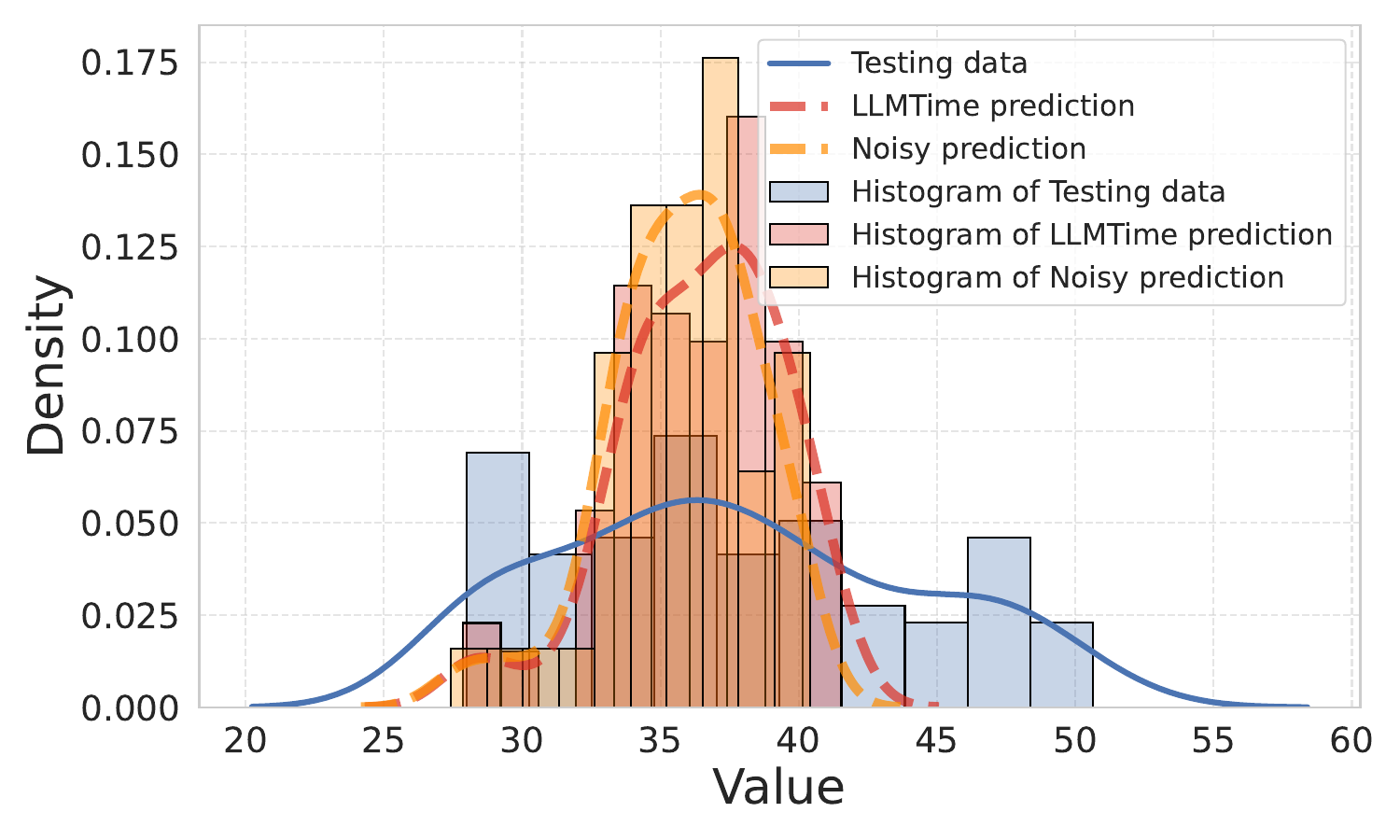} 
& \figref{0.33}{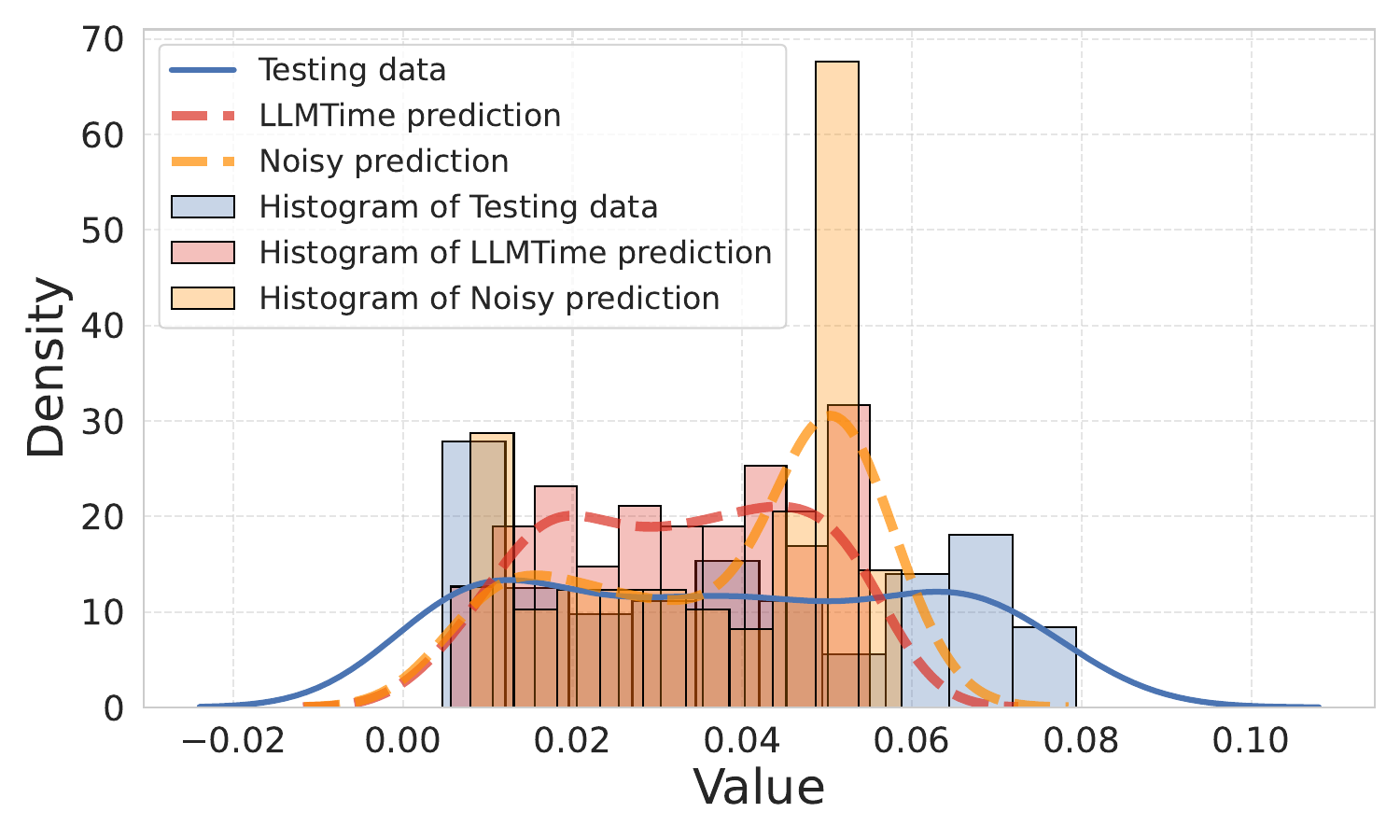} \\
& (a) Input distribution (Traffic) 
& (b) GPT-3.5: LLMTime vs NLTS (Traffic) 
& (c) GLM-Air: LLMTime vs NLTS (Traffic) \\
& \figref{0.33}{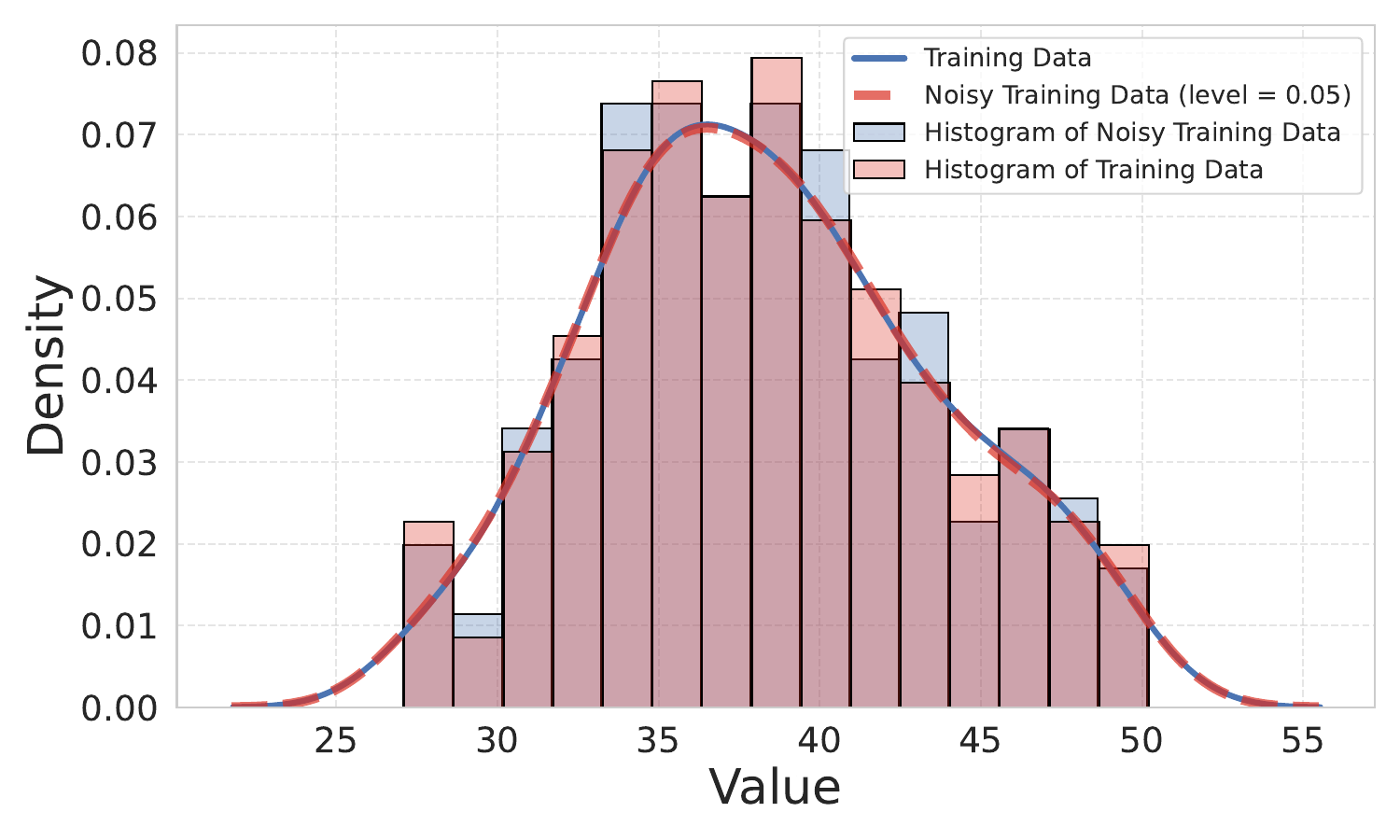} 
& \figref{0.33}{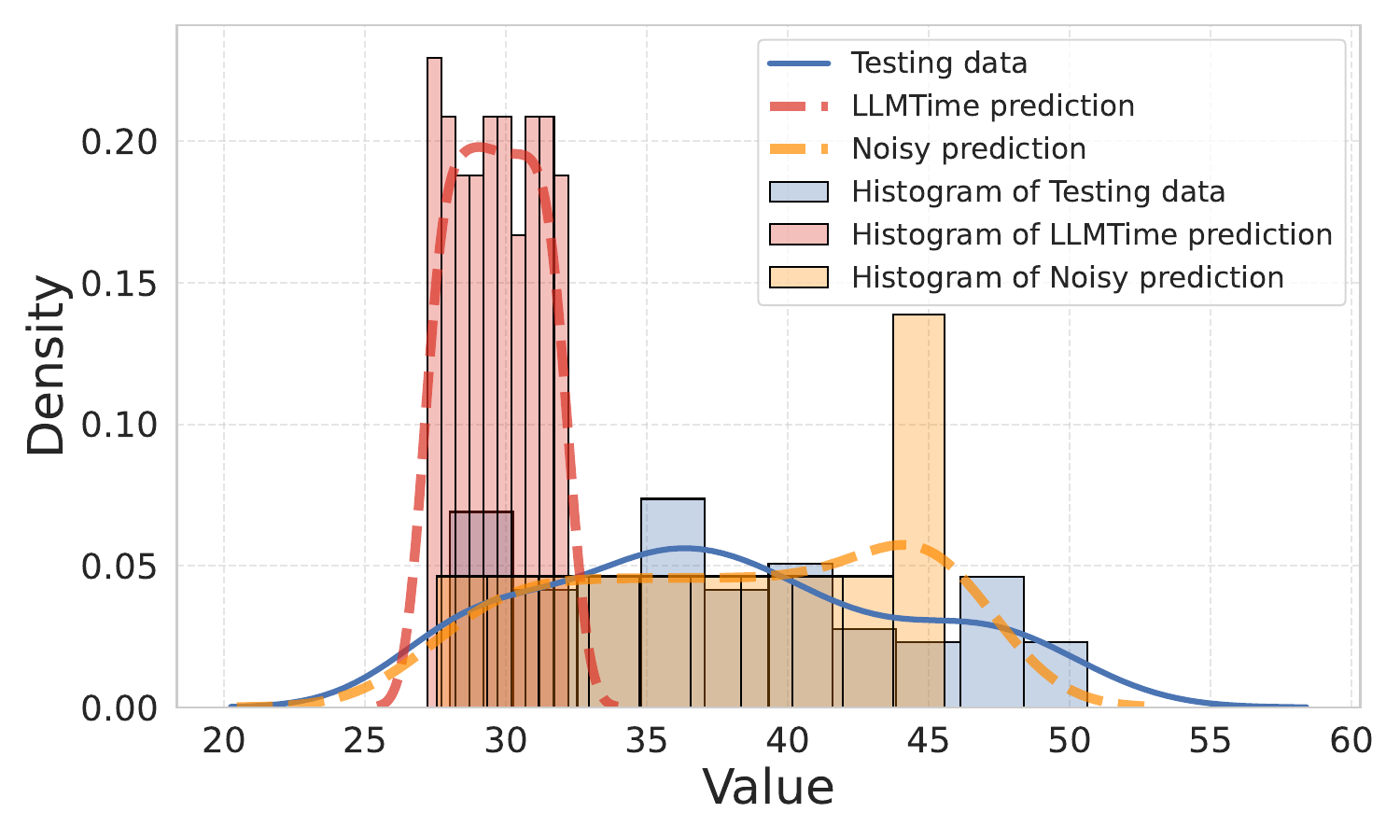} 
& \figref{0.33}{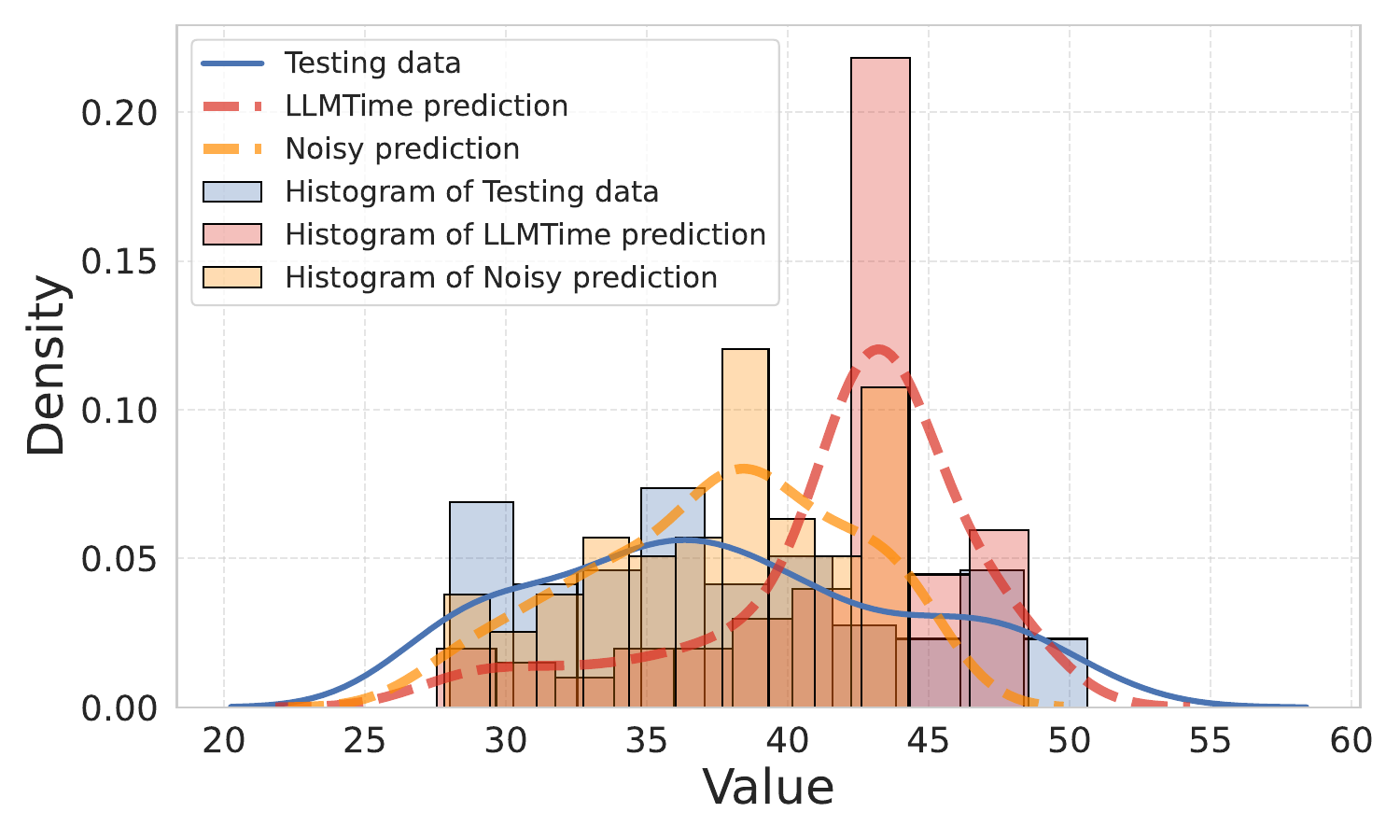} \\
& (d) Input distribution (ETTh2) 
& (e) Claude: LLMTime vs NLTS (ETTh2) 
& (f) GLM-Air: LLMTime vs NLTS (ETTh2)
\end{tabular}
\vspace{-3mm}
\caption{Distribution comparison on the \textbf{Traffic} and \textbf{ETTh2} dataset. Figures (a) and (d) show the empirical distributions of the training data before and after noise injection. Figures (b)-(c) illustrate test set output distributions on Traffic using LLMTime (without noise) and NLTS (with noise), across GPT-3.5-Turbo-Instruct and GLM-Air. Figures (e)-(f) show results on ETTh2 across Claude-3.5-Sonnet and GLM-Air.
In all cases, NLTS outputs align more closely with the true data distribution, highlighting the robustness improvements from noise injection.
}
\label{fig:distribution}
\vspace{-4mm}
\end{figure*}

\subsection{Prompt example for TS forecasting}

In this experiment, the model's task is to perform time series forecasting based on a sequence of historical observations and generate future predictions. 
Using the MonthlyMilk dataset as an example, we employ two different prompt strategies. 

The \textbf{first strategy} is a \textit{raw numeric sequence prompt}, in which historical observations undergo textualization, and then directly provided to the model as the prompt for forecasting. In its noisy variant, controlled levels of noise ($\alpha$) are first injected into the original numerical observations, which are then textualized in the same manner and provided to the model as the prompt.

The \textbf{second strategy} is a \textit{structured chat prompt}, where the input is formatted as a dialogue between the system and the user. The system message defines the forecasting task and constraints, while the user message provides the historical sequence and requests the model to continue it. As with the numeric prompts, noisy variants are also used.

\begin{pythoncodebox}[Raw numeric sequence prompt]
# Raw numeric sequence prompt:
"5 1 8 , 4 7 3 , 6 0 0 , 6 2 6 , 7 4 1 , 6 9 2 , 6 0 0 , 5 3 4 , 4 8 4 , 4 9 9 , 4 6 0 , 5 0 7 , 5 3 6 , 4 8 1 , 6 2 1 , 6 5 4 , 7 6 5 , 7 2 3 , 6 3 3 , 5 6 3 , 5 0 9 , 5 1 5 , 4 8 0 , 5 3 3 ..."

# Noisy prompt with noise level $\alpha = 0.005$:
"5 2 0 , 4 7 1 , 5 9 9 , 6 2 6 , 7 4 0 , 6 9 1 , 6 0 3 , 5 3 5 , 4 8 3 , 4 9 8 , 4 6 1 , 5 0 5 , 5 3 4 , 4 7 9 , 6 2 1 , 6 5 7 , 7 6 7 , 7 2 6 , 6 3 4 , 5 6 2 , 5 0 8 , 5 1 7 , 4 7 7 , 5 3 2 ..."

# Noisy prompt with noise level $\alpha = 0.05$:
"5 2 7 , 4 8 4 , 6 1 6 , 6 3 4 , 7 4 3 , 6 9 0 , 6 0 8 , 5 2 8 , 4 9 8 , 5 0 6 , 4 5 9 , 5 1 0 , 5 3 8 , 4 8 4 , 6 1 7 , 6 7 0 , 7 7 0 , 7 2 8 , 6 3 3 , 5 6 9 , 5 1 1 , 5 2 6 , 4 7 6 , 5 2 5 ..."
\end{pythoncodebox}

\begin{pythoncodebox}[Structured chat prompt]
# Structured Chat Prompt Example
[
  {
    "role": "system",
    "content": "You are a helpful assistant specialized in time series forecasting. The user provides a comma-separated sequence of decimal numbers, and you will predict the following values."
  },
  {
    "role": "user",
    "content": "Please continue the sequence without any additional text or explanation. Only output the predicted numbers.\nSequence:\n5 1 8 , 4 7 3 , 6 0 0 , 6 2 6 , 7 4 1 , 6 9 2 , 6 0 0 , 5 3 4 , 4 8 4 , 4 9 9 , 4 6 0 , 5 0 7 , 5 3 6 , 4 8 1 , 6 2 1 , 6 5 4 , 7 6 5 , 7 2 3 , 6 3 3 , 5 6 3 , 5 0 9 , 5 1 5 , 4 8 0 , 5 3 3 ..."
  }
]

# Corresponding noisy input example
[
  {
    "role": "user",
    "content": "Please continue the sequence without any additional text or explanation. Only output the predicted numbers.\nSequence:\n5 2 0 , 4 7 1 , 5 9 9 , 6 2 6 , 7 4 0 , 6 9 1 , 6 0 3 , 5 3 5 , 4 8 3 , 4 9 8 , 4 6 1 , 5 0 5 , 5 3 4 , 4 7 9 , 6 2 1 , 6 5 7 , 7 6 7 , 7 2 6 , 6 3 4 , 5 6 2 , 5 0 8 , 5 1 7 , 4 7 7 , 5 3 2 ..."
  }
]
\end{pythoncodebox}

\subsection{Algorithm of NLTS}
In \aref{NLTS}, we establish a holistic methodology for improving TS forecasting performance using LLMs. 

\begin{algorithm}
\caption{NLTS for TS forecasting}
\label{algo:NLTS}
\begin{algorithmic}[1]
\Require A TS $ \{ x_t \}_{t=1}^{T} $, scaling factor $\alpha$, tokenization function $\mathcal{Q}$;
\Ensure $\bar{\vx}_{*, i}, \text{Var}(\bar{\vx}_{*, i})$;
\State Noise design and sampling;
\State Perturbation of TS by noise injection;
\State Textualizing the points within TS;
\State Tokenize the noised TS in \eref{token};
\State Provide noisy Tokens as prompts to LLM;
\State Predict tokens for testing points;
\State Invert 
predicted Tokens in \eref{reverse};
\State Calculate $\bar{\vx}_{*, i}, \text{Var}(\bar{\vx}_{*, i})$ to aggregate outputs from LLM.
\end{algorithmic}
\end{algorithm}

\section{Experiment Details and Visualizations}
\label{appendix:Experiment_Details}
\subsection{Dataset}\label{appendix:dataset}

\paragraph{Darts} 
This dataset comprises eight real univariate time series, including \textbf{AirPassengers, AusBeer, GasRateCO2, MonthlyMilk, Sunspots, Wine, Wooly, and HeartRate}. 
These time series are relatively short, each containing \textbf{no more than 800} observations.
The first 80\% of each time series is used \textbf{as prompt input} for the LLMs to generate forecasts, while the last 20\% serves as the test set for evaluation. 
We assess the performance of several methods from the Darts package, including neural network models such as Temporal Convolutional Networks (TCN) 
, N-BEATS 
, and N-HiTS 
, as well as traditional statistical approaches like ARIMA
, and the Spectral Mixture Gaussian Process (SM-GP), a Bayesian nonparametric method. Darts serves as an invaluable tool for benchmarking time series forecasting models, making it particularly well-suited for evaluating the performance of LLMTime.

\paragraph{Memorization} 
This dataset includes three time series sourced from Kaggle: \textbf{Istanbul Traffic, TSMC Stock, and Turkey Power.} 
The Istanbul Traffic time series provides minute-by-minute traffic index data for Istanbul from October 2022 to May 2023. We selected the "TI" column and downsampled the data to an hourly frequency for the period from May 5 to May 18, 2023, resulting in \textbf{267 observations}.
The TSMC Stock contains daily stock market trading data for Taiwan Semiconductor Manufacturing Company (TSMC) for 2022. We used the closing price column, which consists of \textbf{246 observations}. 
The Turkey Power includes hourly electricity generation and consumption data for Turkey from January 1, 2020, to December 31, 2022. We selected the "Total" column and downsampled the data for 2022 to a daily frequency, resulting in \textbf{366 observations}.
These time series are short, each containing \textbf{no more than 400} observations. The last 30 observations from each time series are reserved for testing. 
We evaluate the Memorization dataset using the same models as those applied to the Darts dataset. 

\paragraph{Autoformer} 
This dataset 
consists of nine widely used multivariate TS benchmarks. For our experiments, we select \textbf{ETTh1, ETTh2, ETTm1, ETTm2, Electricity(ECL), Traffic }
\textbf{and ILI.}
These time series are relatively long, with the shortest dataset containing \textbf{more than 950} observations and \textbf{most ranging from 10,000 to 70,000 observations.}
For a more manageable evaluation with LLMTime, we use smaller subsets from each time series, specifically selecting the last univariate series, "OT", and the final 96 and 192 time steps from each series for testing. 
The models are evaluated using the same model settings as in LLMTime, with additional comparisons against more advanced state‑of‑the‑art TS forecasting models, including Informer
,  Autoformer, NSTransformer
, TimesNet
, PatchTST
, and iTransformer

\subsection{Implementation}
\label{hyperparameters}
We utilize the GPyTorch library to implement Gaussian processes (GPs), and the Darts library for modeling with ARIMA, TCN, N-BEATS, and N-HiTS. For any hyperparameters not detailed below, we use the default settings. 
For all LLMs and non-LLM models, we consider the following model settings:
\begin{itemize}
\item {GPT-3.5}:  $\alpha = 0.95$, temperature = 0.7, $\beta = 0.3$, basic = False, and precision = 3, with serializer settings: base = 10, signed = True, and half bin correction = True.

\item {Gemini, Claude, GLM, Qwen, Deepseek}: $\alpha = 0.95$, temperature = 1.0, top P = 0.8, basic = True, precision = 3.

\item {Spectral Mixture Gaussian Process (SM-GP):} We use a Gaussian Process with a kernel consisting of a spectral mixture kernel with 12 components and an RBF kernel. The learning rate is tuned from $\{5e-3, 1e-2, 5e-2, 1e-1\}$.

\item {ARIMA:} We perform a grid search over $p \in \{12, 20, 30\}$, $d \in \{1, 2\}$, and $q \in \{0, 1, 2\}$.

\item {TCN:} We perform a grid search over $\text{input\_chunk\_length} \in \{10, 100, 400\}$, $\text{output\_chunk\_length} \in \{1, 10\}$, $\text{kernel\_size} \in \{3, 5\}$, $\text{num\_filters} \in \{1, 3\}$, and likelihood $\in \{\text{Laplace}, \text{Gaussian}\}$.

\item {N-BEATS:} We perform a grid search over $\text{input\_chunk\_length} \in \{10, 100, 400\}$, $\text{output\_chunk\_length} \in \{1, 10\}$, $\text{layer\_widths} \in \{64, 16\}$, $\text{num\_layers} \in \{1, 2\}$, and likelihood $\in \{\text{Laplace}, \text{Gaussian}\}$.

\item {N-HiTS:} We perform a grid search over $\text{input\_chunk\_length} \in \{10, 100, 400\}$, $\text{output\_chunk\_length} \in \{1, 10\}$, $\text{layer\_widths} \in \{64, 16\}$, $\text{num\_layers} \in \{1, 2\}$, and likelihood $\in \{\text{Laplace}, \text{Gaussian}\}$.
\end{itemize}

Note that we do not use the latest or most advanced LLMs, especially higher versions of ChatGPT, due to the cost issues. The expense of advanced ChatGPT models could be many times that of the standard version. On the other hand, the LLMs currently used are quite popular. These models 
have considerable representativeness and diversity. Furthermore, the focus of the study is to explore whether adding noise to TS can enhance the forecasting ability of LLMs. We believe that the findings obtained on existing LLMs can provide valuable references and insights for the optimization of more advanced LLMs in the future. 



\subsection{Access cost of different LLMs}
In \tref{ModelPricing}, we present the access cost of different LLMs used for both prompt and completion tasks. Each model is listed with the associated costs for processing 1,000 tokens in prompt and completion tasks. 
We provide a clear comparison of the token pricing for both types of tasks, ranging from as low as \$ 0.00021 per 1 K tokens to as high as \$ 0.16 per 1 K tokens.   
It is important to acknowledge that the table presents a single snapshot of historical pricing. Indeed, the pricing of LLMs is subject to flux, largely influenced by the competitive dynamics within the commercial landscape of LLM providers.

The cheapest models for prompts are the Qwen3-4B. The most expensive model for prompts and completion tokens is Claude-3.5-Opus, costing $0.032$ and $0.16$ for 1,000 tokens. 
When choosing a model, it's important to consider both the cost of prompts and completions. Some models are cheaper for prompts but not for completions, and vice versa. Practitioners should weigh the costs against the performance benefits to decide which model is best for their needs, especially when dealing with high volumes of data.

\begin{table}[hbt]
\centering
\begin{tabular}{l|c|l|c|l}
\toprule
{LLMs} & {Number of prompt tokens} & {Prompt price} & {Number of completion tokens} & {Completion price} \\
\midrule
GPT-4 & 1 K & \$ 0.03 & 1 K & \$ 0.06 \\
GPT-3.5-Turbo-Instruct & 1 K & \$ 0.0015 & 1 K & \$ 0.002 \\
Claude-3-Opus & 1 K & \$ 0.032 & 1 K & \$ 0.16 \\
Claude-3.5-Haiku & 1 K & \$ 0.00233 & 1 K & \$ 0.01167 \\
Claude-3.5-Sonnet & 1 K & \$ 0.007 & 1 K & \$ 0.035 \\
Deepseek-V3 & 1 K & \$ 0.0005 & 1 K & \$ 0.002 \\
GLM-4-Air & 1 K & \$ 0.0005 & 1 K & \$ 0.0005 \\
Qwen3-4B & 1 K & \$ 0.00021 & 1 K & \$ 0.00084 \\
\bottomrule
\end{tabular}
\caption{Prices of LLMs for prompt and completion tasks.}
\label{tab:ModelPricing}
\end{table}

\subsection{Performance of diverse LLMs.}

Tables~\ref{tab:llm-mse-mae} report the performance of nine different LLMs, including GPT-4, GPT-3.5-Turbo-Instruct, Moonshot-V1-8k, Claude-3-Opus, Claude-3.5-Sonnet, Claude-3.5-Haiku, DeepSeek-V3, GLM-4-Air, and Qwen3-4B, under varying noise levels on the ETTh2 dataset. Due to space constraints, abbreviated model names are used in the tables. 
Across most noise levels, all evaluated LLMs exhibit a certain degree of performance improvement compared to the Original condition.
Most LLMs exhibit more noticeable performance improvements under low-noise conditions. For example, Claude-3.5-Sonnet and GLM-4-Air achieve the lowest errors at noise levels such as 0.001–0.005 compared to the original setting. This indicates that injecting noise with an appropriate intensity can effectively enhance the predictive performance. 
The sensitivity to noise varies significantly across models. Some maintain stable performance as the noise level increases, while others experience noticeable fluctuations. This disparity may stem from our Off-the-Shelf setting without any fine-tuning or additional pretraining, with the outcome largely influenced by the model’s architecture, scale, and the diversity of its pretraining data.

\begin{table}[H]
\centering
\small
\setlength{\tabcolsep}{1.7pt}
\begin{tabular}{lcccccccccccccccccc}
\toprule
\multirow{2}{*}{Noise Level} 
& \multicolumn{2}{c}{GPT-4} 
& \multicolumn{2}{c}{GPT-3.5} 
& \multicolumn{2}{c}{Moonshot} 
& \multicolumn{2}{c}{Claude-Opus} 
& \multicolumn{2}{c}{Claude-Sonnet} 
& \multicolumn{2}{c}{Claude-Haiku} 
& \multicolumn{2}{c}{Deepseek} 
& \multicolumn{2}{c}{GLM} 
& \multicolumn{2}{c}{Qwen} \\
\cmidrule(lr){2-3} \cmidrule(lr){4-5} \cmidrule(lr){6-7} \cmidrule(lr){8-9}
\cmidrule(lr){10-11} \cmidrule(lr){12-13} \cmidrule(lr){14-15} \cmidrule(lr){16-17} \cmidrule(lr){18-19}
& MSE & MAE & MSE & MAE & MSE & MAE & MSE & MAE & MSE & MAE & MSE & MAE & MSE & MAE & MSE & MAE & MSE & MAE \\
\midrule
Original & 0.064 & 0.186 & 0.042 & 0.148 & 0.021 & 0.114 & 0.012 & 0.084 & 0.044 & 0.176 & 0.0169 & 0.095 & 0.0064 & 0.0648 & 0.024 & 0.128 & 0.273 & 0.355 \\
0.001    & 0.033 & 0.144 & 0.043 & 0.146 & 0.015 & 0.099 & 0.011 & 0.080 & 0.007 & \textbf{0.065} & 0.0162 & 0.093 & 0.0064 & 0.0648 & 0.016 & 0.109 & 0.264 & 0.324 \\
0.005    & 0.054 & 0.190 & 0.039 & 0.144 & 0.017 & 0.103 & 0.010 & 0.073 & \textbf{0.006} & \textbf{0.065} & 0.0148 & 0.091 & 0.0063 & 0.0636 & \textbf{0.007} & \textbf{0.064} & 0.260 & 0.311 \\
0.01     & 0.047 & 0.166 & 0.039 & 0.138 & \textbf{0.013} & \textbf{0.085} & 0.008 & 0.072 & 0.035 & 0.151 & 0.0094 & 0.073 & 0.0064 & 0.0643 & 0.009 & 0.075 & \textbf{0.258} & 0.317 \\
0.02     & \textbf{0.028} & \textbf{0.132} & \textbf{0.024} & \textbf{0.117} & 0.023 & 0.117 & \textbf{0.007} & \textbf{0.061} & 0.050 & 0.183 & \textbf{0.0091} & \textbf{0.070} & \textbf{0.0061} & \textbf{0.0625} & 0.008 & 0.071 & 0.265 & 0.331 \\
0.05     & 0.060 & 0.180 & 0.037 & 0.146 & 0.015 & 0.103 & 0.008 & 0.064 & 0.033 & 0.147 & 0.0093 & 0.074 & 0.0063 & 0.0628 & 0.017 & 0.093 & 0.084 & \textbf{0.251} \\
\bottomrule
\end{tabular}
\caption{MSE and MAE of LLMs on ETTh2 under varying noise levels. “Original” refers to results obtained using LLMTime without noise injection for the same LLM. \textbf{Bold}: the best result for each model across different noise levels.}
\label{tab:llm-mse-mae}
\end{table}

\subsection{Performances of different forecasting horizons}
\label{appendix:forecasting_horizons}
We comprehensively evaluate the performance of several forecasting models across two forecasting horizons, 96 and 192, following the experimental settings established by LLMTime, using seven widely adopted benchmark datasets. 
As shown in \tref{performance_autoformer_mse} and \tref{performance_autoformer_mae}, our model consistently outperforms all baseline models at both horizons, demonstrating strong adaptability and generalization in capturing temporal features.




\begin{table*}[htbp]
\centering
\setlength{\tabcolsep}{3.5pt} 
\begin{tabular}{lcccccccccccccc}
\toprule
\multirow{2}{*}{Model} & \multicolumn{2}{c}{ECL} & \multicolumn{2}{c}{ETTh1} & \multicolumn{2}{c}{ETTh2} & \multicolumn{2}{c}{ETTm1} & \multicolumn{2}{c}{ETTm2} & \multicolumn{2}{c}{Traffic} & \multicolumn{2}{c}{ILI} \\
\cmidrule(lr){2-3} \cmidrule(lr){4-5} \cmidrule(lr){6-7} \cmidrule(lr){8-9} \cmidrule(lr){10-11} \cmidrule(lr){12-13} \cmidrule(lr){14-15}
 & 96 & 192 & 96 & 192 & 96 & 192 & 96 & 192 & 96 & 192 & 96 & 192 & 24 & 36 \\
\midrule
Informer     & 0.124 & 0.195 & 0.039 & 0.149 & 0.174 & 0.363 & 0.004 & 0.019 & 0.075 & 0.160 & 0.079 & 0.171 & 11.047 & 9.579 \\
Autoformer   & 0.235 & 0.534 & 0.071 & 0.073 & 0.184 & 0.210 & 0.009 & 0.102 & 0.077 & 0.449 & 0.073 & 0.560 & 1.077  & 0.662 \\
TimesNet     & 0.076 & 0.246 & 0.080 & 0.058 & 0.314 & 0.220 & 0.008 & \underline{0.025} & \underline{0.021} & 0.361 & 0.077 & 0.383 & 0.766  & 0.700 \\
PatchTST     & 0.124 & 0.303 & 0.071 & 0.059 & 0.331 & 0.264 & 0.006 & 0.037 & 0.026 & 0.403 & 0.075 & 0.439 & 0.979  & 1.048 \\
LLMTime      & \underline{0.027} & \underline{0.028} & \underline{0.008} & \underline{0.030} & \underline{0.042} & \underline{0.026} & 0.004 & 0.019 & 0.028 & \underline{0.083} & \underline{0.010} & \underline{0.019} & \underline{0.084}  & \underline{0.014} \\
iTransformer & 0.084 & 0.252 & 0.038 & 0.073 & 0.214 & 0.264 & \underline{0.003} & 0.031 & \underline{0.021} & 0.385 & 0.091 & 0.403 & 0.426  & 0.868 \\
Ours         & \textbf{0.014} & \textbf{0.018} & \textbf{0.005} & \textbf{0.016} & \textbf{0.024} & \textbf{0.018} & \textbf{0.002} & \textbf{0.009} & \textbf{0.020} & \textbf{0.051} & \textbf{0.001} & \textbf{0.007} & \textbf{0.081}  & \textbf{0.007} \\
\bottomrule
\end{tabular}
\caption{MSE of different models across multiple datasets and forecasting horizons. \textbf{Bold}: the best result under each forecasting horizon, \underline{Underline}: the second best}
\label{tab:performance_autoformer_mse}
\end{table*}

\begin{table*}[htbp]
\centering
\setlength{\tabcolsep}{3.5pt} 
\begin{tabular}{lcccccccccccccc}
\toprule
\multirow{2}{*}{Model} & \multicolumn{2}{c}{ECL} & \multicolumn{2}{c}{ETTh1} & \multicolumn{2}{c}{ETTh2} & \multicolumn{2}{c}{ETTm1} & \multicolumn{2}{c}{ETTm2} & \multicolumn{2}{c}{Traffic} & \multicolumn{2}{c}{ILI} \\
\cmidrule(lr){2-3} \cmidrule(lr){4-5} \cmidrule(lr){6-7} \cmidrule(lr){8-9} \cmidrule(lr){10-11} \cmidrule(lr){12-13} \cmidrule(lr){14-15}
 & 96 & 192 & 96 & 192 & 96 & 192 & 96 & 192 & 96 & 192 & 96 & 192 & 24 & 36 \\
\midrule
Informer & 0.291 & 0.349 & 0.164 & 0.329 & 0.324 & 0.503 & 0.054 & 0.109 & 0.250 & 0.354 & 0.221 & 0.338 & 3.312 & 3.065 \\
Autoformer & 0.364 & 0.535 & 0.234 & 0.208 & 0.324 & 0.379 & 0.078 & 0.300 & 0.249 & 0.591 & 0.213 & 0.642 & 0.969 & 0.633 \\
TimesNet & 0.208 & 0.338 & 0.262 & 0.175 & 0.483 & 0.389 & 0.069 & 0.138 & 0.124 & 0.534 & 0.228 & 0.509 & 0.765 & 0.619 \\
PatchTST & 0.289 & 0.370 & 0.239 & 0.181 & 0.478 & 0.431 & 0.056 & 0.170 & 0.139 & 0.554 & 0.227 & 0.555 & 0.687 & 0.809 \\
LLMTime & \underline{0.127} & \underline{0.123} & \underline{0.077} & \underline{0.143} & \underline{0.148} & \underline{0.131} & 0.045 & \underline{0.123} & 0.142 & \underline{0.236} & \underline{0.069} & \underline{0.111} & \underline{0.115} & \underline{0.103} \\
iTransformer & 0.210 & 0.361 & 0.163 & 0.205 & 0.359 & 0.429 & \underline{0.044} & 0.159 & \underline{0.117} & 0.555 & 0.257 & 0.546 & 0.486 & 0.764 \\
Ours & \textbf{0.094} & \textbf{0.110} & \textbf{0.052} & \textbf{0.097} & \textbf{0.117} & \textbf{0.112} & \textbf{0.035} & \textbf{0.077} & \textbf{0.114} & \textbf{0.189} & \textbf{0.027} & \textbf{0.070} & \textbf{0.106} & \textbf{0.071} \\
\bottomrule
\end{tabular}
\caption{MAE of different models across multiple datasets and forecasting horizons. \textbf{Bold}: the best result under each forecasting horizon, \underline{Underline}: the second best}
\label{tab:performance_autoformer_mae}
\end{table*}

\subsection{Comparative performance of noise levels}\label{appendix:diff_noise_levels}


Our results are based on GPT-3.5-Turbo-Instruction. As shown in ~\tref{mse-mae-noise}, injecting noise at various levels consistently improves forecasting performance across all datasets compared to the original input. 
However, sensitivity to noise levels varies among datasets, and there is no single noise level that universally optimizes performance. The effectiveness of noise injection is influenced by multiple factors, including data distribution, task complexity, and model capacity.
More specifically, experiments demonstrate that moderate noise levels (approximately 0.005 to 0.02) yield the best performance improvements, as such noise helps the model capture key features in the data and enhances prediction accuracy. In contrast, higher noise levels (e.g., 0.05) may excessively disrupt the semantic structure of inputs. In future work, we will focus on identifying the optimal noise level and type tailored to different tasks.

\begin{table}[htbp]
\centering
\setlength{\tabcolsep}{5.5pt}
\begin{tabular}{lcccccccccc}
\toprule
\multirow{2}{*}{Noise Level} 
& \multicolumn{2}{c}{Traffic} 
& \multicolumn{2}{c}{ETTh2} 
& \multicolumn{2}{c}{AusBeer} 
& \multicolumn{2}{c}{GasRateCO2} 
& \multicolumn{2}{c}{MonthlyMilk} \\
\cmidrule(lr){2-3} \cmidrule(lr){4-5} \cmidrule(lr){6-7} \cmidrule(lr){8-9} \cmidrule(lr){10-11}
& MSE & MAE & MSE & MAE & MSE & MAE & MSE & MAE & MSE & MAE \\
\midrule
Original & 0.010 & 0.069 & 0.042 & 0.148 & 0.0011 & 0.026 & 0.038 & 0.160 & 0.0054 & 0.062 \\
0.001    & 0.004 & 0.047 & 0.043 & 0.146 & 0.0009 & 0.024 & 0.034 & 0.150 & 0.0053 & 0.058 \\
0.005    & 0.002 & 0.033 & 0.039 & 0.144 & \textbf{0.0007} & \textbf{0.020} & 0.025 & 0.124 & \textbf{0.0024} & \textbf{0.042} \\
0.01     & 0.002 & 0.034 & 0.039 & 0.138 & 0.0008 & 0.021 & \textbf{0.019} & \textbf{0.106} & 0.0035 & \textbf{0.042} \\
0.02     & \textbf{0.001} & \textbf{0.027} & \textbf{0.024} & \textbf{0.117} & 0.0008 & 0.022 & 0.022 & 0.121 & 0.0044 & 0.054 \\
0.05     & 0.005 & 0.058 & 0.037 & 0.146 & 0.0011 & 0.025 & 0.035 & 0.151 & 0.0033 & 0.045 \\
\bottomrule
\end{tabular}
\caption{MSE and MAE across different noise levels for multiple datasets. \textbf{Bold}: the best result for each dataset}
\label{tab:mse-mae-noise}
\end{table}



\subsection{Impact of sampling strategy}

To investigate how sampling strategies affect the robustness of LLMTime, we evaluate its performance on the ETTh2 dataset by varying both the trial count (1, 5, 10, 15, 20) and the aggregation method (mean vs. median).
As shown in ~\tref{llmtime-sample-agg-mse} and ~\tref{llmtime-sample-agg-mae}, the aggregation strategy and trial count jointly impact performance. 
Specifically, moderately increasing the trial count from 1 to 5 or 10 generally leads to noticeable improvements in MSE and MAE, enhancing result stability and accuracy. However, further increases to 15 or 20 yield diminishing returns.
Moreover, median aggregation consistently produces better optimal results than mean aggregation, indicating its effectiveness in mitigating outliers and reducing variability.

Balancing both model performance and computational cost, we adopt a trial count of 10 and median aggregation for our main experiments.
\begin{table}[htbp]
\centering
\setlength{\tabcolsep}{4pt} 
\begin{tabular}{lcccccccccc}
\toprule
\multirow{2}{*}{Noise Level} & \multicolumn{2}{c}{Trial count = 1} & \multicolumn{2}{c}{Trial count = 5} & \multicolumn{2}{c}{Trial count = 10} & \multicolumn{2}{c}{Trial count = 15} & \multicolumn{2}{c}{Trial count = 20} \\
 & Median & Mean & Median & Mean & Median & Mean & Median & Mean & Median & Mean \\
\midrule
original & 0.067 & 0.055 & 0.044 & 0.050 & 0.042 & 0.036 & 0.050 & 0.053 & 0.042 & 0.048 \\
0.001   & 0.034 & 0.026 & 0.033 & \textbf{0.017} & 0.043 & 0.041 & \textbf{0.015} & 0.038 & 0.027 & 0.046 \\
0.005   & 0.031 & 0.051 & \textbf{0.030} & \textbf{0.027} & 0.039 & \textbf{0.031} & 0.027 & \textbf{0.029} & 0.033 & 0.037 \\
0.01    & \textbf{0.019} & \textbf{0.025} & 0.031 & 0.034 & 0.039 & 0.035 & 0.038 & 0.038 & 0.041 & 0.036 \\
0.02    & 0.036 & 0.036 & 0.037 & 0.033 & \textbf{0.024} & 0.037 & 0.047 & 0.038 & 0.031 & 0.035 \\
0.05    & 0.057 & 0.031 & 0.046 & 0.036 & 0.037 & 0.040 & \textbf{0.035} & 0.033 & 0.035 & 0.037 \\
\bottomrule
\end{tabular}
\caption{MSE under different trial counts and aggregation methods. “Trial count” refers to the number of repeated runs. “Mean” denotes averaging the results across all runs, while “Median” denotes taking the median across time steps within each run.}
\label{tab:llmtime-sample-agg-mse}
\end{table}

\begin{table}[htbp]
\centering
\setlength{\tabcolsep}{4pt} 
\begin{tabular}{lcccccccccc}
\toprule
\multirow{2}{*}{Noise Level} & \multicolumn{2}{c}{Trial count = 1} & \multicolumn{2}{c}{Trial count = 5} & \multicolumn{2}{c}{Trial count = 10} & \multicolumn{2}{c}{Trial count = 15} & \multicolumn{2}{c}{Trial count = 20} \\
 & Median & Mean & Median & Mean & Median & Mean & Median & Mean & Median & Mean \\
\midrule
original & 0.198 & 0.181 & 0.147 & 0.163 & 0.148 & 0.153 & 0.179 & 0.177 & 0.159 & 0.169 \\
0.001 & 0.160 & 0.127 & 0.137 & \textbf{0.103} & 0.146 & 0.157 & \textbf{0.093} & 0.150 & \textbf{0.129} & 0.164 \\
0.005 & 0.133 & 0.167 & 0.138 & 0.136 & 0.144 & \textbf{0.135} & 0.121 & \textbf{0.130} & 0.139 & 0.154 \\
0.01  & \textbf{0.111} & 0.137 & 0.137 & 0.138 & 0.138 & 0.146 & 0.150 & 0.147 & 0.154 & 0.146 \\
0.02  & 0.163 & 0.159 & \textbf{0.130} & 0.137 & \textbf{0.117} & 0.148 & 0.174 & 0.152 & 0.132 & 0.148 \\
0.05  & 0.186 & \textbf{0.124} & 0.172 & 0.143 & 0.146 & 0.146 & 0.144 & 0.142 & 0.140 & \textbf{0.144} \\
\bottomrule
\end{tabular}
\caption{MAE under different trial counts and aggregation methods. }
\label{tab:llmtime-sample-agg-mae}
\end{table}


\subsection{Forecasting performance under different noise types}

\tref{mse-mae-noise-type} reports the MSE and MAE results of our method on the ETTh2 dataset.
The evaluation is conducted under six different noise types: Uniform, Laplace, Geometric, Gamma, Beta, and Gaussian. The theoretical details of these noise distributions are provided in \Aref{noise_distribution}.

\begin{table}[H]
\centering
\setlength{\tabcolsep}{4pt} 
\begin{tabular}{lcccccccccccc}
\toprule
\multirow{2}{*}{Noise Level} 
& \multicolumn{2}{c}{Uniform} 
& \multicolumn{2}{c}{Laplace} 
& \multicolumn{2}{c}{Geometric} 
& \multicolumn{2}{c}{Gamma} 
& \multicolumn{2}{c}{Beta} 
& \multicolumn{2}{c}{Gaussian} \\
\cmidrule(lr){2-3} \cmidrule(lr){4-5} \cmidrule(lr){6-7} \cmidrule(lr){8-9} \cmidrule(lr){10-11} \cmidrule(lr){12-13}
 & MSE & MAE & MSE & MAE & MSE & MAE & MSE & MAE & MSE & MAE & MSE & MAE \\
\midrule
Original & 0.046 & 0.173 & 0.036 & 0.147 & 0.044 & 0.160 & 0.038 & 0.155 & 0.046 & 0.172 & 0.042 & 0.148 \\
0.001    & 0.031 & 0.146 & 0.027 & 0.133 & 0.024 & 0.123 & \textbf{0.025} & \textbf{0.120} & \textbf{0.028} & \textbf{0.124} & 0.043 & 0.146 \\
0.005    & 0.034 & 0.149 & \textbf{0.024} & \textbf{0.118} & 0.040 & 0.154 & 0.034 & 0.144 & 0.036 & 0.139 & 0.039 & 0.144 \\
0.01     & 0.036 & 0.154 & 0.032 & 0.133 & 0.043 & 0.152 & 0.030 & 0.137 & 0.040 & 0.154 & 0.039 & 0.138 \\
0.02     & \textbf{0.025} & \textbf{0.119} & 0.035 & 0.136 & \textbf{0.024} & \textbf{0.120} & 0.031 & 0.139 & 0.039 & 0.139 & \textbf{0.024} & \textbf{0.117} \\
0.05     & 0.030 & 0.134 & 0.027 & 0.131 & 0.034 & 0.141 & 0.028 & 0.121 & 0.034 & 0.137 & 0.037 & 0.146 \\
\bottomrule
\end{tabular}
\caption{MSE and MAE under different noise types.
\textbf{Bold}: the best result for each noise level.}
\label{tab:mse-mae-noise-type}
\end{table}

\subsection{Average forecasting performance of models}
\label{appendix:performances}


For short- and long-term forecasting tasks, we evaluate three major TS benchmarks: Darts, Memorization, and Autoformer, which contain 8, 3, and 7 sub-datasets. For each benchmark, we report the average performance across its sub-datasets. For Autoformer, following the LLMTime setting, we present results for forecasting horizons of 96 and 192 steps, as illustrated in \fref{MAE_combined}


\begin{table*}[tbph]
\centering
\small
\setlength{\tabcolsep}{2pt} 
\begin{tabular}{lcccccccccccccc}
\toprule
\multirow{2}{*}{Datasets} & 
\multicolumn{2}{c}{\shortstack{NLTS 
}} & 
\multicolumn{2}{c}{\shortstack{LLMTime 
}} & 
\multicolumn{2}{c}{\shortstack{N-HiTS 
}} & 
\multicolumn{2}{c}{\shortstack{N-BEATS 
}} & 
\multicolumn{2}{c}{\shortstack{TCN 
}} & 
\multicolumn{2}{c}{\shortstack{SM-GP 
}} & 
\multicolumn{2}{c}{\shortstack{ARIMA 
}} \\
\cmidrule(lr){2-3} 
\cmidrule(lr){4-5} 
\cmidrule(lr){6-7} 
\cmidrule(lr){8-9} 
\cmidrule(lr){10-11} 
\cmidrule(lr){12-13} 
\cmidrule(lr){14-15} 
& MSE & MAE & MSE & MAE & MSE & MAE & MSE & MAE & MSE & MAE & MSE & MAE & MSE & MAE \\
\midrule
Darts (avg.) & \textbf{0.015} & \textbf{0.082} & 0.025 & \underline{0.110} & 0.044 & 0.154 & 0.052 & 0.165 & 0.037 & 0.152 & 0.043 & 0.159 & \underline{0.024} & 0.117 \\
Memorization (avg.) & \textbf{0.020} & \textbf{0.073} & 0.046 & \underline{0.126} & 0.092 & 0.178 & 0.149 & 0.287 & \underline{0.043} & 0.130 & 0.086 & 0.199 & 0.207 & 0.314 \\
\bottomrule
\end{tabular}
\caption{
Average zero-shot forecasting performance on the short-term benchmarks.
\textbf{Bold} and \underline{underline}: the best and the second-best results.
}
\label{tab:mse-mae-short-avg}
\end{table*}

\begin{table*}[tbph]
\centering
\small
\setlength{\tabcolsep}{4pt}
\begin{tabular}{lccccccccccccccc}
\toprule
\multirow{2}{*}{Prediction Length} & 
\multicolumn{2}{c}{\shortstack{NLTS 
}} & 
\multicolumn{2}{c}{\shortstack{iTransformer 
}} & 
\multicolumn{2}{c}{\shortstack{LLMTime 
}} & 
\multicolumn{2}{c}{\shortstack{PatchTST 
}} & 
\multicolumn{2}{c}{\shortstack{TimesNet 
}} & 
\multicolumn{2}{c}{\shortstack{Autoformer 
}} & 
\multicolumn{2}{c}{\shortstack{Informer 
}} \\
\cmidrule(lr){2-3} \cmidrule(lr){4-5} \cmidrule(lr){6-7} \cmidrule(lr){8-9} \cmidrule(lr){10-11} \cmidrule(lr){12-13} \cmidrule(lr){14-15}
 & MSE & MAE & MSE & MAE & MSE & MAE & MSE & MAE & MSE & MAE & MSE & MAE & MSE & MAE \\
\midrule
Autoformer-96 (avg.) & \textbf{0.021} & \textbf{0.078} & 0.125 & 0.234 & \underline{0.029} & \underline{0.103} & 0.230 & 0.302 & 0.192 & 0.306 & 0.247 & 0.347 & 1.649 & 0.660 \\
Autoformer-192 (avg.) & \textbf{0.018} & \textbf{0.104} & 0.325 & 0.431 & \underline{0.031} & \underline{0.139} & 0.365 & 0.439 & 0.285 & 0.386 & 0.370 & 0.470 & 1.519 & 0.721 \\
\bottomrule
\end{tabular}
\caption{
Average zero-shot forecasting performance on the long-term benchmarks
for different prediction lengths. 
\textbf{Bold} and \underline{underline}: the best and the second best results.
}
\label{tab:mse-mae-long-avg}
\end{table*}

\begin{figure*}[htbp]
\small
\centering
\begin{tabular}{p{0.0mm}*{2}{c}}
& \figref{0.35}{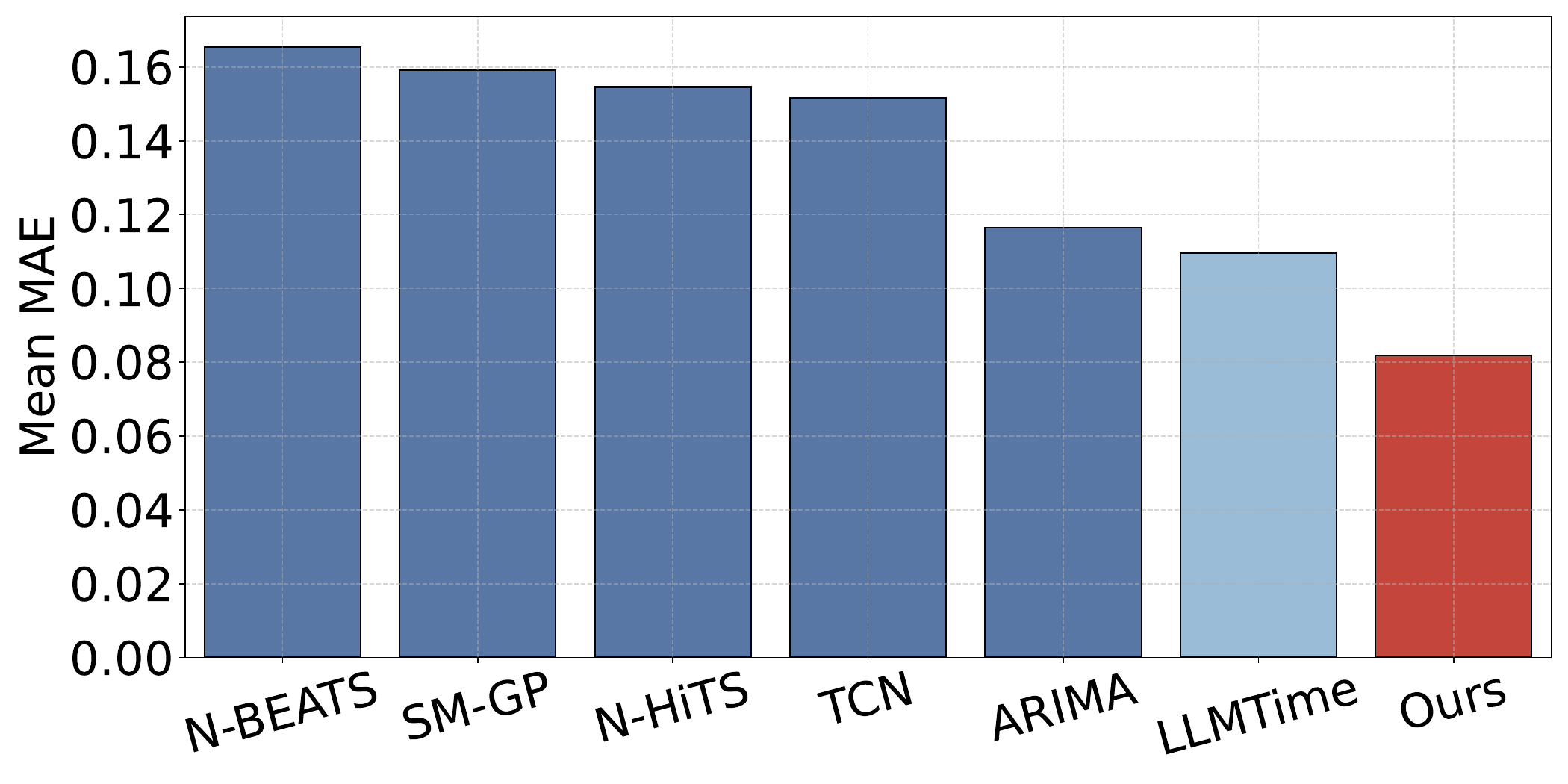} & \figref{0.35}{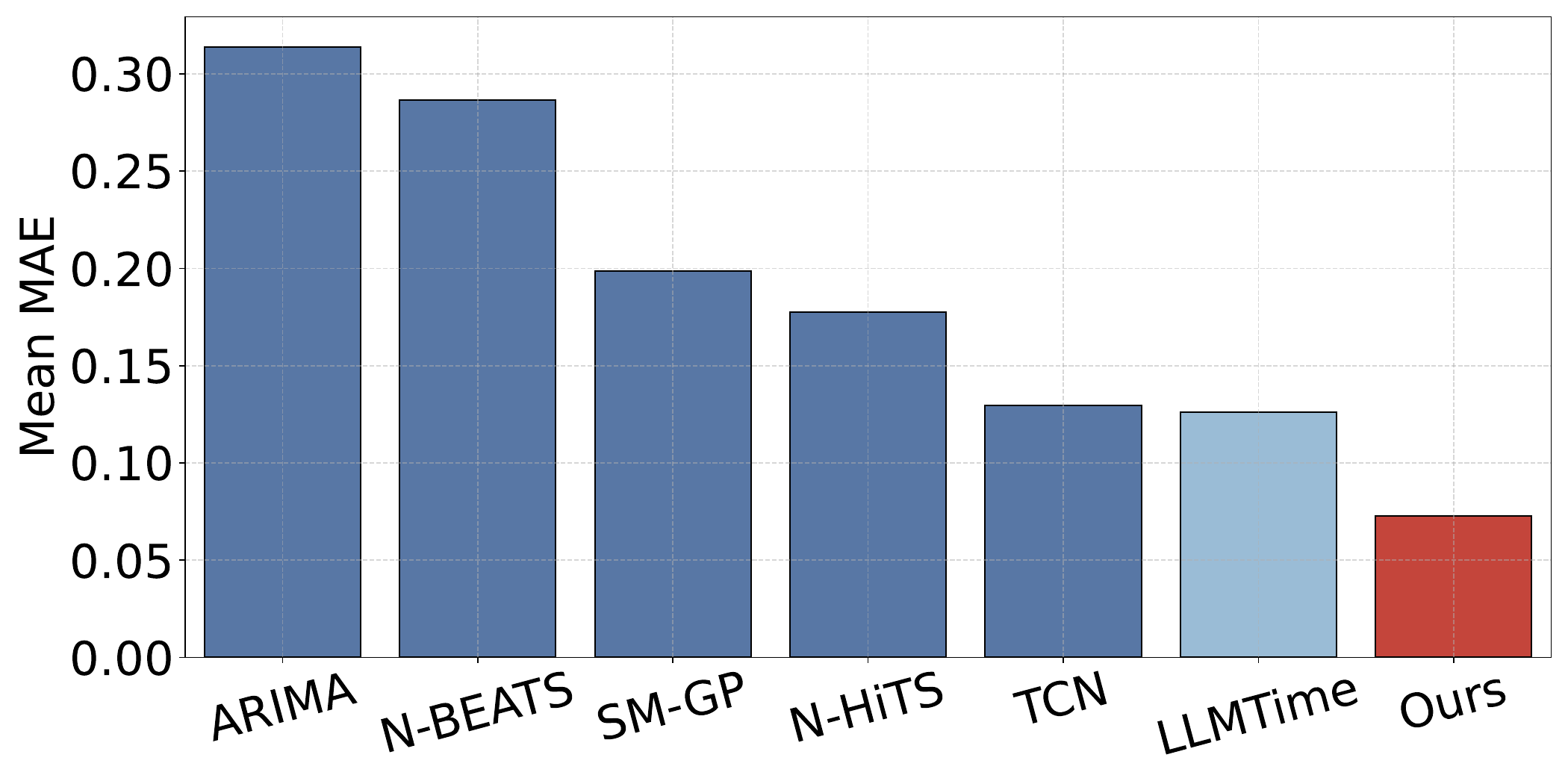} \\
& (a) Darts & (b) Memorization \\
& \figref{0.35}{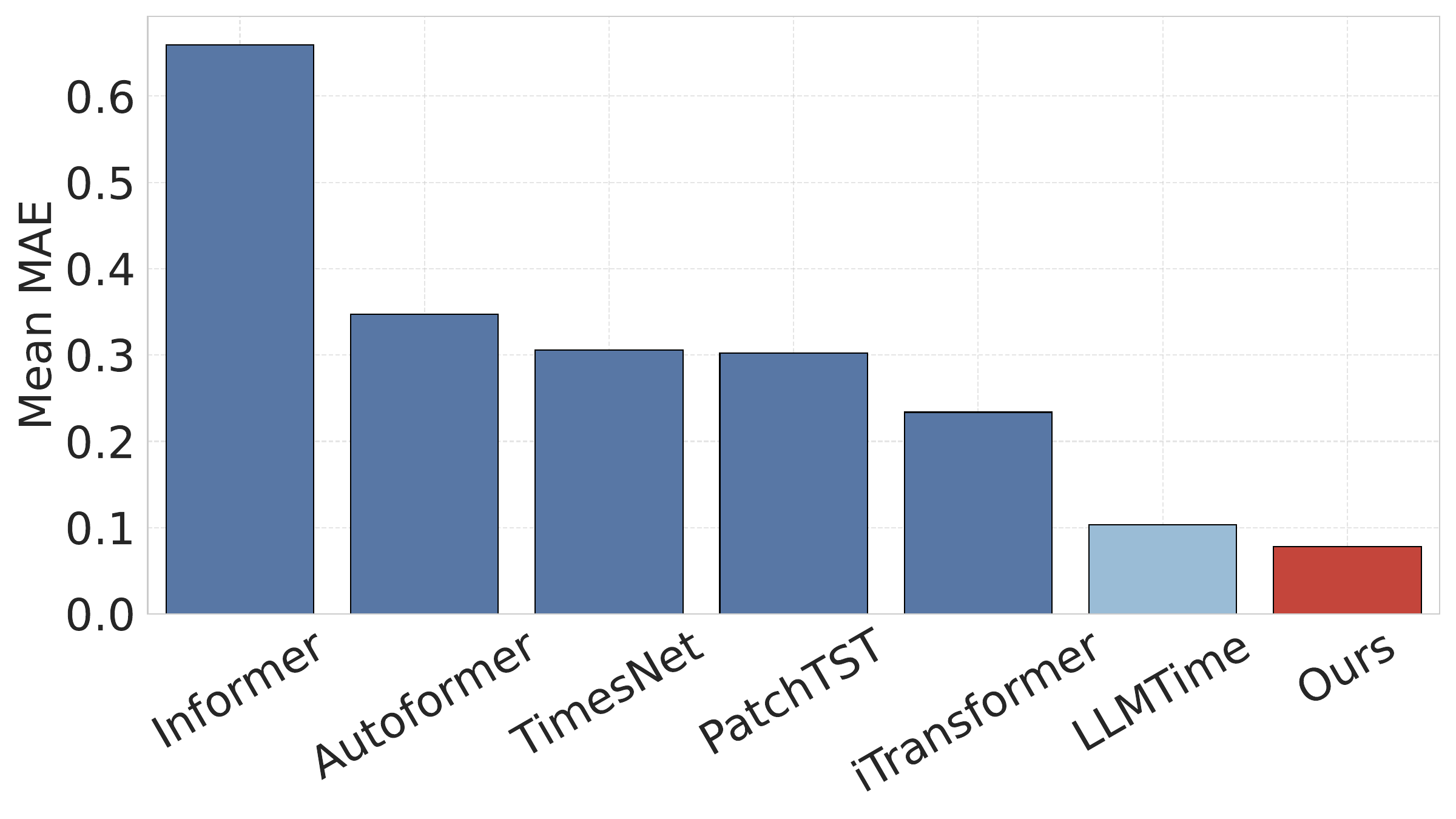} & \figref{0.35}{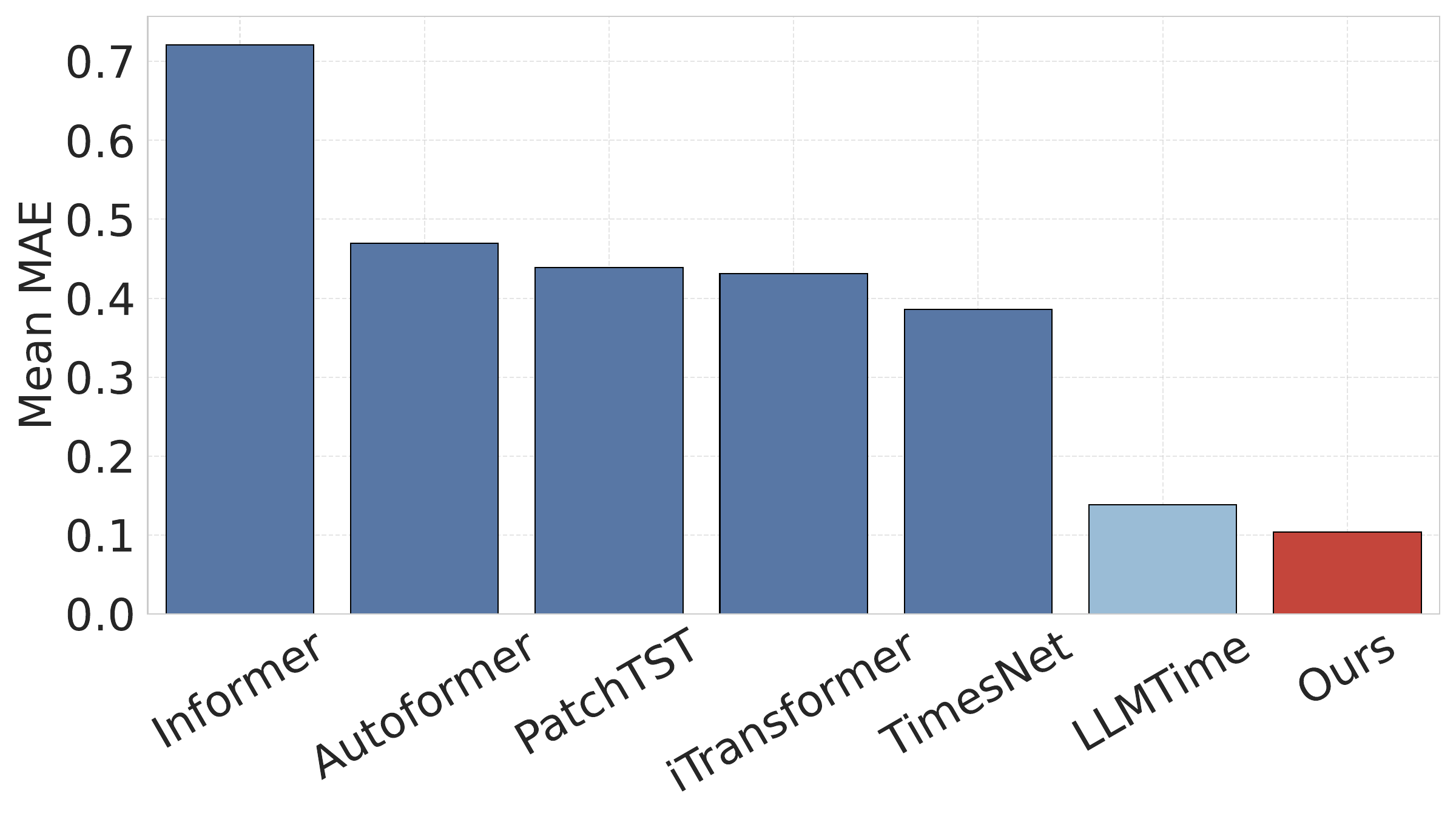} \\
& (c) Autoformer (horizon=96) & (d) Autoformer (horizon=192)
\end{tabular}
\caption{
Average zero-shot forecasting performance (MAE) across the Darts, Memorization, and Autoformer benchmarks.
}
\label{fig:MAE_combined}
\end{figure*}

\clearpage

\subsection{Visualization of forecasting across all datasets}
Due to the limited page space, we selectively present the forecasting plots. 

\begin{figure}[H]
\renewcommand{\tabcolsep}{0.5mm}
\small
\centering
\begin{tabular}{p{0.0mm}*{3}{c}}
& \figref{0.32}{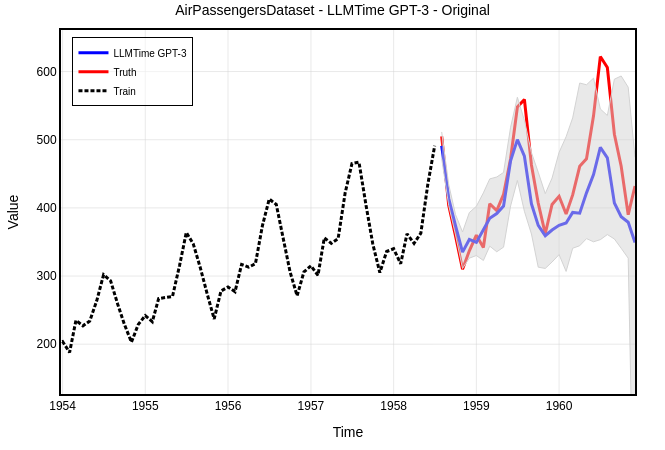} & \figref{0.32}{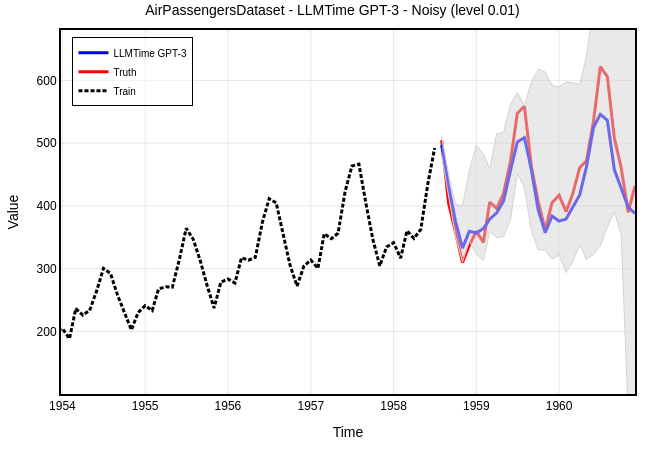} & \figref{0.32}{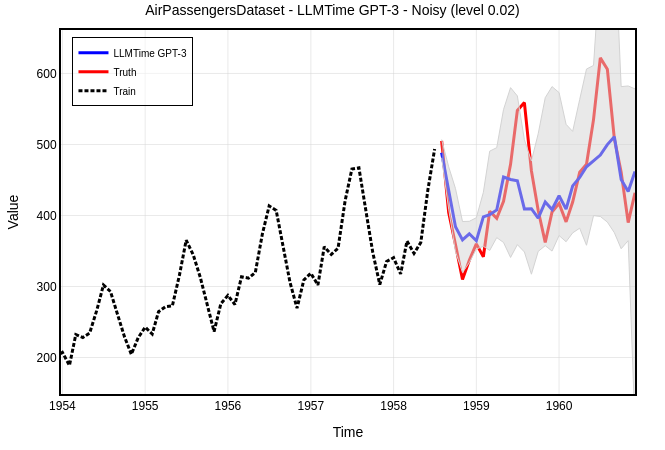} \\
& (a) GPT 3.5 without NLTS & (b) GPT 3.5 with NLTS ($\alpha=0.01$) & (c) GPT 3.5 with NLTS ($\alpha=0.02$).  \\
& \figref{0.32}{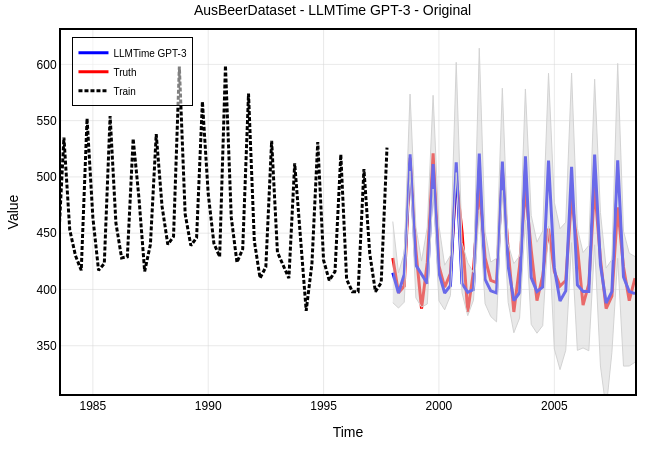} & \figref{0.32}{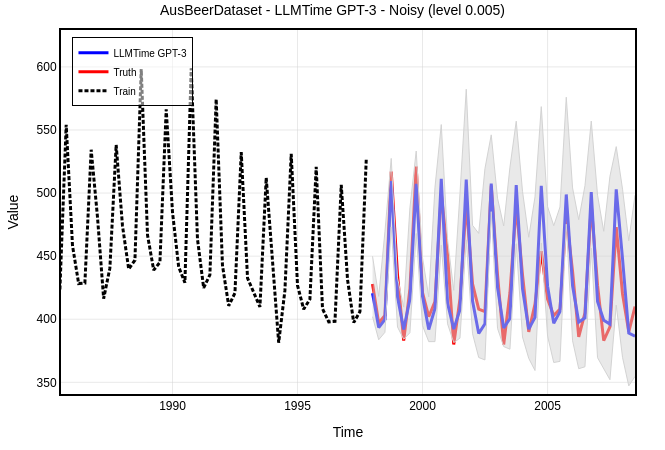} & \figref{0.32}{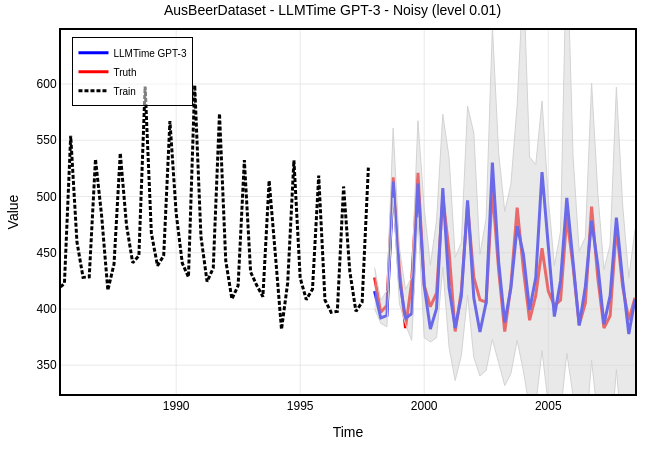} \\
& (a) GPT 3.5 without NLTS & (b) GPT 3.5 with NLTS ($\alpha=0.005$) & (c) GPT 3.5 with NLTS ($\alpha=0.01$).  \\
& \figref{0.32}{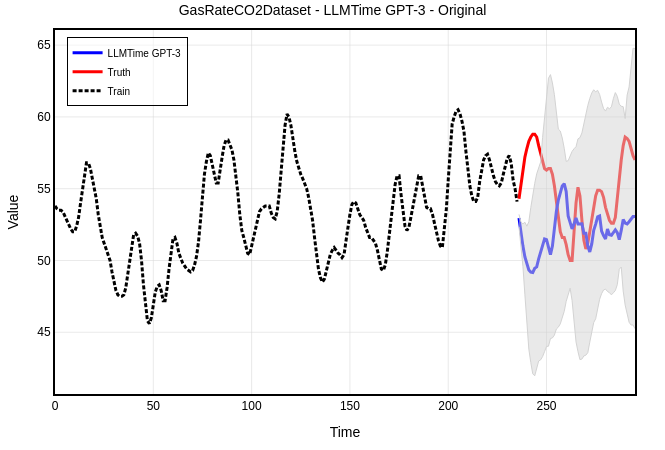} & \figref{0.32}{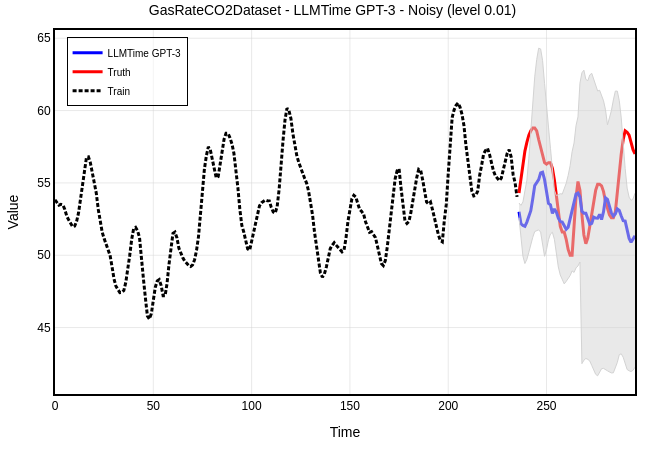} & \figref{0.32}{GasRateCO2Dataset_LLMTimeGPT-3_noisy_0.02.png} \\
& (a) GPT 3.5 without NLTS & (b) GPT 3.5 with NLTS ($\alpha=0.01$) & (c) GPT 3.5 with NLTS ($\alpha=0.02$).  \\
& \figref{0.32}{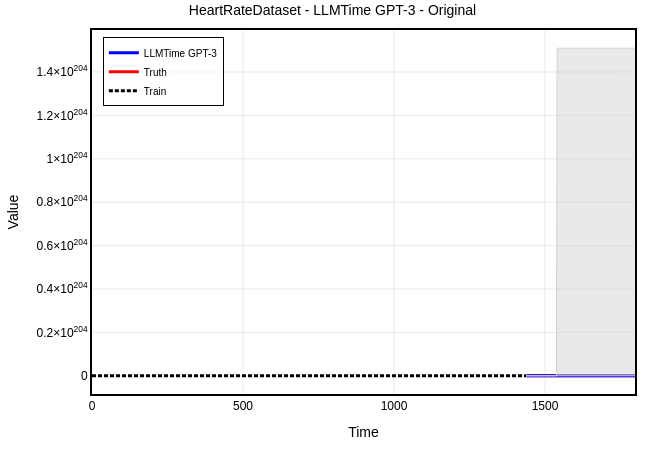} & \figref{0.32}{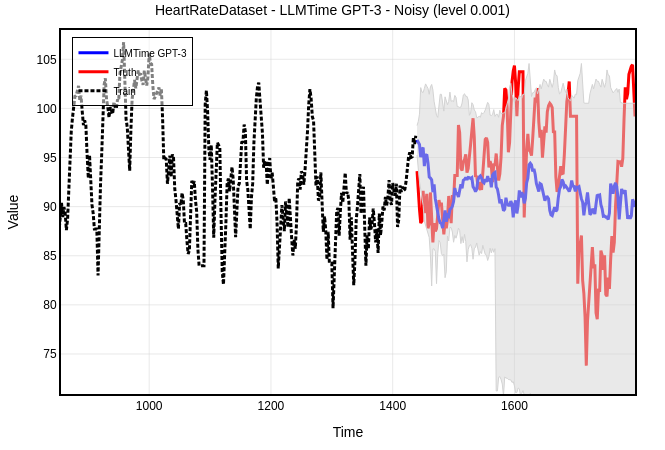} & \figref{0.32}{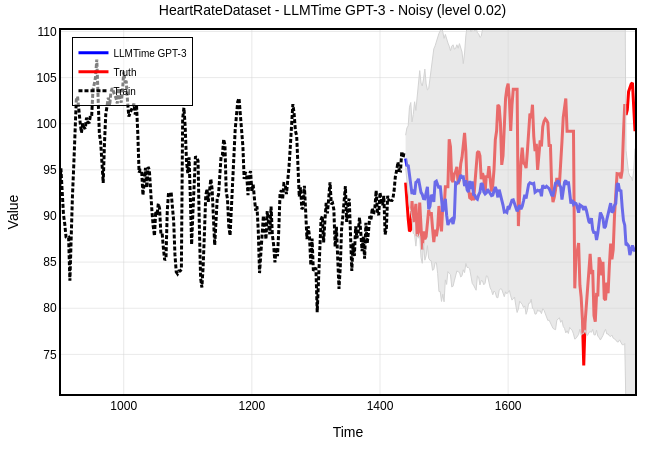} \\
& (a) GPT 3.5 without NLTS & (b) GPT 3.5 with NLTS ($\alpha=0.001$) & (c) GPT 3.5 with NLTS ($\alpha=0.02$).  \\
\end{tabular}\vspace{-3mm}
\caption{Forecasting results of GPT-3.5 on multiple time series from the Darts dataset under different noise injection levels. Each row corresponds to a specific dataset (AirPassengers, AusBeer, GasRateCO2, and HeartRate), with (a) showing the original prediction without noise, and (b)--(c) showing predictions with increasing levels of noise ($\alpha$).}\label{fig:pred-darts1}
\end{figure}

\begin{figure}[H]
\renewcommand{\tabcolsep}{0.5mm}
\small
\centering
\begin{tabular}{p{0.0mm}*{3}{c}}
& \figref{0.32}{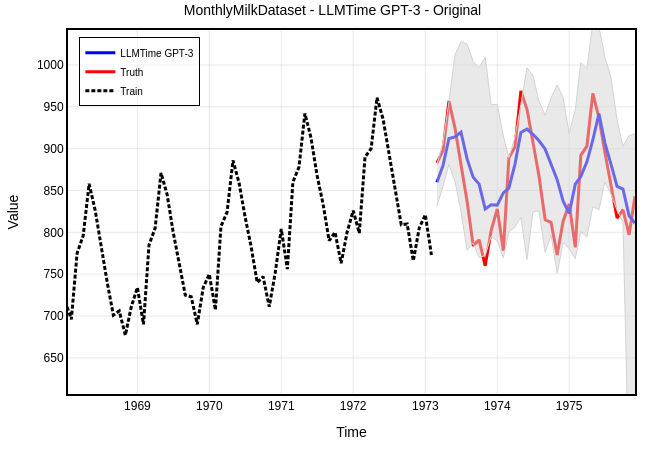} & \figref{0.32}{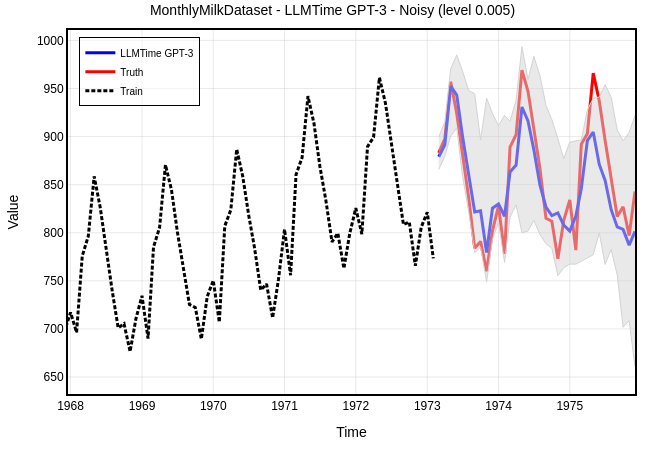} & \figref{0.32}{MonthlyMilkDataset_LLMTimeGPT-3_noisy_0.01.png} \\
& (a) GPT 3.5 without NLTS & (b) GPT 3.5 with NLTS ($\alpha=0.005$) & (c) GPT 3.5 with NLTS ($\alpha=0.01$).  \\
& \figref{0.32}{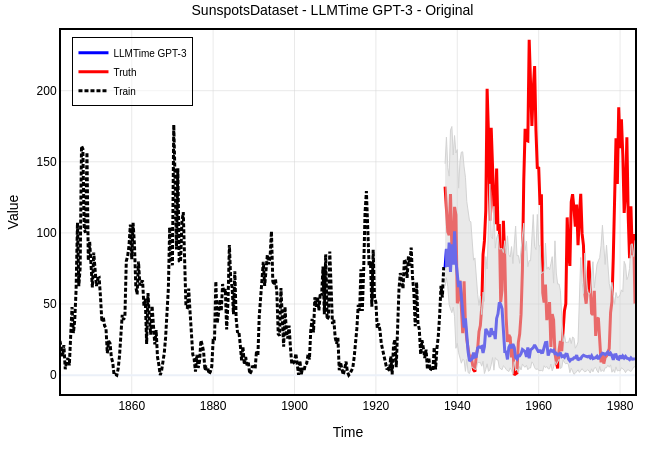} & \figref{0.32}{SunspotsDataset_LLMTimeGPT-3_noisy_0.01.png} & \figref{0.32}{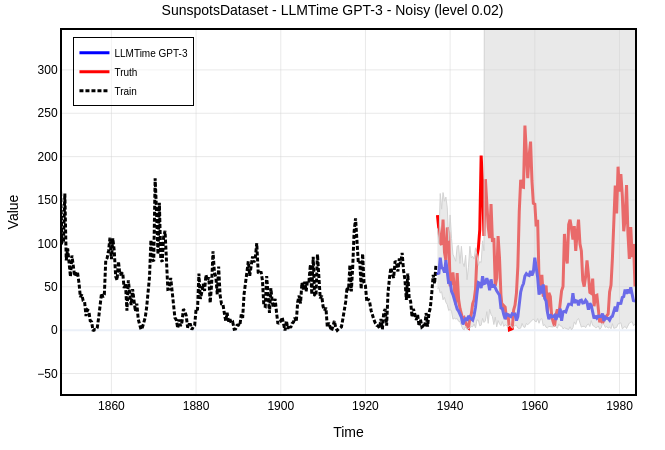} \\
& (a) GPT 3.5 without NLTS & (b) GPT 3.5 with NLTS ($\alpha=0.01$) & (c) GPT 3.5 with NLTS ($\alpha=0.02$).  \\
& \figref{0.32}{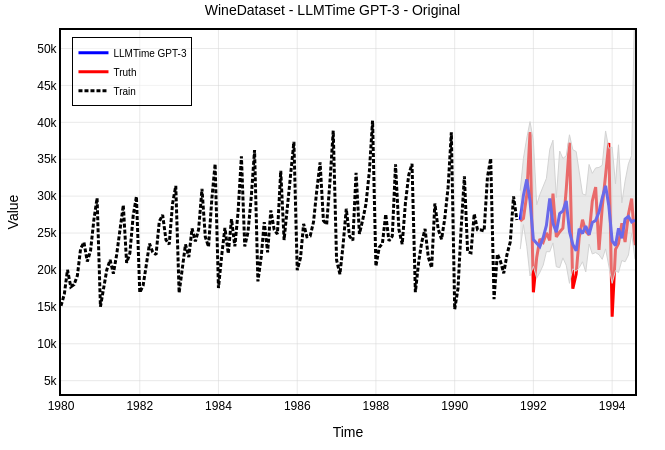} & \figref{0.32}{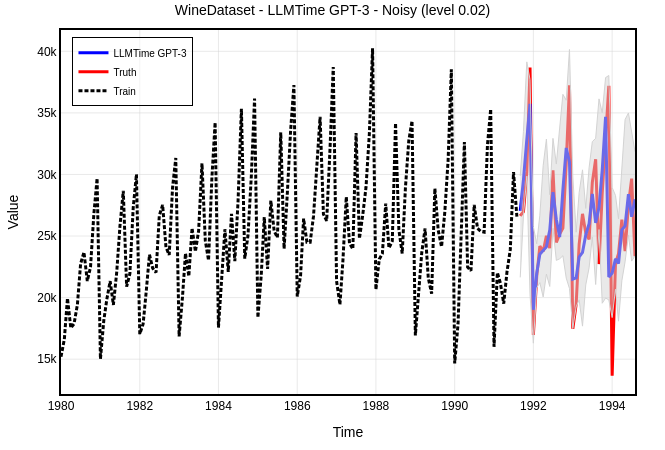} & \figref{0.32}{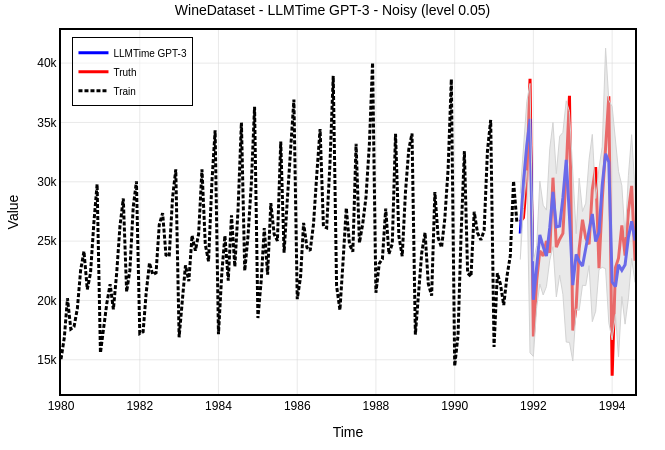} \\
& (a) GPT 3.5 without NLTS & (b) GPT 3.5 with NLTS ($\alpha=0.02$) & (c) GPT 3.5 with NLTS ($\alpha=0.05$).  \\
& \figref{0.32}{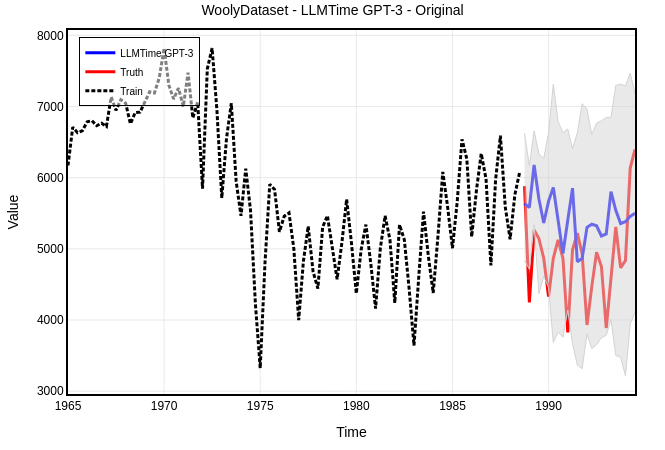} & \figref{0.32}{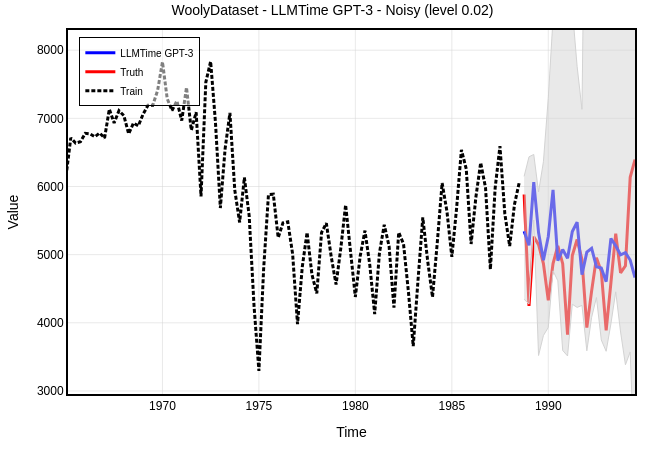} & \figref{0.32}{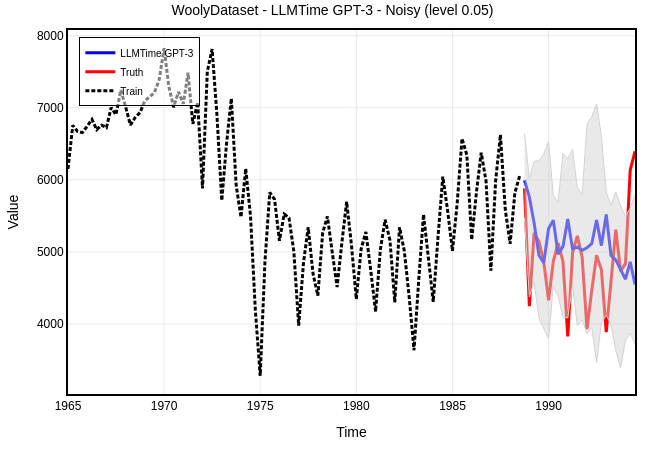} \\
& (a) GPT 3.5 without NLTS & (b) GPT 3.5 with NLTS ($\alpha=0.02$) & (c) GPT 3.5 with NLTS ($\alpha=0.05$).  \\
\end{tabular}\vspace{-3mm}
\caption{Forecasting results of GPT-3.5 on multiple time series from the Darts dataset under different noise injection levels. Each row corresponds to a specific dataset (MonthlyMilk, Sunspots, Wine, Woolly), with (a) showing the original prediction without noise, and (b)--(c) showing predictions with increasing levels of noise ($\alpha$).}\label{fig:fig:pred-darts2}
\end{figure}

\begin{figure}[H]
\renewcommand{\tabcolsep}{0.5mm}
\small
\centering
\begin{tabular}{p{0.0mm}*{3}{c}}
& \figref{0.32}{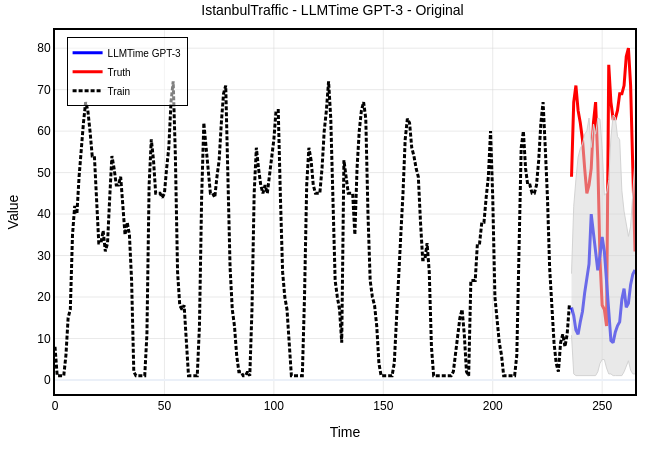} & \figref{0.32}{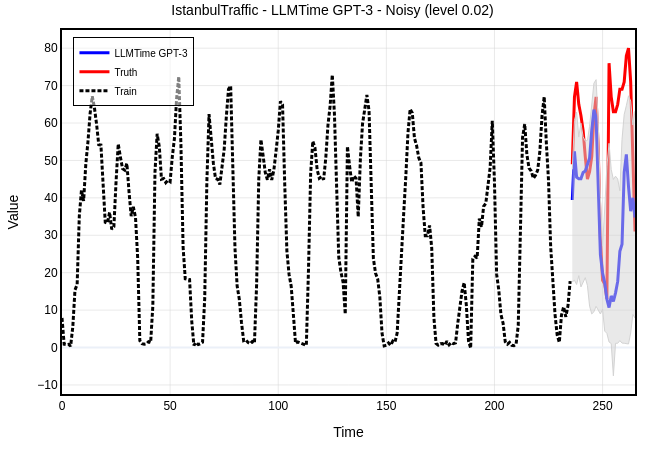} & \figref{0.32}{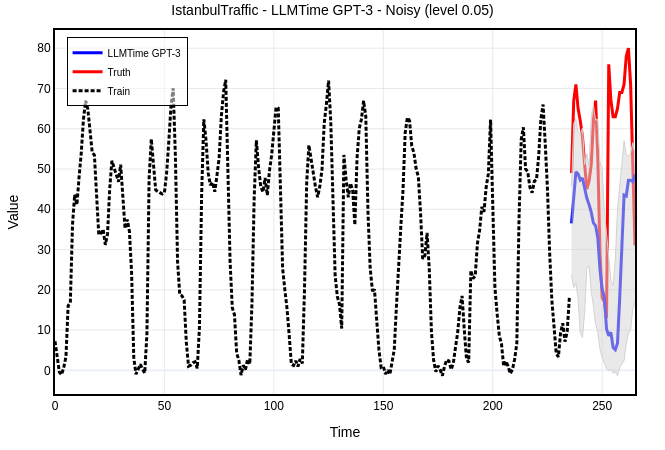} \\
& (a) GPT 3.5 without NLTS & (b) GPT 3.5 with NLTS ($\alpha=0.02$) & (c) GPT 3.5 with NLTS ($\alpha=0.05$).  \\
& \figref{0.32}{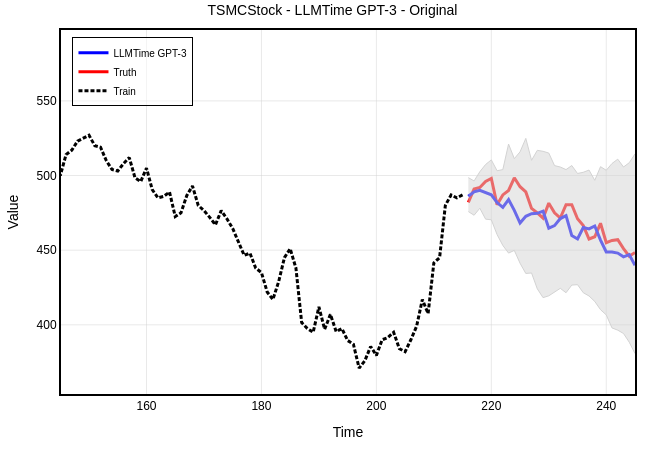} & \figref{0.32}{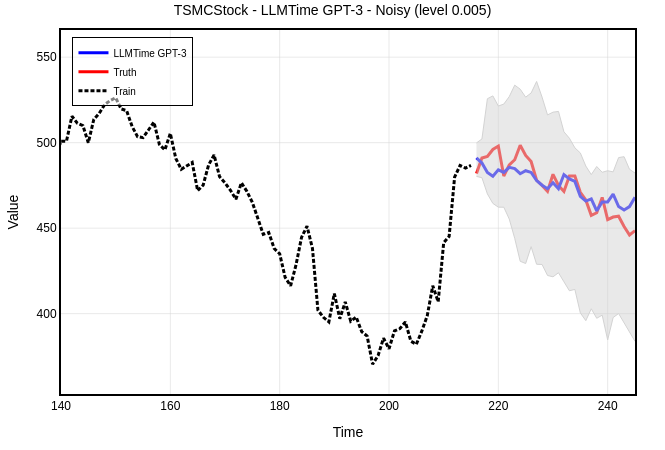} & \figref{0.32}{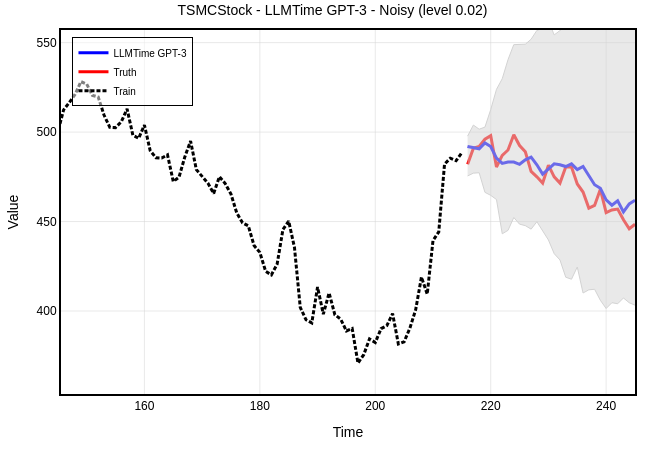} \\
& (a) GPT 3.5 without NLTS & (b) GPT 3.5 with NLTS ($\alpha=0.005$) & (c) GPT 3.5 with NLTS ($\alpha=0.02$).  \\
& \figref{0.32}{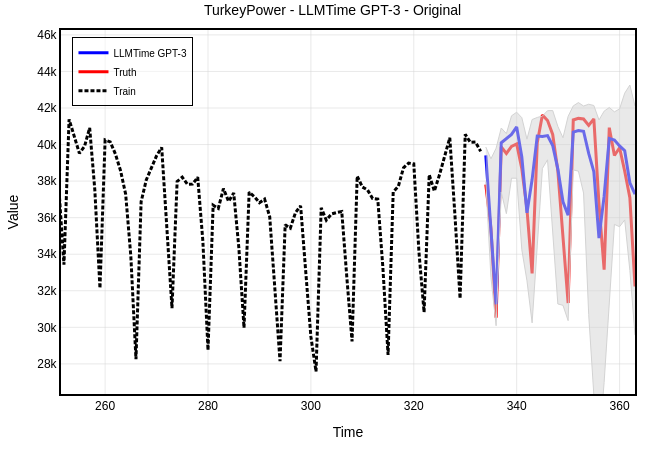} & \figref{0.32}{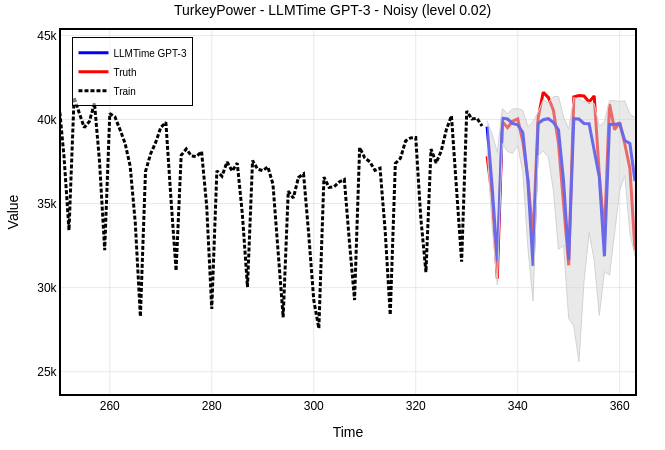} & \figref{0.32}{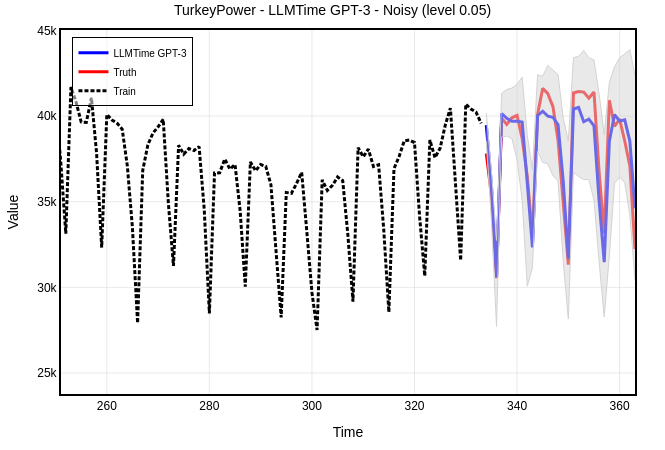} \\
& (a) GPT 3.5 without NLTS & (b) GPT 3.5 with NLTS ($\alpha=0.02$) & (c) GPT 3.5 with NLTS ($\alpha=0.05$).  \\
\end{tabular}\vspace{-3mm}
\caption{Forecasting results of GPT-3.5 on multiple time series from the Memorization dataset under different noise injection levels. Each row corresponds to a specific dataset (IstanbulTraffic, TSMCStock, TurkeyPower), with (a) showing the original prediction without noise, and (b)--(c) showing predictions with increasing levels of noise ($\alpha$).}\label{fig:pred-Memorization}
\end{figure}

\begin{figure}[H]
\renewcommand{\tabcolsep}{0.5mm}
\small
\centering
\begin{tabular}{p{0.0mm}*{3}{c}}
& \figref{0.32}{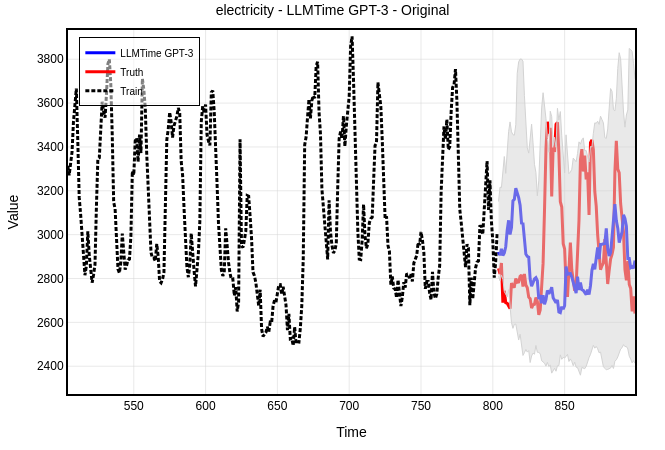} & \figref{0.32}{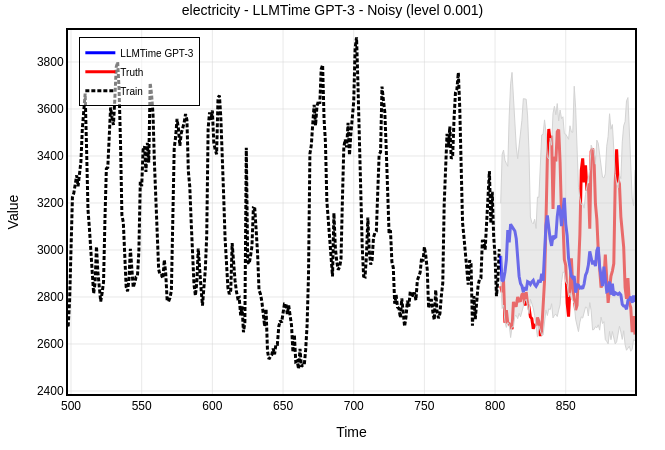} & \figref{0.32}{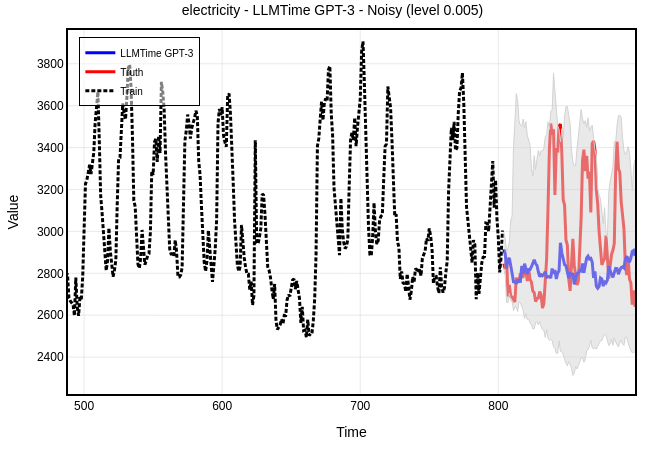} \\
& (a) GPT 3.5 without NLTS & (b) GPT 3.5 with NLTS ($\alpha=0.001$) & (c) GPT 3.5 with NLTS ($\alpha=0.001$).  \\
& \figref{0.32}{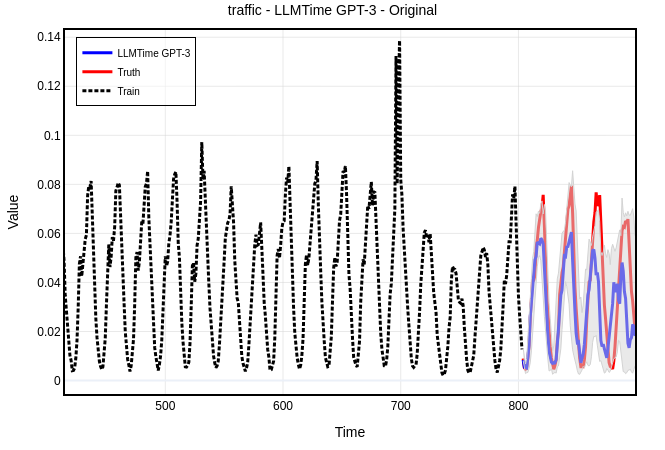} & \figref{0.32}{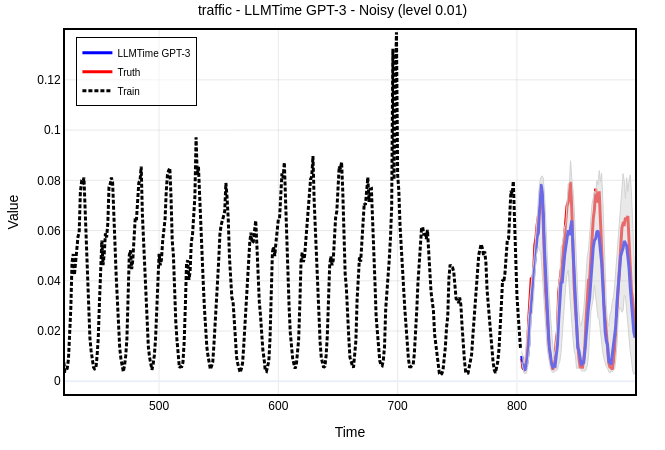} & \figref{0.35}{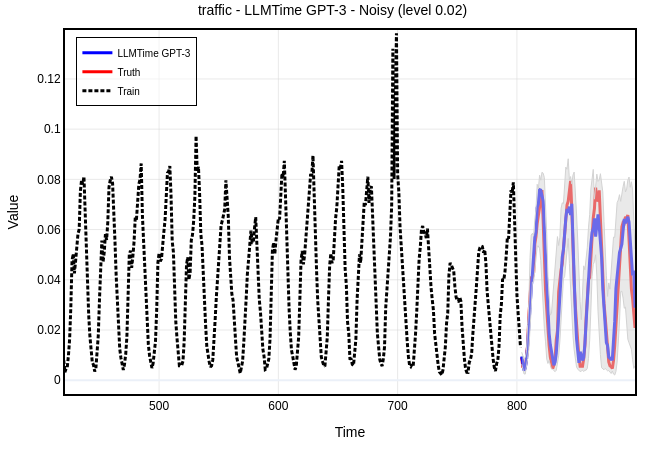} \\
& (a) GPT 3.5 without NLTS & (b) GPT 3.5 with NLTS ($\alpha=0.01$) & (c) GPT 3.5 with NLTS ($\alpha=0.02$).  \\
& \figref{0.32}{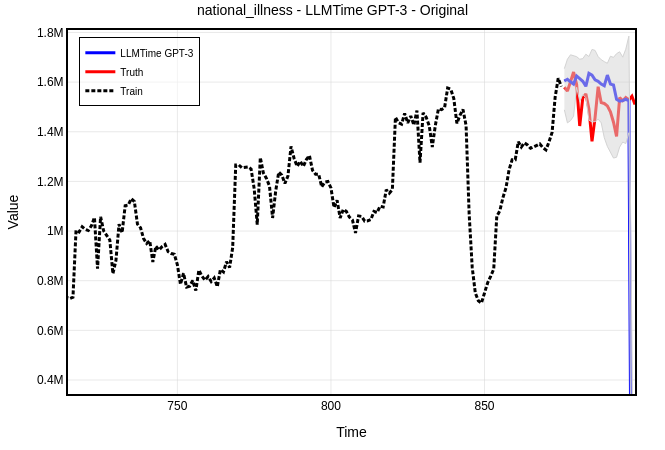} & \figref{0.32}{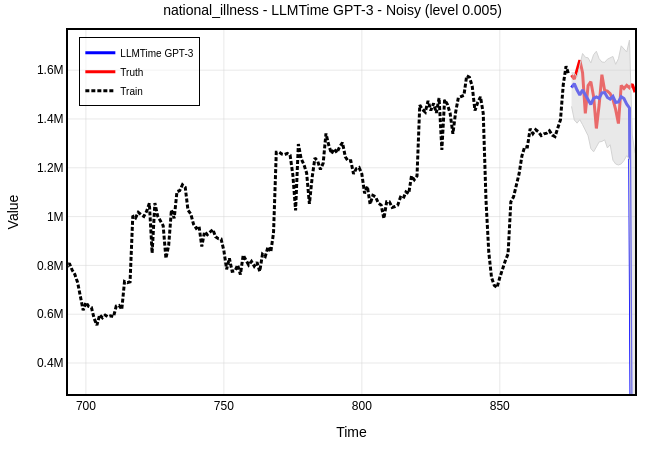} & \figref{0.32}{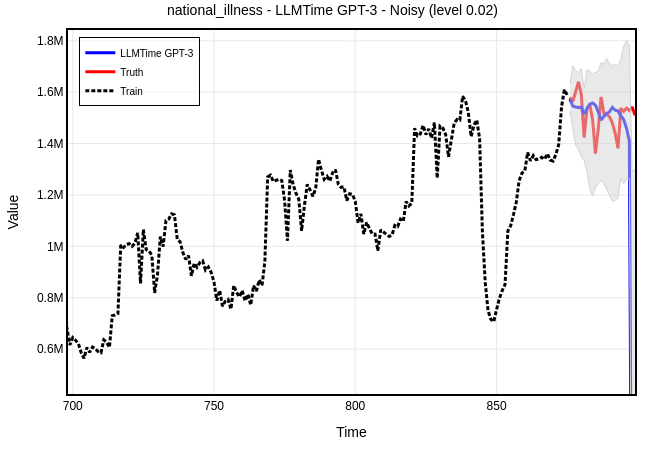} \\
& (a) GPT 3.5 without NLTS & (b) GPT 3.5 with NLTS ($\alpha=0.005$) & (c) GPT 3.5 with NLTS ($\alpha=0.02$).  \\
\end{tabular}\vspace{-3mm}
\caption{Forecasting results of GPT-3.5 on multiple time series from the Autoformer dataset under different noise injection levels. Each row corresponds to a specific dataset (Electricity, Traffic, ILI), with (a) showing the original prediction without noise, and (b)--(c) showing predictions with increasing levels of noise ($\alpha$).}\label{fig:pred-autoformer1}
\end{figure}

\begin{figure}[H]
\renewcommand{\tabcolsep}{0.5mm}
\small
\centering
\begin{tabular}{p{0.0mm}*{3}{c}}
& \figref{0.32}{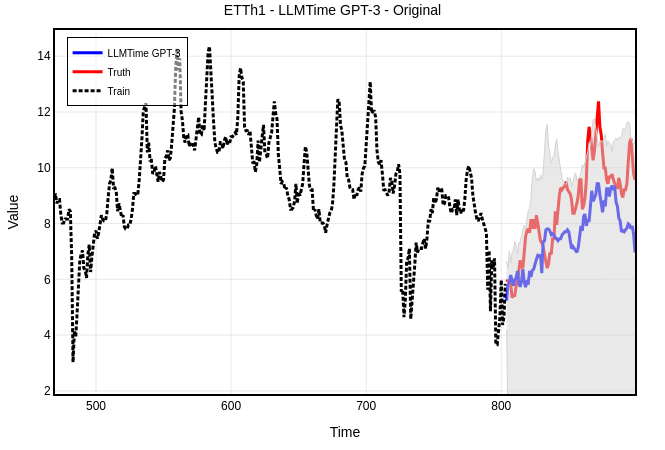} & \figref{0.32}{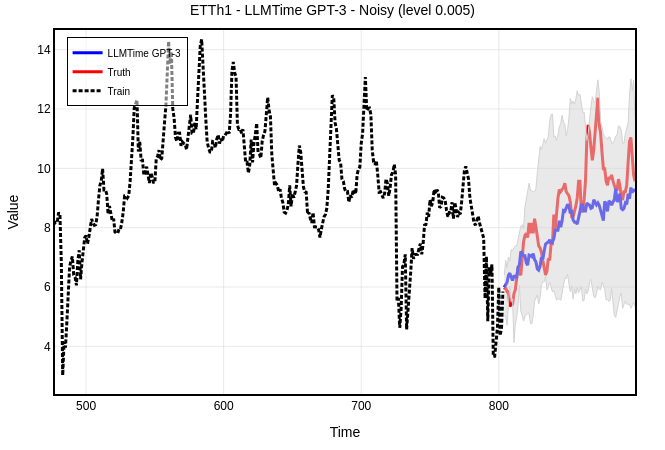} & \figref{0.32}{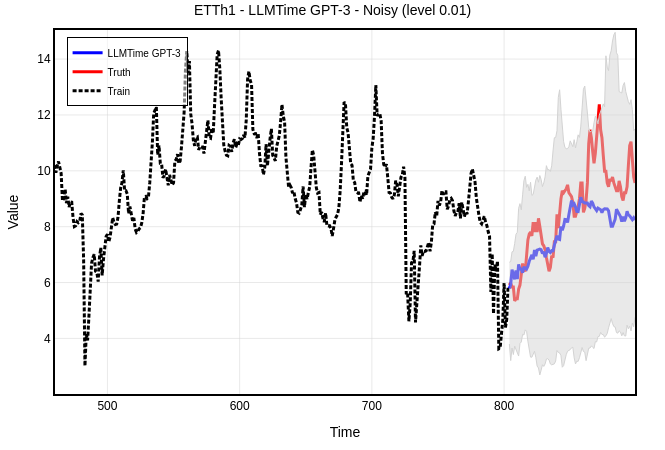} \\
& (a) GPT 3.5 without NLTS & (b) GPT 3.5 with NLTS ($\alpha=0.005$) & (c) GPT 3.5 with NLTS ($\alpha=0.01$).  \\
& \figref{0.32}{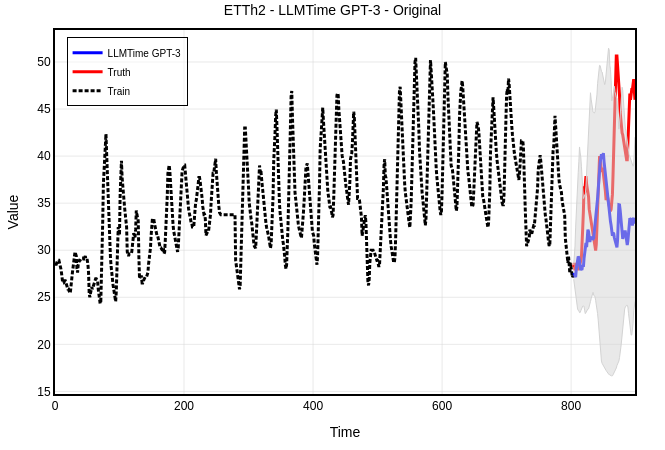} & \figref{0.32}{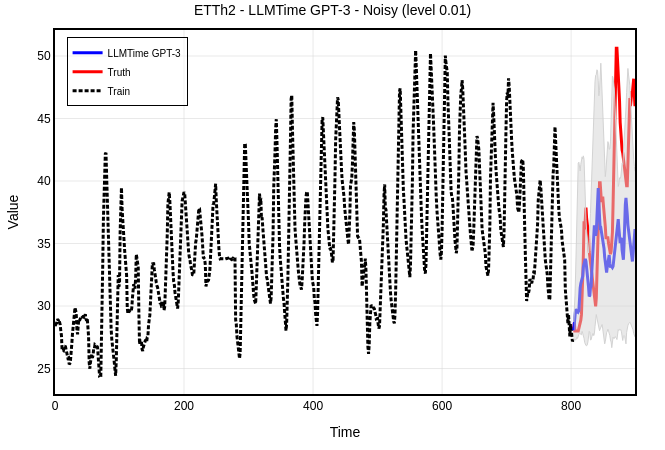} & \figref{0.32}{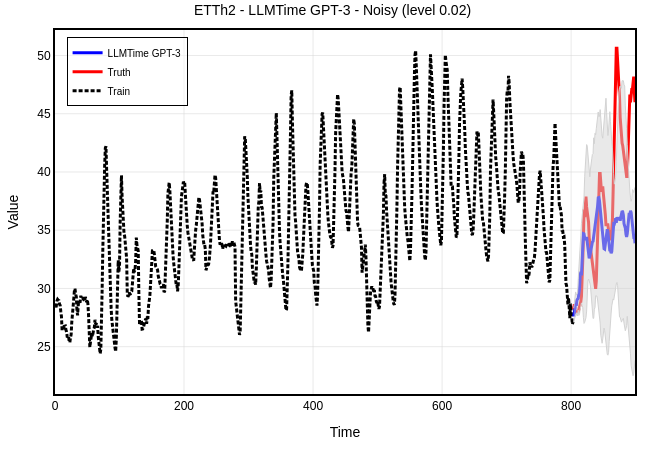} \\
& (a) GPT 3.5 without NLTS & (b) GPT 3.5 with NLTS ($\alpha=0.01$) & (c) GPT 3.5 with NLTS ($\alpha=0.02$).  \\
& \figref{0.32}{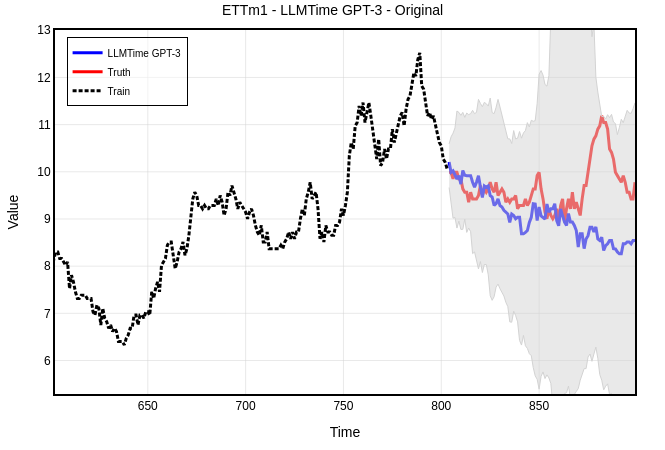} & \figref{0.32}{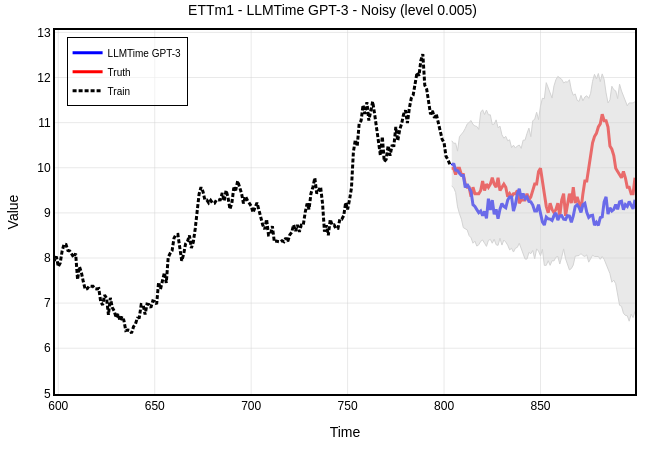} & \figref{0.32}{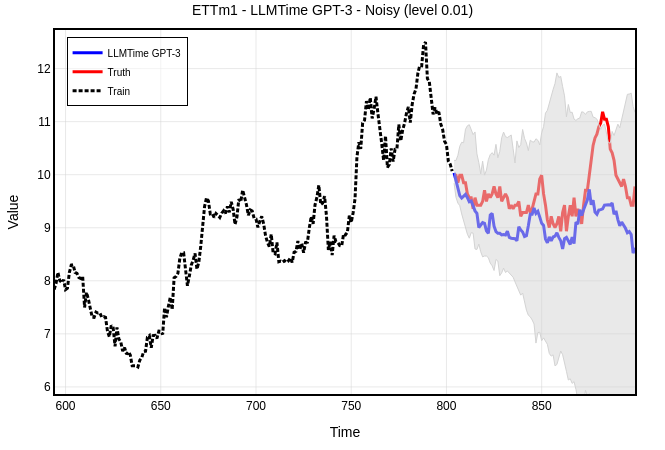} \\
& (a) GPT 3.5 without NLTS & (b) GPT 3.5 with NLTS ($\alpha=0.005$) & (c) GPT 3.5 with NLTS ($\alpha=0.01$).  \\
& \figref{0.32}{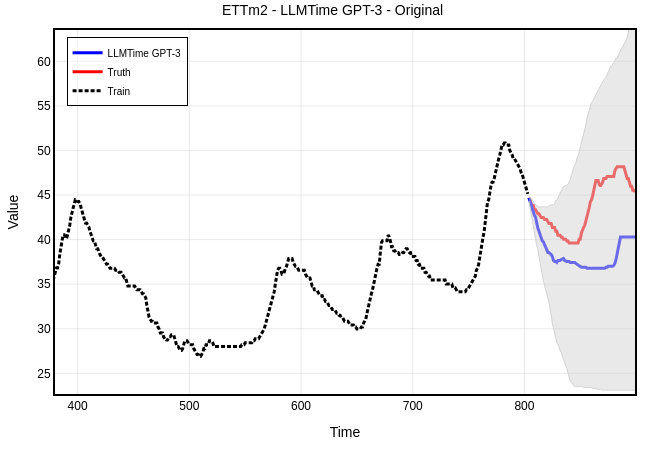} & \figref{0.32}{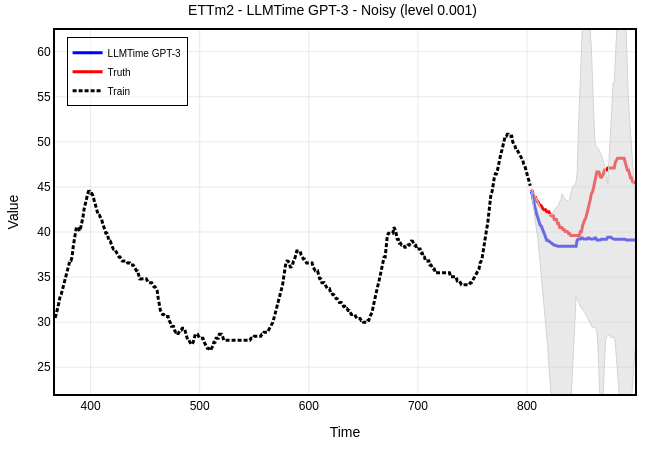} & \figref{0.32}{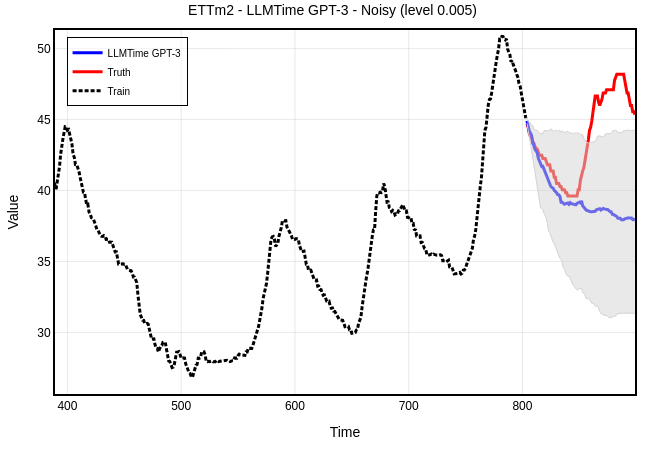} \\
& (a) GPT 3.5 without NLTS & (b) GPT 3.5 with NLTS ($\alpha=0.001$) & (c) GPT 3.5 with NLTS ($\alpha=0.005$).  \\
\end{tabular}\vspace{-3mm}
\caption{Forecasting results of GPT-3.5 on multiple time series from the Autoformer dataset under different noise injection levels. Each row corresponds to a specific dataset (ETTh1, ETTh2, ETTm1, ETTm2), with (a) showing the original prediction without noise, and (b)--(c) showing predictions with increasing levels of noise ($\alpha$).}\label{fig:pred-autoformer2}
\end{figure}

\clearpage

\section{Impact of data contamination}
\label{appendix:without-data-conta}

\subsection{Dataset}
\paragraph{Synthetic TS}
We employ multiple Gaussian Process kernel functions, including \textbf{ExpSineSquared, Linear, Mat\'ern, Polynomial, Rational Quadratic (RQ), and Radial Basis Function (RBF)}, to generate \textbf{1,000 univariate time series} for each kernel. The last \textbf{30} time steps of each series are held out as the forecasting target. For benchmarking, we compare our method against state-of-the-art models such as Autoformer, NSTransformer, TimesNet, and iTransformer, the classical ARIMA model, and zero-shot models like LLMTime. All models are configured following the settings described in Appendix \ref{hyperparameters}

\paragraph{Latest Stock TS}

We collected stock price data and constructed \textbf{eight real-world} time series datasets covering multiple stock indices across different countries. 
For the DJIA and SPX indices, the datasets contain 434 points at hourly frequency and 1612 points at 15-minute frequency. The HS300 index contains hourly and minute-level data with 248 and 992 points, while the SZ300 index comprises hourly and minute-level data with 100 and 960 points, respectively. 

All datasets are collected \textbf{after April 2025}, ensuring that mainstream LLMs (e.g., GPT-3.5, GPT-4, Claude) are unlikely to have seen them during pretraining.
In our experimental setup, the last \textbf{seven time steps} of each series are reserved as forecasting targets, allowing us to evaluate model performance in short-term prediction scenarios. For benchmarking, we adopt the same set of representative baselines as in the Synthetic TS experiments, and the experimental results are reported in Table \ref{tab:mse-mae-stock}.

\begin{table*}[tbph]
\centering
\setlength{\tabcolsep}{4pt}
\begin{tabular}{llccccccccccccccc}
\toprule
\multirow{2}{*}{Datasets} & 
\multicolumn{2}{c}{\shortstack{NLTS 
}} & 
\multicolumn{2}{c}{\shortstack{iTransformer 
}} & 
\multicolumn{2}{c}{\shortstack{LLMTime 
}} & 
\multicolumn{2}{c}{\shortstack{TimesNet 
}} & 
\multicolumn{2}{c}{\shortstack{NSTransformer 
}} & 
\multicolumn{2}{c}{\shortstack{Autoformer 
}} & 
\multicolumn{2}{c}{\shortstack{ARIMA 
}} \\
\cmidrule(lr){2-3} 
\cmidrule(lr){4-5} 
\cmidrule(lr){6-7} 
\cmidrule(lr){8-9} 
\cmidrule(lr){10-11} 
\cmidrule(lr){12-13} 
\cmidrule(lr){14-15}
& MSE & MAE & MSE & MAE & MSE & MAE & MSE & MAE & MSE & MAE & MSE & MAE & MSE & MAE \\
\midrule
DJIAh   & \textbf{0.0004} & \textbf{0.017} & 0.039 & 0.156 & \underline{0.0006} & \underline{0.020} & 0.034 & 0.142 & 0.025 & 0.132 & 0.137 & 0.329 & 0.0016 & 0.033 \\
DJIAm   & \textbf{0.0002} & \textbf{0.009} & 0.020 & 0.121 & 0.0007 & 0.022 & 0.009 & 0.069 & 0.028 & 0.145 & 0.034 & 0.169 & \underline{0.0004} & \underline{0.017} \\
SPXh    & \textbf{0.0002} & \textbf{0.013} & 0.012 & 0.082 & \underline{0.0002} & \underline{0.013} & 0.016 & 0.093 & 0.012 & 0.084 & 0.027 & 0.116 & 0.0007 & 0.020 \\
SPXm    & \underline{0.0002} & \underline{0.011} & 0.010 & 0.086 & 0.1182 & 0.142 & 0.012 & 0.093 & 0.015 & 0.098 & 0.008 & 0.073 & \textbf{0.0001} & \textbf{0.010} \\
SZ300h  & \textbf{0.00031} & \textbf{0.016} & 0.046 & 0.187 & 0.00075 & 0.023 & 0.012 & 0.084 & 0.014 & 0.106 & 0.146 & 0.345 & \underline{0.00082} & \underline{0.023} \\
SZ300m  & \textbf{0.00001} & \textbf{0.003} & 0.001 & 0.022 & 0.00017 & 0.011 & 0.001 & 0.018 & 0.001 & 0.028 & 0.092 & 0.292 & \underline{0.00005} & \underline{0.006} \\
HS300h  & \textbf{0.022} & \textbf{0.089} & 1.276 & 0.950 & 0.120 & \underline{0.188} & 1.385 & 0.968 & 0.923 & 0.821 & 0.878 & 0.794 & \underline{0.054} & 0.199 \\
HS300m  & \textbf{0.0005} & \textbf{0.018} & 0.070 & 0.228 & 0.0011 & 0.030 & 0.044 & 0.185 & 0.029 & 0.139 & 0.147 & 0.347 & \underline{0.001} & \underline{0.026} \\
\bottomrule
\end{tabular}
\caption{
Forecasting performance (MSE / MAE) on Latest Stock TS. \textbf{Bold} and \underline{underline}: the best and the second best results.}
\label{tab:mse-mae-stock}
\end{table*}

\subsection{Average forecasting performance of models}

\begin{figure*}[htbp]
\centering
\begin{tabular}{p{0.0mm}*{3}{c}}
& \figref{0.45}{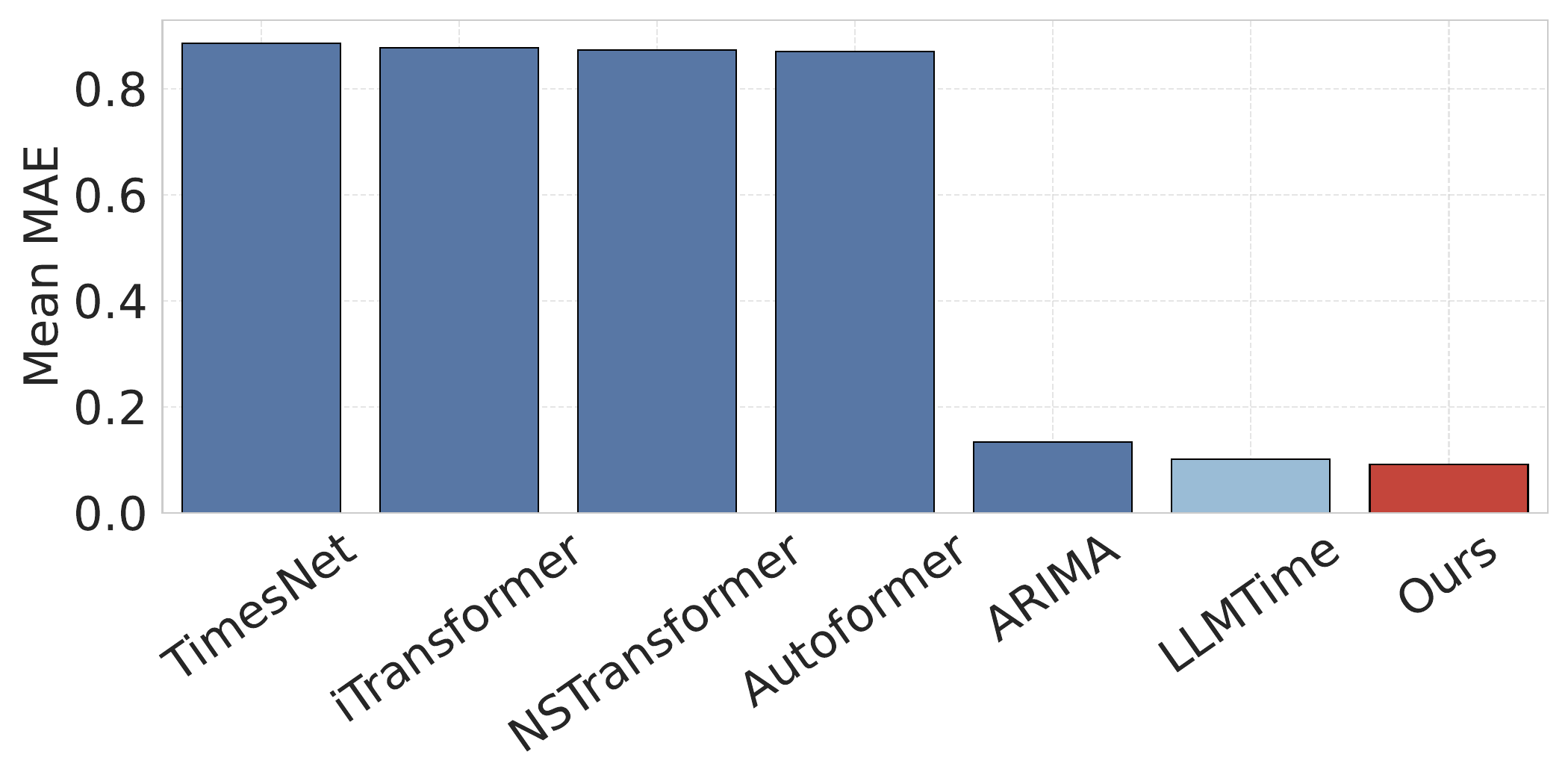} 
& \figref{0.45}{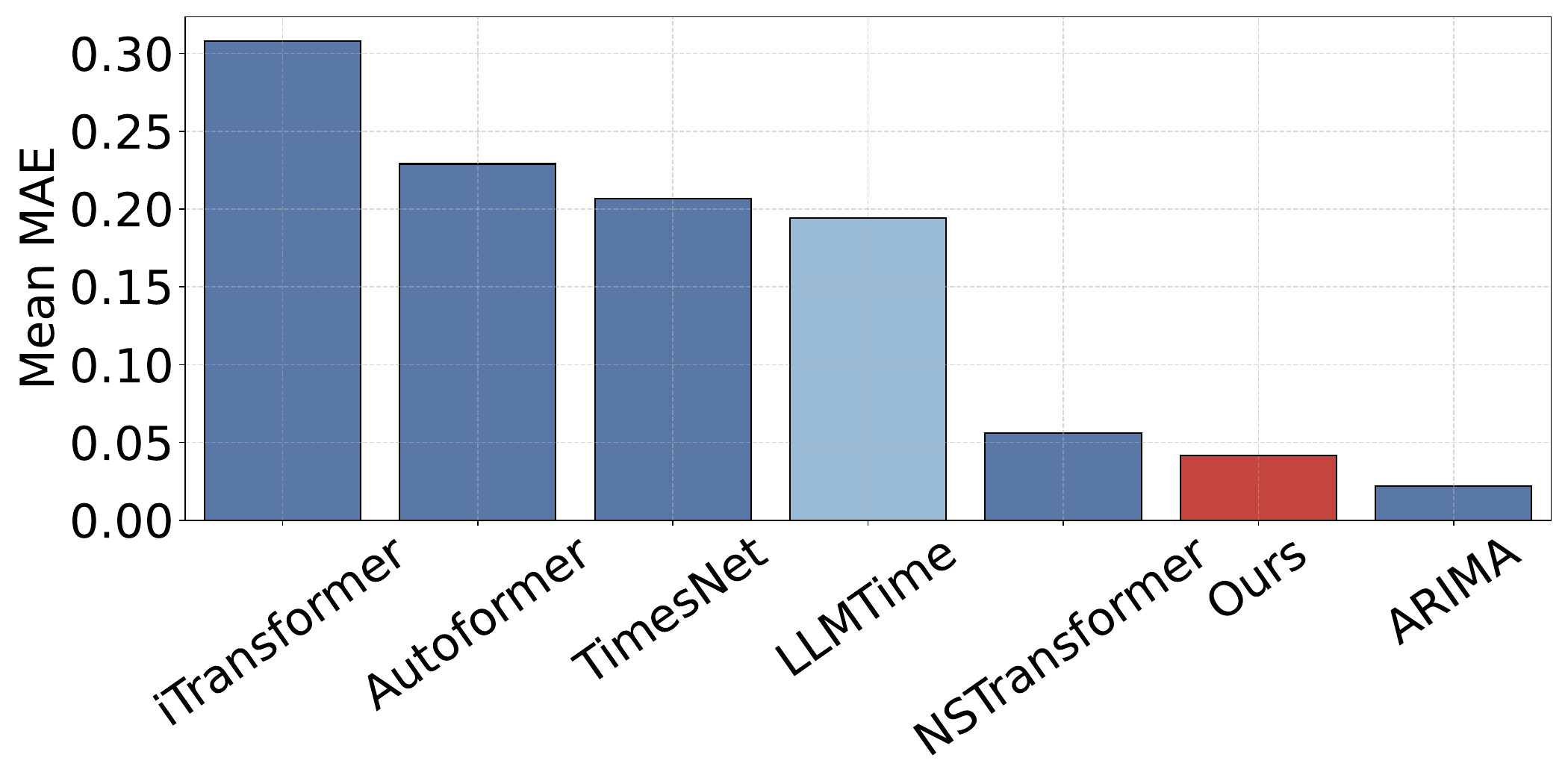} \\
& (a) Synthetic TS
& (c) Latest Stock TS
\end{tabular}
\caption{
Average zero-shot forecasting performance on the synthetic dataset and the latest stock market datasets.
}
\label{fig:unseen data}
\end{figure*}

For contamination-free evaluation, we assess the Synthetic and Latest Stock TS datasets, which contain 6 and 8 sub-datasets. 
\tref{mse-mae-synth-stock} and Figure \ref{fig:unseen data} reports the average zero-shot forecasting results on both the Synthetic and Latest Stock TS datasets. 
Overall, our proposed NLTS model consistently achieves the lowest MSE and MAE across benchmarks, demonstrating superior forecasting capability. 
Since both benchmarks are constructed to eliminate data contamination, the results further indicate that the superior performance of LLM-based approaches (such as ours and LLMTime) does not stem from memorization of the test data, but rather reflects the LLM’s genuine ability to model time series patterns in a zero-shot setting.
Traditional methods, such as ARIMA, achieve relatively good results on real stock datasets, whereas some specialized deep learning methods perform worse in the same tasks. These findings prompt us to reconsider whether conventional time series forecasting benchmarks have become outdated and highlight the need to design more representative benchmarks to accurately evaluate the actual predictive capabilities of models.

\begin{table*}[tbph]
\centering
\setlength{\tabcolsep}{2pt}
\begin{tabular}{llccccccccccccccc}
\toprule
\multirow{2}{*}{Datasets} & 
\multicolumn{2}{c}{\shortstack{NLTS 
}} & 
\multicolumn{2}{c}{\shortstack{iTransformer 
}} & 
\multicolumn{2}{c}{\shortstack{LLMTime 
}} & 
\multicolumn{2}{c}{\shortstack{TimesNet 
}} & 
\multicolumn{2}{c}{\shortstack{NSTransformer 
}} & 
\multicolumn{2}{c}{\shortstack{Autoformer 
}} & 
\multicolumn{2}{c}{\shortstack{ARIMA 
}} \\
\cmidrule(lr){2-3} 
\cmidrule(lr){4-5} 
\cmidrule(lr){6-7} 
\cmidrule(lr){8-9} 
\cmidrule(lr){10-11} 
\cmidrule(lr){12-13} 
\cmidrule(lr){14-15}
& MSE & MAE & MSE & MAE & MSE & MAE & MSE & MAE & MSE & MAE & MSE & MAE & MSE & MAE \\
\midrule
Synthetic TS (avg.) 
& \textbf{0.014} & \textbf{0.091} 
& 1.215 & 0.878 
& \underline{0.016} & \underline{0.102}
& 1.193 & 0.874 
& 1.197 & 0.871 
& 0.029 & 0.135 
& 1.234 & 0.886 \\
Latest Stock TS (avg.)
& \textbf{0.003} & \textbf{0.022} 
& 0.184 & 0.229
& 0.030 & 0.056 
& 0.189 & 0.207
& 0.131 & 0.194
& 0.183 & 0.308 
& \underline{0.007} & \underline{0.041} \\
\bottomrule
\end{tabular}
\caption{
Average zero-shot forecasting performance on synthetic and latest stock TS datasets. 
\textbf{Bold} and \underline{underline}: the best and the second best results in each row.}
\label{tab:mse-mae-synth-stock}
\end{table*}

